\documentclass[letterpaper, 10 pt, journal, twoside]{IEEEtran}
\usepackage{lineno}
\ifCLASSINFOpdf
\else
\fi
\usepackage{graphicx}
\usepackage{graphics}
\usepackage{float}
\usepackage{amsmath}
\graphicspath{{./figures/}}

\usepackage{tabularx}
\usepackage{booktabs}

\usepackage{tikz}
\usetikzlibrary{positioning,decorations.pathreplacing,arrows.meta}
\tikzstyle{EmptyBlock}=[rectangle, draw=blue!80,fill=blue!20, node distance=6em, text centered, minimum height=3em, minimum width=9em, line width=1.5pt, opacity=0]
\tikzstyle{BlueBlock}=[rectangle, draw=blue!80,fill=blue!20, node distance=6em, text centered, minimum height=3em, minimum width=9em, line width=1.5pt]
\tikzstyle{OrangeBlock}=[rectangle, draw=red!80,fill=red!20, node distance=12em,  align=center, text width=0.25*\columnwidth,minimum height=1cm, line width=1.5pt]
\tikzstyle{EndBlock}=[rectangle, draw=red!0,fill=red!0, node distance=12em,  align=left, text width=0.25*\columnwidth,minimum height=1cm]
\tikzstyle{Arrow} = [-{Latex[length=2mm,width=2mm]}, line width=0.4mm]
\tikzstyle{ArrowText} = [sloped, anchor=center,above]

\usepackage{amsmath}
\usepackage{amssymb}  
\usepackage{url}

\newcommand{\Tbf}{\mathbf{T}}

\newcommand{\Ical}{\mathcal{I}}

\newcommand{\Lcal}{\mathcal{L}}

\newcommand{\Vcal}{\mathcal{V}}

\usepackage{subcaption}
\usepackage{multirow}

\usepackage{bm}
\usepackage{algorithm}
\usepackage{algorithmicx}
\usepackage{algpseudocode}

\usepackage[font={small}]{caption}

\usepackage{textcomp}
\usepackage{lipsum}
\usepackage{siunitx}
\usepackage{soul}

\begin{document}

\title{Large-Scale Dense 3D Mapping Using Submaps Derived From Orthogonal Imaging Sonars}

\author{ 
        John McConnell, Ivana Collado-Gonzalez, Paul Szenher,       
        and~Brendan~Englot,~\IEEEmembership{Senior Member,~IEEE}
\vspace{-3mm}
\thanks{
John McConnell is with the Department of Weapons, Robotics and Controls, U.S. Naval Academy, Annapolis, MD, USA, (e-mail: \{jmcconne\}@usna.edu). Ivana Collado-Gonzalez, Paul Szenher and Brendan Englot are with the Department of Mechanical Engineering, Stevens Institute of Technology, Hoboken, NJ, USA (e-mail: \{icollado, pszenher, benglot\}@stevens.edu). 

}
}

\maketitle

\begin{abstract}
3D situational awareness is critical for any autonomous system. However, when operating underwater, environmental conditions often dictate the use of acoustic sensors. These acoustic sensors are plagued by high noise and a lack of 3D information in sonar imagery, motivating the use of an orthogonal pair of imaging sonars to recover 3D perceptual data. Thus far, mapping systems in this area only use a subset of the available data at discrete timesteps and rely on object-level prior information in the environment to develop high-coverage 3D maps. Moreover, simple repeating objects must be present to build high-coverage maps. In this work, we propose a submap-based mapping system integrated with a simultaneous localization and mapping (SLAM) system to produce dense, 3D maps of complex unknown environments with varying densities of simple repeating objects. We compare this submapping approach to our previous works in this area, analyzing simple and highly complex environments, such as submerged aircraft. We analyze the tradeoffs between a submapping-based approach and our previous work leveraging simple repeating objects. We show where each method is well-motivated and where they fall short. Importantly, our proposed use of submapping achieves an advance in underwater situational awareness with wide aperture multi-beam imaging sonar, moving toward generalized large-scale dense 3D mapping capability for fully unknown complex environments. 

\end{abstract}


\IEEEpeerreviewmaketitle

\section{Introduction}
Autonomous underwater vehicles (AUVs) provide great capability to various end users, including offshore oil and gas, renewable energy, and defense. These users utilize AUVs for inspection, environmental monitoring, and security. 

Situational awareness is critical for any autonomous system to complete any task. Autonomous systems use perceptual sensors to interrogate the environment to build an understanding of the vehicle state and the appearance of the environment. However, underwater, these sensing systems are challenged by the environment itself. Conditions such as high water turbidity, low ambient lighting, and water absorption of light make standard sensing systems challenging to apply. Sensing systems such as cameras and LIDAR (Light Detection and Ranging) require clear water conditions or high power requirements, hindering their applicability. 
In contrast, acoustic perceptual systems are robust to environmental conditions and have comparably low power requirements. However, these acoustic systems have a low signal-to-noise ratio and low resolution. 

Acoustic perceptual systems come in wide varieties, including sidescan sonar, profiling sonar, and wide aperture multibeam imaging sonar. Sidescan sonar is typically employed for bottom imaging, especially when searching the seafloor for targets such as shipwrecks or aircraft crash sites. Multibeam profiling sonar is often employed in a downward-looking modality to perform bathymetric mapping. Moreover, profiling sonar can be placed in a forward-looking arrangement onboard a vehicle to detect and avoid obstacles. Profiling sonars are characterized by a narrow vertical beam width, making them precise tools for geometry estimation but challenging to leverage for panoramic situational awareness. In contrast, wide aperture multibeam imaging sonar samples a large volume of water. While imaging sonar has a large sensor volume, it only provides a flattened 2D image of the observed 3D space. 
The question then becomes, how can an autonomous underwater vehicle (AUV) use wide aperture multibeam imaging sonar to achieve the requisite situational awareness for autonomous operation? 

When considering situational awareness, we mainly consider the concept of mapping, i.e., building a data structure that allows the autonomous system to understand the geometry of its environment and how to navigate safely. We focus on building large-scale, low-resolution maps with dense coverage of the environment, supporting tasks like planning and eventually interrogating certain areas with higher-resolution tools. Consider an AUV building a low-resolution map of its work area using acoustic sensors due to the environmental conditions. Using the map, the AUV can plan a path to a goal position, collecting high-resolution data with cameras at close range, or even interrogating a structure with a non-destructive-testing (NDT) probe. 

This article builds on our previous work building 3D maps using wide-aperture multibeam imaging sonars. To address the lack of 3D information in a single sonar image, we use a system that combines a pair of orthogonal sonars onboard a small remotely operated vehicle (ROV), proposed in \cite{McConnell-2020}. However, because 3D information is only recoverable where the sonars overlap, we considered an inference-based mapping system where simple, repeating objects in the environment accelerate mapping and enhance situational awareness \cite{McConnell-2021}. In this evolutionary paper, we extend the work described in these two prior conference papers, using pairs of wide aperture multibeam imaging sonars to build large-scale dense 3D maps of complex underwater environments, using a submapping approach that has never previously been applied in this setting. In the following sections, we will discuss related work, define the problem, discuss our system in detail, and compare the newest mapping system to our previous work in this area. The paper is structured as follows: Section \ref{lit-rev} shows the related work. Section \ref{problem-def} shows
the problem definition. Section \ref{SLAM} introduces the simultaneous
localization and mapping (SLAM) formulation used throughout
this work. Section \ref{sonar-fuse} discusses fusing orthogonal sonars as
introduced in \cite{McConnell-2020,McConnell-2021}. Section \ref{infer_map} shows mapping using object-specific inference from \cite{McConnell-2021}. Section \ref{submapping-main} shows the authors’
newest work, submapping. Section \ref{experiments} shows the experiments.
Finally, Section \ref{conc} concludes this article.

\section{Related Work}
\label{lit-rev}
\subsection{3D Reconstruction Using Wide Aperture Imaging Sonar}
Wide aperture multibeam imaging sonars stand in contrast to other sensors in this space; they have reasonable power requirements and are robust to environmental conditions such as water quality and ambient lighting. However, imaging sonar has a critical issue that hinders its usability as a tool for 3D situational awareness, the lack of elevation angle in its imagery. While the sonar observes a 3D volume of water, the sensor only reports acoustic intensity, distance, and bearing angle, leaving the elevation angle unknown.

The challenges associated with wide-aperture multi-beam imaging sonars have inspired an impressive body of work to address the fundamental limitations of their under-constrained measurements. Firstly, work from Aykin \cite{Aykin-2013, Aykin-2013-1} estimates the elevation angle of sonar image pixels in scenes where objects are lying on the seafloor.

The work of Westman \cite{Westman-2019} extends the work of Aykin and shows excellent results in a constrained nearshore pier environment. 
However, these methods rely on several assumptions that may often be violated. Firstly, both \cite{Aykin-2013-1} and \cite{Westman-2019} assume that all objects in view have their range returns monotonically increase or decrease with elevation angle. While this assumption may hold true for some objects, it hinders the application of their methods to arbitrary objects and scenes. Additionally, \cite{Aykin-2013-1} requires visibility of the leading and trailing edge of an observed object; this is obtained by examining the shadow area behind a segmentation created by the sonar's downward grazing angle. In contrast, \cite{Westman-2019} only needs to identify the leading \textit{or} trailing edge of the object; however, in the experiments
shown, the leading edge is always the closest return because of the sonar's downward grazing angle. Using a downward grazing angle makes the problem significantly simpler to solve, but comes at a price. By tilting the sonar downward, and making the upper edge of the sonar beam parallel with the water plane, the AUV's situational awareness may be hampered. In cluttered environments, an AUV would be unable to see above it before transiting upward, and a safe navigation solution may not always be possible. Moreover, if the vehicle is perturbed in a way that violates this geometric assumption, the perception system could be driven to inaccuracy. 

We also note that a recent technique has employed deep learning with convolutional neural networks to estimate the elevation angle associated with imaging sonar observations \cite{DeBortoli-2019}. More recently, \cite{Wang-2022} uses a cost volume and multiple views to perform training. Neural Radiance Fields (NeRFs) have also been leveraged with imaging sonar \cite{Qadri-2023}. NeRFs, however, have yet to be extended to arbitrary scenes, rather than single objects with imaging sonar. Lastly, when considering machine learning, \cite{Arnold-2022} uses a truncated signed distance function to render a 3D polygon mesh. Further, a vehicle may lack opportunities for prior training and exposure to the subsea objects and scenes it may encounter in a given mission.  

Another method proposed by Aykin \cite{Aykin-2017} applies a space carving approach to produce surface models from an image's low-intensity background, which outer-bound the objects of interest. The min-filtering voxel grid modeling approach from Guerneve \cite{Guerneve-2018} similarly removes voxels from an object model based on observations of low-intensity pixels. Westman \cite{Westman-2020} improves on space carving using Fermat paths. Westman \cite{Westman-2020-2} proposes a volumetric framework for recovering 3D geometry, testing with both wide and narrow aperture sonar. These approaches require the objects of interest to be observed from multiple vantage points to achieve accurate reconstructions. Moreover, it is unclear how these methods would be applied when the robot pose estimates evolve, changing as constraints are introduced, resulting in significant pose estimate updates. 

We also note that an alternative approach is to employ a single sonar that scans in 3D. Mechanically scanning 3D sonars are widely available, and utilized for underwater 3D mapping applications, although they are typically mounted at a static location on the seafloor, and are not designed to be carried by mobile robots \cite{blueview, 3d-scanning}. Electronically scanning, beam steering 3D sonars have been proposed and prototyped \cite{3d-fls-1}, and can be implemented as forward looking sonars that are portable aboard an underwater vehicle. However, such sonars are presently high in cost, and require significant time to sweep the desired 3D volume \cite{3d-fls-2}.
More compact 3D sonars are emerging \cite{waterlinked}, but are presently severely limited in sensing range.



Elevation angle recovery has also been addressed from a simultaneous localization and mapping (SLAM) perspective. Rather than trying to estimate the elevation of pixels in a single frame, these works extract features from the imagery and use a series of views combined with a pose graph optimization back-end \cite{Kaess-2008} to determine 3D structure. This concept was proposed by Huang \cite{Huang-2016} in acoustic structure from motion (ASFM). This initial implementation has limitations, the chief of which is the reliance on manually extracted features. This work was later built on by Wang \cite{Wang-2019}, incorporating automated feature extraction and tracking. \cite{pedro-2019} shows a SLAM-based approach that recovers the geometry of a ship hull rather than focusing on terrain reconstruction. The limitation of these methods for AUVs in clutter is the perception system requiring a series of frames to recover 3D information rather than a single timestep. 

\subsection{Fusing Multiple Wide Aperture Multi-beam Imaging Sonars}
Combining a pair of imaging sonars to recover the missing data at each vantage point provides unique advantages to an underwater vehicle. First proposed in \cite{Assalih-2009, Negahdaripour-2020} and later implemented onboard an underwater vehicle in \cite{McConnell-2020, McConnell-2021}, the most recent performance analysis has been applied to specific classes of objects \cite{Sadjoli-2022}. McConnell \cite{McConnell-2020} associates the objects in view, then extracts and matches pixel features using the object associations as an optimization constraint. McConnell \cite{McConnell-2021} makes improvements to the orthogonal sonar fusion algorithm by using sonar range as an optimization constraint, enhancing performance in complex scenes. 

These techniques require no assumptions about scene geometry, removing any object-level geometric assumptions from a reconstruction system and paving the way for mapping arbitrary scenes. Critically, when considering situational awareness, a pair of orthogonal sensors provides a set of fully defined 3D points at \textit{every timestep.} However, using a pair of orthogonal sensors yields 3D data only inside the overlapping region of their fields of view. The hardware used in this paper and \cite{McConnell-2020, McConnell-2021} gives an overlapping area of 20$^{\circ} $-by-20$^{\circ}$, making large-scale mapping a potentially lengthy process. The crucial challenge for these systems is integrating the 3D data they generate into a complete mapping solution and accounting for their limited field of view. 

\subsection{Inference Aided 3D Reconstruction and Mapping}
A large body of work applies probabilistic inference to enhance 3D mapping and reconstruction. Most closely related are the methods that use inference to improve point cloud or voxel mapping, which often use insights from large data sets or object models provided a priori, which is not always available in an underwater robotics setting.  

The use of a variational auto-encoder to infer the 3D distribution of an object given a single view from an RGB camera is explored in \cite{Yu-2018, Yu-2019}. As mentioned above, this work requires an extensive data set to pre-train the network weights. Yang \cite{Yang-2019} explores synthetic data from a generative adversarial network to avoid this requirement. Yang \cite{Yang-2019} uses a single view voxel grid as the input for 3D reconstruction; the focus is 3D-to-3D densification rather than the 2D-to-3D inference considered in our work. A similar concept to our work is explored in \cite{Moreno-2013}, where mapping is performed at an object level. Here, object models are produced using high-quality depth camera scans from a controlled setting. These scans are used to improve a 6-DoF SLAM solution. Similar to \cite{Yu-2018, Yu-2019}, the applications of this approach in underwater robotics are limited due to the need for object models a priori. A notable use of inference in an underwater setting is \cite{Guerneve-2017}, where CAD (computer aided drawing) models of objects of interest are provided a priori and are used to improve the map output. 

A related set of methods use probabilistic inference to enhance occupancy grid maps  \cite{Wang-2016, Doherty-2017, Meadhra-2018, Gan-2020}. While these methods are excellent at improving occupancy map coverage, they solve mapping under sparse inputs by performing gap-filling and semantic inference on a data structure of the same dimensionality as those sparse inputs. In contrast, we focus on 2D-to-3D inference to enhance the dimensionality of inputs that are not directly observed in 3D, whose inference is conditioned on prior 3D observations of the same class.

In our previous work, we propose using Bayesian inference to accelerate the mapping process. We again note the limited overlapping area for a pair of orthogonal wide aperture imaging sonars, in our case 20$^{\circ} $-by-20$^{\circ}$. This limited overlapping area, combined with a 130$^{\circ}$ horizontal field of view, motivates our work in \cite{McConnell-2021}. \cite{McConnell-2021} leverages a pre-trained semantic classification model and online estimation of specific object classes in the environment, both simple and repeating objects (such as pier pilings). Using these simple repeating objects, more of the 130$^{\circ}$ field-of-view sonar image is used for mapping, and the density of map coverage is greatly enhanced. However, this method requires two items for functionality and performance: a trained model with known classes in the environment, and the presence of those objects. 

\subsection{Submapping}
In many robot mapping methods, discrete timesteps are selected to perform pose estimation and enter data into a map. However, it is often the case that when implementing a system like this, the sensor data between discrete timesteps is discarded. Conversely, submapping does not discard this data between timesteps; instead, it is captured and leveraged.

A notable example of submapping in underwater robot mapping using imaging sonar is \cite{Teixeira-2016}. \cite{Sodhi-2019} exploits submapping to create an updatable occupancy grid, enabling a grid structure to be used when robot poses experience significant updates. This submapping technique can be used to create dense 3D maps, especially when input data may be sparse. In this paper, we will utilize the submapping paradigm to enhance mapping and dispense with the need for prior information.   

\section{Problem Definition}
\label{problem-def}
In this work, we consider 3D mapping using a pair of orthogonal imaging sonars with an overlapping field of view. A robot visits a series of poses $x_t$, with transformations  $\Tbf \in \mathbb{R}^{4\times 4}$. Each pose has associated observations $z_t$, with two components: horizontal sonar observations $z^h$ and vertical sonar observations $z^v$. Each set of observations is defined as an intensity image in spherical coordinates with range $R\in \mathbb{R}_+$, bearing $\theta\in \Theta$, and elevation $\phi \in \Phi$, with $\Theta,\Phi \subseteq [-\pi,\pi)$, and an associated intensity value $\gamma \in \mathbb{R}_+$. These measurements can be converted to Cartesian space: 
\begin{linenomath*}
\begin{align}
    \begin{pmatrix}  X \\  Y \\  Z \end{pmatrix} 
    = R\begin{pmatrix} \cos{\phi} \cos{\theta} \\ 
    \cos{\phi}\sin{\theta} \\ 
    \sin{\phi} \end{pmatrix}.
    \label{eq:tx_to_cartesian}
\end{align}
\end{linenomath*}
Each recorded measurement $z^h$ and $z^v$ is characterized by an omitted degree-of-freedom (DoF), 
and in the robot frame, due to the orthogonality of the two sonars, these DoFs differ:  
\begin{linenomath*}
\begin{align}
    z^{h} &= (R^{h},\theta,\gamma^{h})^\top, &z^{v} &= (R^{v},\phi,\gamma^{v})^\top.
    \label{eq:observations}
\end{align}
\end{linenomath*}
We associate measurements across concurrent, orthogonal images to fully define the measurements in 3D, yielding Equation \eqref{eq:fused}:
\begin{linenomath*}
\begin{align}
    {z}^{Fused}=\left(\frac{R^{h}+R^{v}}{2},\theta^{(h)},\phi^{(v)} \right)^\top.
    \label{eq:fused}
\end{align} 
\end{linenomath*}
We complete the 3D mapping problem by placing each set of observations into a fixed frame
$\Ical$ as in Eq. \eqref{eq:map}. 
\begin{linenomath*}
    \begin{align}
    \mathcal{M} = \{\hat{z}^{(\Ical)}  | \hat{z}^{(\Ical)} = \Tbf \hat{z}^{Fused} \ \forall \  \hat{z}\in  \widehat{z}\}
    \label{eq:map}
\end{align}
\end{linenomath*}
We note that the lack of 3D information in the sonar imagery 
introduces ambiguity into the application of Eqs. \eqref{eq:tx_to_cartesian} and \eqref{eq:map}, requiring the association of observations to determine their location in 3D space as in Eq. \eqref{eq:fused}.

\section{Simultaneous Localization and Mapping (SLAM)}
\label{SLAM}
In this section, we will describe our pose estimation system. While not the focus of this work, nor a claim of novelty, this module influences the system level design of the downstream mapping pipeline. We utilize a pose SLAM formulation to estimate our robot's pose history through time. We restrict our formulation to 3-DoF estimation in the plane to provide an efficient and robust SLAM pipeline that prioritizes the DoFs of greatest uncertainty (surge, sway, and yaw). In this section, we will describe the handling of incoming sonar images, dead reckoning data, and our keyframe-based SLAM system.

\subsection{Processing Sonar Observations}
While our vehicle has a dual sonar system, only the horizontal sonar observations $z^h$ are used to support state estimation. This horizontal sonar image is segmented as described in Section \ref{feature-ex}. The segmentation identifies which image pixels correspond to structures in the environment. The sonar image segmentation is converted to a \textit{planar} point cloud by setting the unknown elevation angle $\phi$ to zero. The following sections will use this planar point cloud to support vehicle state estimation. 

\subsection{Dead Reckoning}
While our system uses sonar observations to perform state estimation, it is built on a dead reckoning system to provide initial estimates for robot poses, $x_t$. We consider a robot with an inertial measurement unit (IMU) and a doppler velocity log (DVL). While specific sensor rates may vary, the IMU refresh rate in our system is 200 Hz, and the DVL refresh rate is 5Hz, with combined ``dead reckoning`` data output at 5Hz.  

The IMU provides an estimate of the rotation matrix between the vehicle frame $\Vcal$ and $\Ical$, denoted as $R^{\Ical}_{\Vcal} \in \mathbb{R}^{3\times 3}$. Complimenting the IMU, the DVL provides velocity estimates in the vehicle frame, in the surge, sway, and heave directions. These sensors are combined to estimate the vehicle location without perceptual observations, using dead reckoning only. We denote the dead reckoning based pose estimate as $o_t$. The fusion of these two sensors is not at the core of this paper's discussion. However, we must note that we use this odometry system as an initial guess for our keyframe-based SLAM system. Moreover, when between keyframes, to provide live pose estimates, we use the dead reckoning system to provide pose estimates relative to the latest keyframe.  

\subsection{Keyframe Graph Based Pose SLAM}
Our SLAM system follows the graph based pose SLAM paradigm, where discrete time steps are selected, and measurements are developed to support their estimation. These discrete time steps known as \textit{keyframes} have three relevant quantities: a provided horizontal sonar image $z^h$, a provided dead reckoning pose estimate $o_t$, and an unknown robot pose $x_t$. Upon receipt of $z^h$, and $o_t$, we check if a new keyframe needs to be instantiated. Keyframes are added when the current distance or rotation relative to the previous keyframe is large enough per the dead reckoning system. Note that keyframes are not added at a fixed rate, but are dependent on vehicle velocity. 

When a keyframe is instantiated, we add a node in the SLAM graph to estimate the robot location $x_t$. We then add a series of measurements or edges in the graph structure. We use two measurements: sequential scan matching (SSM) and non-sequential scan matching (NSSM). First, we compute the SSM factor using iterative closest point (ICP) using the dead reckoning pose, $o_t$, to provide ICP's required initialization. Here we compare the newest keyframe's horizontal sonar observations $z_t^h$ to the last keyframe's $z_{t-1}^h$. Recall that observations are planar inside the SLAM system, as is the resultant transformation estimate from ICP. This transformation is formulated as a factor in the factor graph, an SSM factor, $f^{\text{SSM}}$. In the event of an ICP failure, to preserve the graph connectivity, we insert an odometry factor, which is simply the pose estimate from dead reckoning relative to the last keyframe, $f^{\text{O}}$.

Next, we consider NSSM factors, also known as loop closures. We compare the most recent sonar observations $z_t^h$ to the aggregated keyframes in the fixed frame $\Ical$ using ICP. This transformation is between the current pose $x_t$ and another pose in the graph, excluding the several previous frames. We denote this loop closure factor as $f^{\text{NSSM}}$. NSSM factors are subject to a rigorous outlier rejection scheme, Pairwise Consistent Measurement Set Maximization (PCM) \cite{Mangelson-2018}. NSSM factors approved by PCM are entered into the pose graph, completed below:

\begin{linenomath*}
\begin{flalign*}
\mathbf f(\boldsymbol \Theta) = \mathbf  f^{\text{0}}(\boldsymbol \Theta_0) & \prod_i \mathbf f^{\text{O}}_{i}(\boldsymbol \Theta_i) \prod_j \mathbf f^{\text{SSM}}_j(\boldsymbol \Theta_j) \prod_q \mathbf f^{\text{NSSM}}_q(\boldsymbol \Theta_q).
\end{flalign*}
\end{linenomath*}

This planar SLAM solution is used to provide estimates of surge, sway and yaw; the remaining degrees of freedom come from our vehicle's pressure sensor and inertial measurement unit (IMU). We enable this by operating our vehicle at a \textit{fixed depth} and noting that roll, pitch, and depth are directly observable. The planar SLAM system performs inference over relative measurements to estimate the degrees of freedom with the most uncertainty, as they are not directly observable. This combination yields a vehicle state estimate with 6 degrees of freedom. We implement this system using the GTSAM \cite{GTSAM} implementation of iSAM2 \cite{Kaess-2011}. We note that this system relies on the features in view, and that in their absence, the system relies solely on dead-reckoning.

\section{Fusing Orthogonal Concurrent Wide Aperture Sonar Images}
\label{sonar-fuse}
\begin{figure}[t]
\centering
{\includegraphics[height=2.5cm]{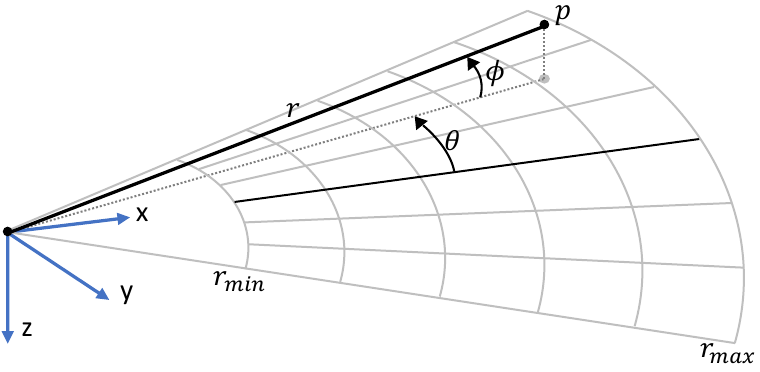}}
\caption{\textbf{Example sonar geometry.} Sensor origin is shown at left, with max range denoted. The sensor swath, or horizontal field of view is shown as $\theta$ and the vertical aperture is shown as $\phi$. We note that an example $\theta$ and $\phi$ are shown to point $p$.}
\vspace{-2mm}
\label{fig:sonar_geo}
\end{figure}

This section will describe our method for recovering 3D information from a pair of orthogonal sonar images \cite{McConnell-2020}. Recall that each sonar image, whose geometry is illustrated Fig. \ref{fig:sonar_geo} contains partial 3D information, each image with a missing angle. First, we will identify features in sonar imagery, then associate those features. Next, we use an outlier rejection system to cull low-quality feature associations. The output of this subsystem will be a 3D point cloud used to support various 3D mapping systems. 

\begin{linenomath*}
    \begin{figure}[t]
\centering
\includegraphics[height= 2cm]{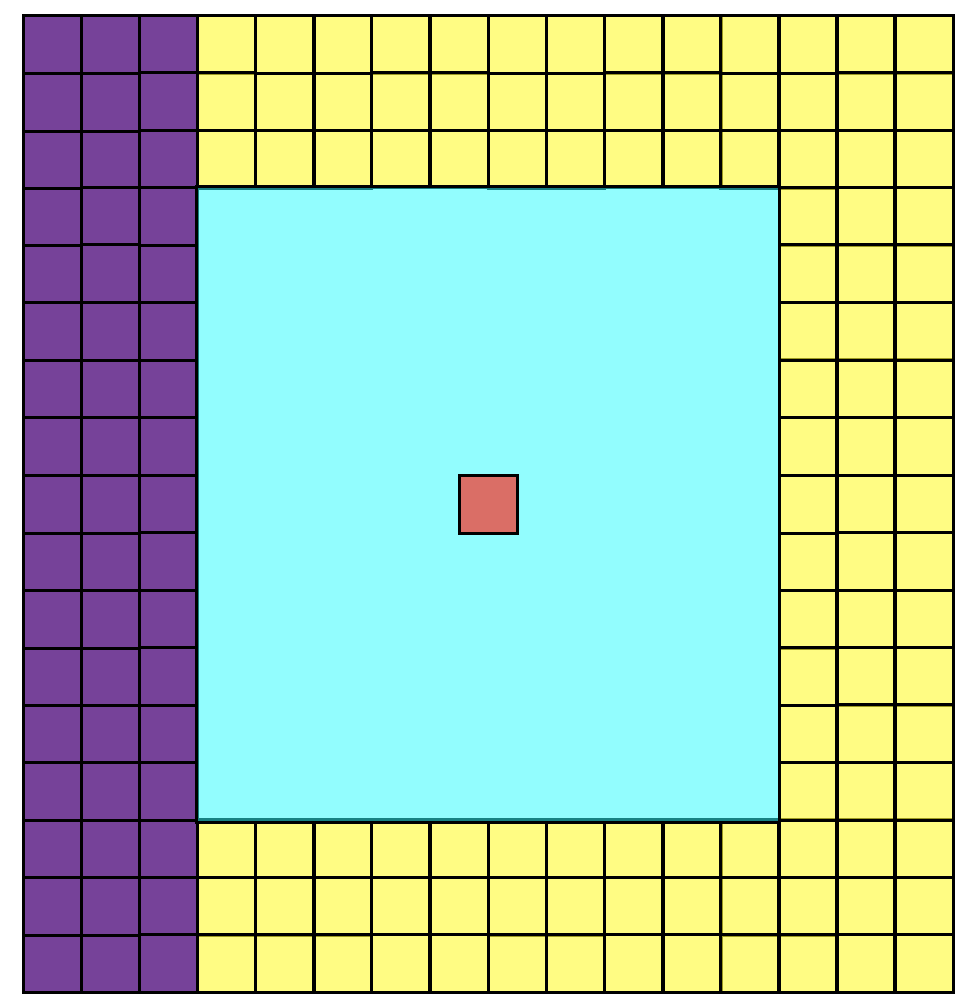} \ \
\includegraphics[height= 2cm]{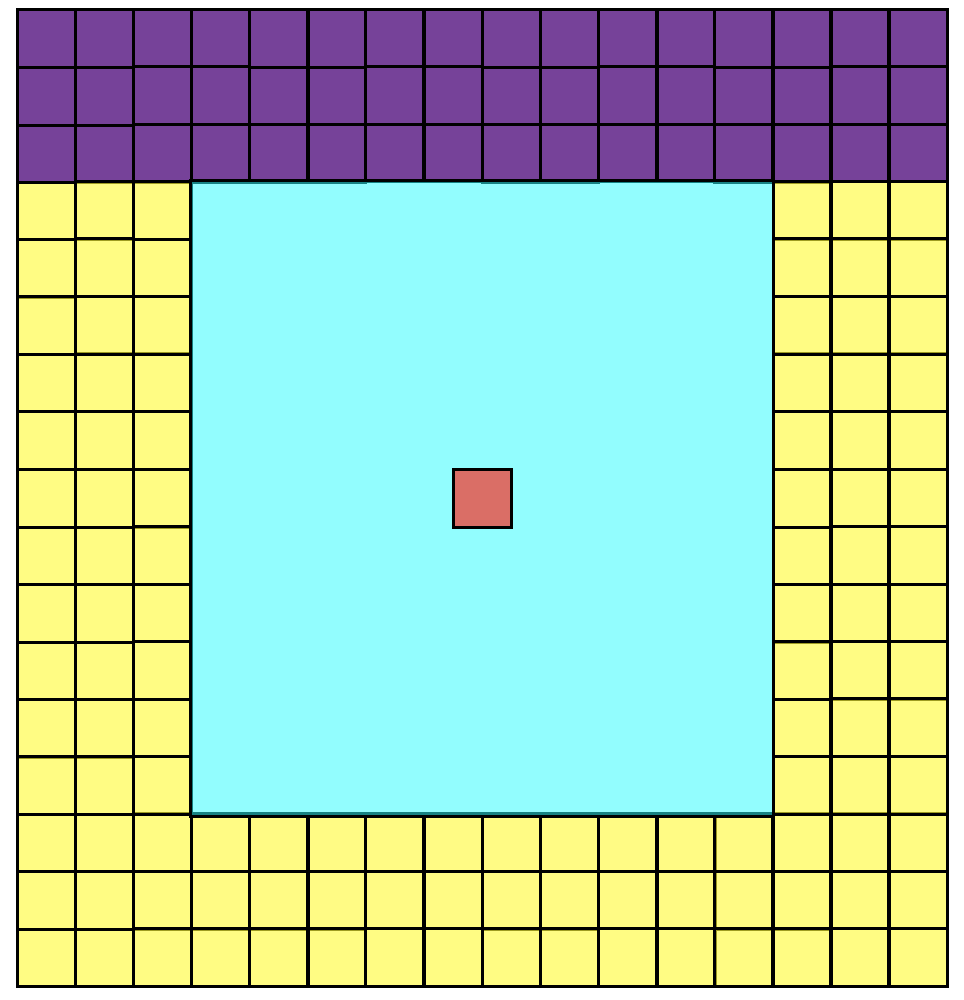} \ \
\includegraphics[height= 2cm]{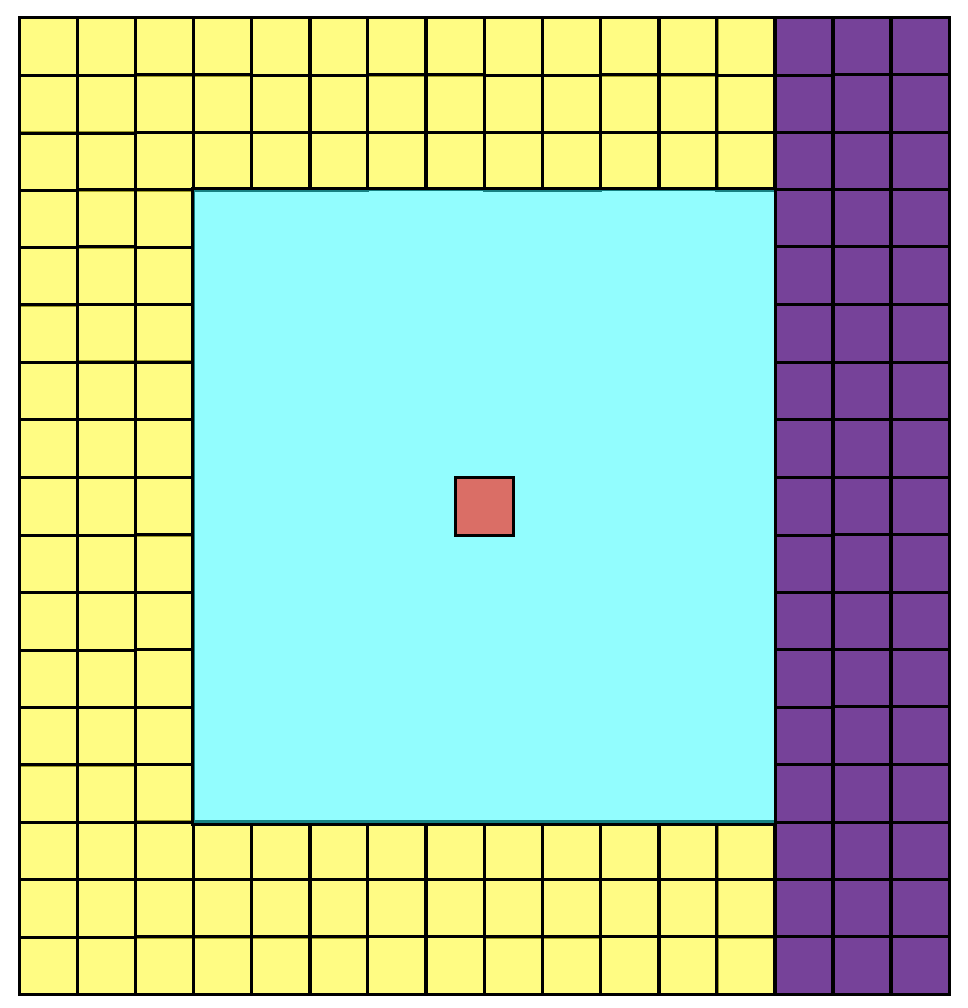}\ \
\includegraphics[height= 2cm]{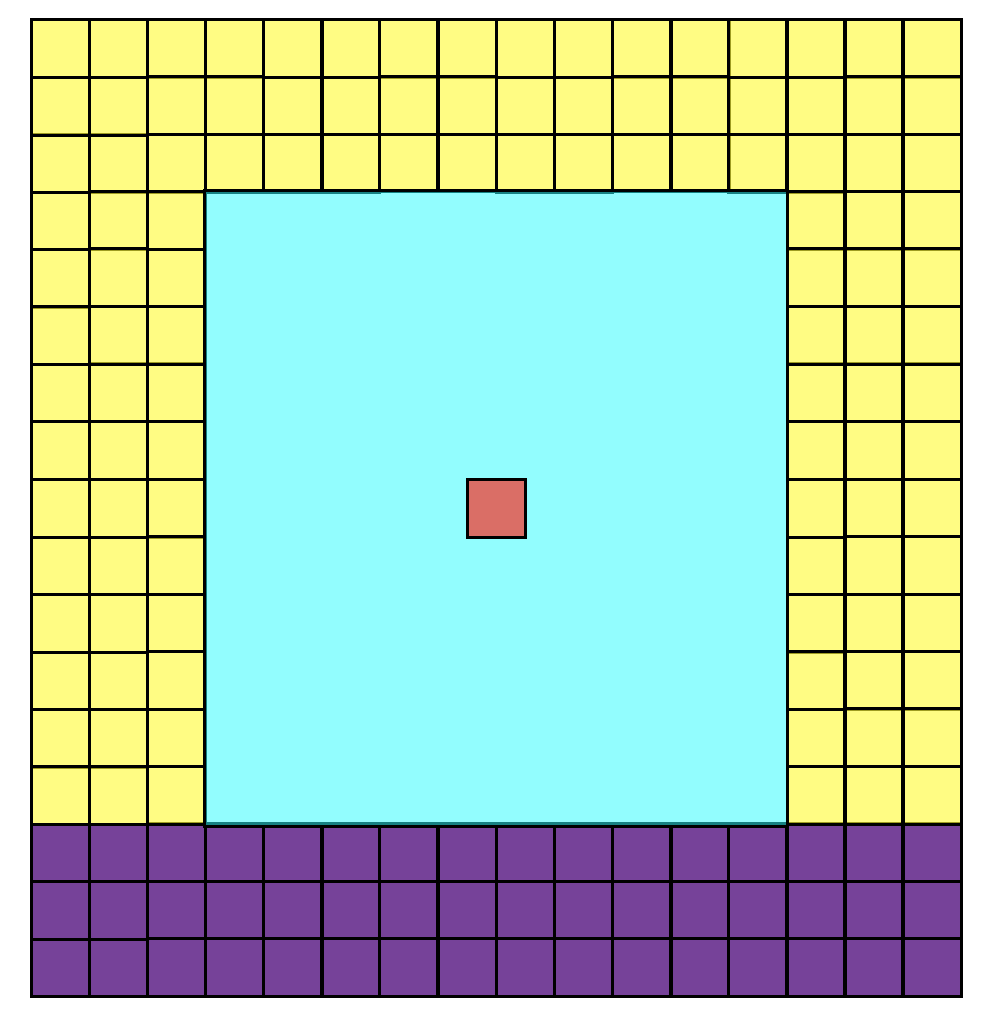}
\caption{\textbf{SOCA-CFAR overview.} Purple cells show training cells, blue the guard cells, and red the cell under test. }
\label{fig:cfar}
\end{figure}
\end{linenomath*}

\subsection{Feature Extraction}
\label{feature-ex}
The first step in our orthogonal sonar fusion system is to identify which sonar image pixels correspond to contact with the environment. This is done using the constant false alarm rate (CFAR) algorithm, specifically smallest of cell averages (SOCA-CFAR) \cite{Richards-2005}. SOCA-CFAR is selected as it alleviates issues with CFAR when multiple targets are present in the image.

SOCA-CFAR uses a simple threshold to determine whether a pixel in the image is a contact; however, this threshold is computed dynamically by characterizing the noise around the image pixel. Consider the pixel shown in red in Figure \ref{fig:cfar}. Four averages are computed around the pixel; the average of the training pixels is shown in purple in Figure \ref{fig:cfar}. The average is simply 
\begin{linenomath*}
    \begin{align}
\mu= \frac{1}{N} \Sigma_{i=0}^{N} p_i,
\end{align}
\end{linenomath*}
where $N$ is the number of training pixels and the value of each pixel is $p_i$. We only record the minimum of the four averages, $\mu_{min}$, the minimum of the areas shown in purple in Figure \ref{fig:cfar}. Each pixel is wrapped in guard cells, shown as light blue in Figure \ref{fig:cfar}, to prevent the signal from leaking into the noise estimate. Next, the detection constant, $\alpha$, is computed: 
\begin{linenomath*}
    \begin{align}
\alpha = N (P_{fa}^{-1/N} - 1 ),
\end{align}
\end{linenomath*}
where $N$ is the number of training pixels and $P_{fa}$ is the user's false alarm rate. Lastly, $\beta$ is computed and compared against the user-set threshold as follows:
\begin{linenomath*}
\begin{align}
\beta = \mu_{min} \alpha.
\end{align}
\end{linenomath*}
At this stage in the system, we have identified pixels in the vertical and horizontal sonar images representing contact with the environment. In the next section, we will address the association of the image pixels across concurrent sonar images. 

\subsection{Sonar Fusion}
After SOCA-CFAR is applied, our system now has a pair of segmented sonar images, identifying contacts with the environment. However, these images still lack 3D information, with each image lacking a single angle. To fully define measurements in 3D as denoted in Eq. \eqref{eq:fused}, we need to associate pixels across the images. 

Here we divide the image-pixel matching problem into multiple smaller subproblems. Recall that range $R$ is discretized within a sonar image, and since associated pixels should be at the same range, we use range to define these subproblems. Each sonar image's pixels at the same range are gathered and processed using intensity-based association. The cost function used here is defined as follows:
\begin{linenomath*}
    \begin{align}
    \Lcal(z_i^{h},z_j^{v}) = || \nu^h - \nu^v||,
    \label{eq:cost}
\end{align}
\end{linenomath*}
where $\nu^h$ and $\nu^v$ are square patches of the sonar image around the given pixel. Note that $\nu^v$ is rotated 90 degrees to account for the orthogonality of the images. Moreover, before this comparison is made, the images' intensity values are normalized at every timestep. The cost function in Eq. \eqref{eq:cost} is used to find the solution that minimizes the sum of costs between features for each subproblem.

To estimate our confidence in these matches, we compare the two best solutions for each feature association \cite{Hu-Mor-2010}: 
\begin{linenomath*}
\begin{align}
    C = \frac{\Lcal(z_i^{h},z_j^{v})_{min 2}- \Lcal(z_i^{h},z_j^{v})_{min}}{\Sigma_{0,0}^{i,j} \Lcal(z_i^{h},z_j^{v})}.
    \label{eq:uncer}
\end{align}
\end{linenomath*}
While we compute subproblem cost totals to find sets of associated features in Eq. \eqref{eq:cost}, confidence is evaluated on a pixel-wise basis, comparing the costs for each pixel association made in the given solution. This comparison gives us a simple metric with which to cull uncertain associations. If a match has confidence outside our set requirements, it is not retained. We now have a set of fully defined points in 3D space, $z^{Fused}$, and can map them into Cartesian space using Equation \eqref{eq:tx_to_cartesian}, obtaining a 3D point cloud from a pair of orthogonal sonar images. 

\subsection{Sonar Fusion Mapping}
\label{fusion_mapping}
Once 3D point clouds are recovered from the sonar fusion system, we need to place these clouds in a fixed frame to build the map in Equation \eqref{eq:map}. Recall that our SLAM system estimates robot poses at discrete time steps. Consequently, we only perform mapping at these same time steps. In our implementation, each keyframe in the SLAM system retains $z^{Fused}$ as a 3D point cloud, registering it to the fixed frame $\Ical$ per Equation \eqref{eq:map}, denoted as \textit{sonar fusion mapping}. 

\begin{figure*}[t]
\centering
{\includegraphics[height=3.00cm]{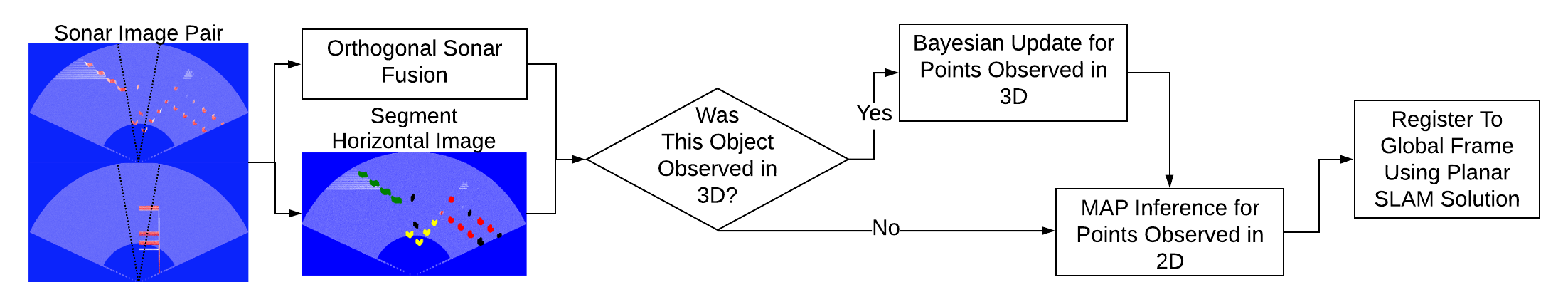}}
\caption{\textbf{Mapping via object specific Bayesian inference block diagram.} A pair of orthogonal sonar images is provided as input (black lines bound the region of overlap between the two sonar fields-of-view). The images are processed according to Section \ref{sonar-fuse}. The horizontal image is segmented as in Section \ref{sonar-seg} (colors denote different object classes - seawall in green, rectangular pilings in yellow, cylindrical pilings in red). The resulting 3D points enrich each object's model (Eq. \eqref{eq:bayes rule}), while MAP inference is applied to 2D points (Eqs. \eqref{eq:query}, \eqref{eq:query2}).
We then use the planar SLAM solution to register the resulting point cloud. The synthetic sonar images shown here are sampled from the virtual environment depicted in Fig. \ref{fig:marina_2_pic}. }
\label{fig:object_mapping_flow}
\end{figure*}

\section{Mapping via Object Specific Bayesian Inference}
In this section, we will briefly describe our object specific Bayesian mapping system \cite{McConnell-2021}. We note that only part of the sonar images can be used to recover 3D information where the sonars overlap. This part of our system aims to utilize the remaining parts of the sonar images, not only the 20$^{\circ} $x20$^{\circ} $ overlapping area. First, we will identify and classify objects; then, if an object is present inside the overlapping area, we update our estimate of that object class's geometry. If an object is present outside the overlapping area, we will query the object class's geometry to fill in the missing dimension in the sonar image. This object-specific mapping will be used on objects that are both simple and repetitive to accelerate the mapping process. 

\subsection{Object Identification and Classification}
\label{sonar-seg}
Our object-specific mapping system's first step is to identify and classify objects in the horizontal sonar images. We start with the image pixels classified as contact from Section \ref{feature-ex}. These image pixels are clustered into object instances of unknown classes using DBSCAN \cite{Ester-1996}. DBSCAN is used because it is a density-based clustering algorithm that does not require the number of clusters to be specified \textit{a priori}. Once each object is identified using DBSCAN, a bounding box is developed around each object. This bounding box is cropped from the original sonar image and fed to a convolutional neural network (CNN) for semantic classification. 

In this work, we use a simple neural network to perform semantic labeling of object instances. Specifically, we use a CNN that accepts a 40x40 pixel bounding box in grayscale with two convolutional layers. Inputs are generated by fitting a bounding box around each object identified in a sonar image and resizing the bounding box into 40x40 pixels. 
We utilize Monte-Carlo dropout in this CNN to reject outliers and uncertain classifications by making $m$ predictions for each object. This way, we can assess the network's confidence in the predictions, as shown in \cite{Loquercio-2020}. Uncertain predictions are simply provided with the label ``unknown class." This pipeline in action is shown in Fig. \ref{fig:object_mapping_flow}.

To train this CNN, a small hand-annotated data set of representative sonar imagery is used, which is not included in the sequences used for validation in this work. For training the simulation model we use 200 samples per class and for real world data we use 300 samples per class. Generating sufficient training samples to train the model properly requires data augmentation. We augment our data by applying Gaussian noise, random flips, and random rotations. Further augmentation is not employed as satisfactory performance is achieved with the above methods. 

\subsection{Bayesian Inference for Objects Observed in 3D}
\label{bayes_update}
Each detected object in the horizontal sonar image is now represented by a cluster of pixels with a class label. These pixels have a range, bearing, and unknown elevation angle. At this step, the dual sonar fusion system provides an elevation angle for a subset of these pixels, which lie inside the small region with overlapping fields of view. These are the pixels we concern ourselves with in this subsection. 

We assume objects of the same class will have similar geometries, as is typical in the humanmade littoral environments, populated with piers, used to validate this algorithm.
Semantic classes are defined with this goal in mind, so that objects with similar geometries are grouped together.  

We use a Bayesian inference framework to estimate the conditional 
distribution P($z_Z^h | z_R^h, z_\theta^h$) for each object class incrementally and online. Note that in this process, we estimate Cartesian $z_Z$ and not elevation angle. 
$z_Z$ is a more accurate indicator of the absolute, rather than relative, height at which an object is observed, since in this work we consider scenarios in which our robot maps the environment at a fixed depth, employing planar SLAM.
An object's distribution is updated for every measured 3D point per Bayes rule:
\begin{linenomath*}
    \begin{align}
P({z_Z^h} | {z_R^h}, {z_\theta^h}) = \dfrac{P(  {z_R^h}, {z_\theta^h} | {z_Z^h})  P({z_Z^h})}{P(  {z_R^h}, {z_\theta^h})}.
\label{eq:bayes rule}
\end{align}
\end{linenomath*}
Elevation angles measured by the dual sonar fusion system are treated as measurements of ${z_Z^h}$ at the given range and bearing, corrupted with zero-mean Gaussian noise, $\mathcal{N}(\mu,\,\sigma^{2})$, forming the measurement likelihood $P({z_R^h}, {z_\theta^h} | {z_Z^h})$. $z_Z^h$ is different from $z_Z$ in that it is specifically modeling the value from the perspective of the horizontal sonar. The prior, ${P(z_Z^h})$, is simply the existing distribution corresponding to the ${z_R^h}$ and ${z_\theta^h}$ of the newly observed 3D point. In practice, these distributions are maintained as a discrete set of probabilities across the whole elevation bound. We note that these distributions are maintained throughout the whole time-history of the robot's mission, so they incorporate observations from the current frame and from all previous frames. An initial uniform distribution is used if an update has never been performed previously. 

\begin{table*}[h]
\vspace{-0mm}
\centering
\begin{tabular}{cccc}
\toprule
System & Semantic Model & Repeating Objects for High Coverage & Short Term Dead Reckoning \\
\midrule
Sonar Fusion Mapping    & x & x & x \\
Inference Based Mapping & \checkmark & \checkmark & x \\
Submapping              & x & x & \checkmark \\
\toprule
\end{tabular}
\caption{\textbf{Comparison of system requirements.} We compare our three mapping systems, each with its operational requirements. A check denotes the requirement of a category, while an x means it is not necessarily required. }
\label{table:system_reqs}
\end{table*}

At times we may view an object class at different distances and orientations; for this reason, we register each object to a \textit{reference coordinate frame} before we apply Bayes rule in Eq. \eqref{eq:bayes rule}. The first time we see an object, we note its minimum range and median bearing as the reference frame's origin. These object points are then maintained as a ``reference point cloud" to register the object class's future instances to this coordinate frame. When an object is detected, and its distribution P(${z_Z^h} | {z_R^h}, {z_\theta^h}$) is updated, the object is first registered to the given object class's reference coordinate frame using ICP. The transformed points are evaluated via Eq. \eqref{eq:bayes rule} and are added to the points used to register future object sightings to the reference frame. Our object-specific distributions P(${z_Z^h} | {z_R^h}, {z_\theta^h}$) will next allow us to predict the height of the sonar returns observed only in 2D.

\subsection{Predicting 3D Structure via MAP Estimation}
At each time step, after the process detailed in Section \ref{bayes_update} is completed for the objects measured in 3D, we again consider all objects with class labels comprised of a minimum number of pixels. We now use the posterior distribution of each object's geometry, $P({z_Z^h} | {z_R^h}, {z_\theta^h})$, to predict the height of all 2D points lacking this information. 

Suppose an object belongs to a class with a posterior updated by at least one application of Eq. \eqref{eq:bayes rule}. In that case, we first proceed with registration to the object class's reference frame as described above without adding new points to the reference point cloud. Due to the sonar's ambiguity, there may be more than one true ${z_Z^h}$ for a given range and bearing. For this reason, we break maximum a posteriori (MAP) estimation into two steps, as shown in Eqs. \eqref{eq:query}, \eqref{eq:query2}.
\begin{linenomath*}
\begin{align}
{z_Z^h}  = {argmax}  P({z_Z^h} | {z_R^h}, {z_\theta^h}), {z_Z^h} \leq 0 \label{eq:query}\\
{z_Z^h}  = {argmax}  P({z_Z^h} | {z_R^h}, {z_\theta^h}), {z_Z^h} > 0
\label{eq:query2}
\end{align}
\end{linenomath*}

If one or both maxima correspond to confidence exceeding a designated threshold, those values are adopted for inclusion in the robot's map. 
Eq. \eqref{eq:tx_to_cartesian} is solved to provide an output in local Cartesian coordinates, $[X,Y,Z]^T$. This process is completed for all objects in view -- the result is a horizontal sonar image with more observations fully defined in 3D, rather than just the few observations inside the region of dual-sonar overlap. The observations are converted to a point cloud and registered to the global map frame per Eq. \eqref{eq:map}.

\subsection{Inference Based Mapping}
\label{infer_map}
A 3D point cloud has been recovered at this stage in the inference-based mapping system. In exactly the same way as described in Section \ref{fusion_mapping}, we can only add to an aggregate 3D map at the discrete timesteps in the SLAM solution. This new inferred 3D point cloud is retained at the given keyframe and registered to the fixed frame $\Ical$ per Equation \eqref{eq:map}. 

\section{Submapping With Orthogonal Concurrent Wide Aperture Sonar Images}
\label{submapping-main}
This section will describe our newest contribution to 3D mapping using orthogonal sonar images. As noted in Sections \ref{fusion_mapping} and \ref{infer_map}, we only add to the 3D map from the SLAM keyframe poses. These SLAM keyframes are only generated when enough distance or rotation has accumulated to warrant adding a new keyframe. However, sonar imagery is available at 5 Hz, leaving most of the sonar image pairs unused in the above 3D mapping systems. This motivates our newest iteration on this system, building \textit{submaps}. In this subsystem, we will retain the sonar imagery between keyframes to build denser, more detailed 3D maps without requiring the prior information used by our inference based mapping system. We note this key difference from our other versions of the system: submapping retains the data at and between the keyframes, rather than \textit{only} at keyframes, as in the methods presented in Sections \ref{fusion_mapping} and \ref{infer_map}.

\subsection{Submap Construction}
\label{submapping}
Constructing submaps is a simple process using the dead reckoning pose $o_t$ and the fused sonar images $z^{Fused}$. We note that keyframes are discrete steps in time, denoted by iterator $k$. Once a keyframe is added, we log all the $z^{Fused}$ and their poses relative to the most recent keyframe. A submap at step $k$, $\mathcal{S}_k$, is defined as the set of observations between keyframes registered in the local reference frame: 
\begin{linenomath*}
    \begin{align}
    \mathcal{S}_k = \{      \Tbf_t^0 \hat{z}_0^{Fused} , 
                            \Tbf_t^{1} \hat{z}_{1}^{Fused},
                            \Tbf_t^{2} \hat{z}_{2}^{Fused}, 
                            ...,
                            \Tbf_t^N \hat{z}_N^{Fused}  \}.
\end{align}
\end{linenomath*}
Note that the submap $\mathcal{S}_k$ contains the $N$ fused sonar observations,  $\hat{z}_i^{Fused}$, where $N$ is the number of observations until the next keyframe is instantiated. Each of the $N$ observations, $\hat{z}_i^{Fused}$, has an associated transformation, $\Tbf_t^i$, that registers $\hat{z}_i^{Fused}$ into the keyframe's local reference frame at step $k$, derived from the dead reckoning system providing position. When considering time synchronization issues, linear interpolation is used to solve for a transform between dead reckoning steps, if required. The \textit{robot map}, $\mathcal{M}$ can be built from the submaps:
\begin{linenomath*}
\begin{align}
    \mathcal{M} = \{\Tbf_0 \mathcal{S}_0, 
                            \Tbf_1 \mathcal{S}_1, 
                            \Tbf_2 \mathcal{S}_2, 
                            ...,
                            \Tbf_t \mathcal{S}_t,   \}
\end{align}
\end{linenomath*}
Where $\Tbf_k$ is from the SLAM based pose estimate, $x$.
The key difference between submapping and the other mapping techniques in this paper is the collection and use of sonar data \textit{between} the discrete SLAM keyframes. Note, this system does require a source of accurate dead reckoning over the short term between keyframes. 

This sets up the tradeoffs to be examined in the Experiments section of this paper. Is a semantic labeling model available for the environment? Are repeating objects present? Does the vehicle have an accurate dead reckoning system? We will compare systems using these questions and their applicability in varying use cases.

\section{Experiments}
\label{experiments}
In this section, we will perform experimental validation and comparison of three versions of our mapping system. These systems will be referred to as below:
\begin{enumerate}
    \item \textbf{Sonar fusion mapping:} using only sonar image pairs at discrete timesteps (Section \ref{fusion_mapping})
    \item \textbf{Inference Based Mapping:} using only sonar image pairs at discrete timesteps and inference on those timesteps (Section \ref{infer_map})
    \item \textbf{Submapping:} using sonar image pairs \textit{at and between} timesteps (Section  \ref{submapping})
\end{enumerate}
We note that each method has its advantages, disadvantages, and requirements for operation. Sonar fusion mapping may provide inadequate detail due to its generally poor coverage. Inference based mapping improves on this but requires a trained model to segment sonar images. Moreover, inference based mapping's coverage rate is correlated with the number of simple, repeating objects in the environment, for example, circular pier pilings. Lastly, submapping will increase map coverage by retaining the data between timesteps but requires a dead reckoning system that is accurate between keyframes. A summary of operational requirements is shown in Table \ref{table:system_reqs}.

\subsection{Hardware Overview}
In order to perform real-world experiments and derive a simulation environment for this work, we use our customized BlueROV2-heavy robot, shown in Fig. \ref{fig:suny_img}. This vehicle is equipped with an onboard Pixhawk, Raspberry Pi, and NVIDIA Jetson Nano for control and computation. We use a Rowe  SeaPilot doppler velocity log (DVL), VectorNav VN100 inertial measurement unit (IMU), KVH DSP-1760 3-axis fiber optic gyroscope, and a Bar30 pressure sensor. We use a pair of wide aperture multi-beam imaging sonars for perceptual sensors, a Blueprint subsea Oculus M750d and M1200d. We use the M750d as our horizontal sonar and the M1200d as the vertical sonar. Note that the entirety of this work takes place with these sonars and their simulated versions operating at a range of 30 meters, with a 5cm range resolution. 

In order to manage the BlueROV's sensors, SLAM system, and companion 3D mapping system, we use the Robot Operating System \cite{Quigley-2009}, both for operating the vehicle and for playback of its data. The 3D mapping algorithms are applied to real-time playback of our data using a computer equipped with an NVIDIA Titan RTX GPU and Intel i9 CPU. Note that \textbf{all} experiments take place at a fixed depth. 

\subsection{Simulation Study}
\begin{figure*}[t]%
 \centering
 \subfloat[\textbf{Simulated Marina 1} ]{\includegraphics[height= 3cm]{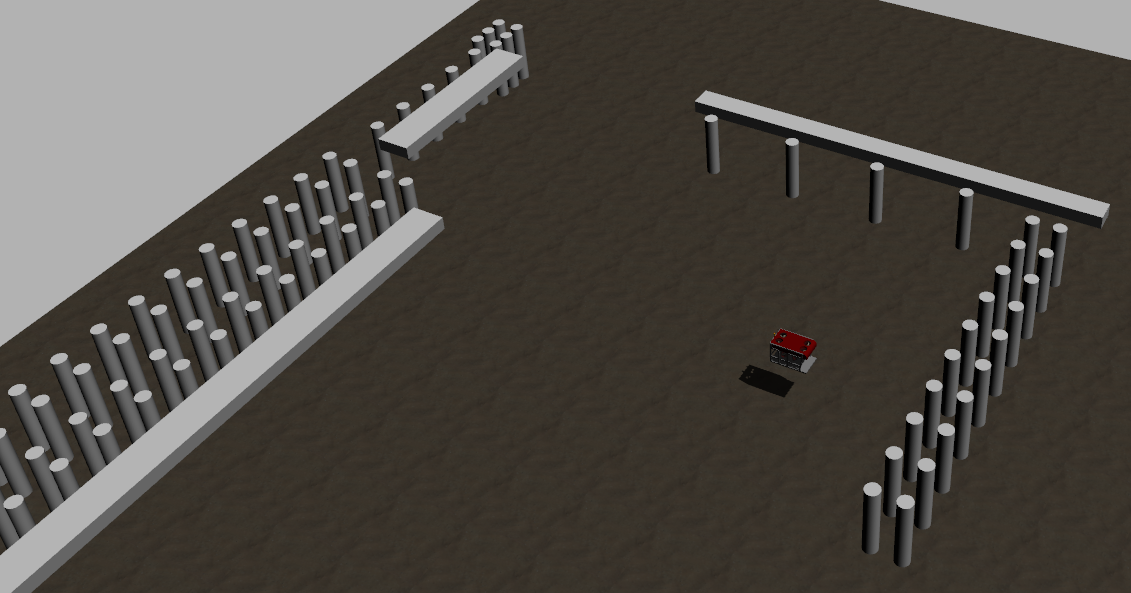}\label{fig:marina_1_pic}}%
 \subfloat[\textbf{Simulated Marina 2} ]{\includegraphics[height= 3cm]{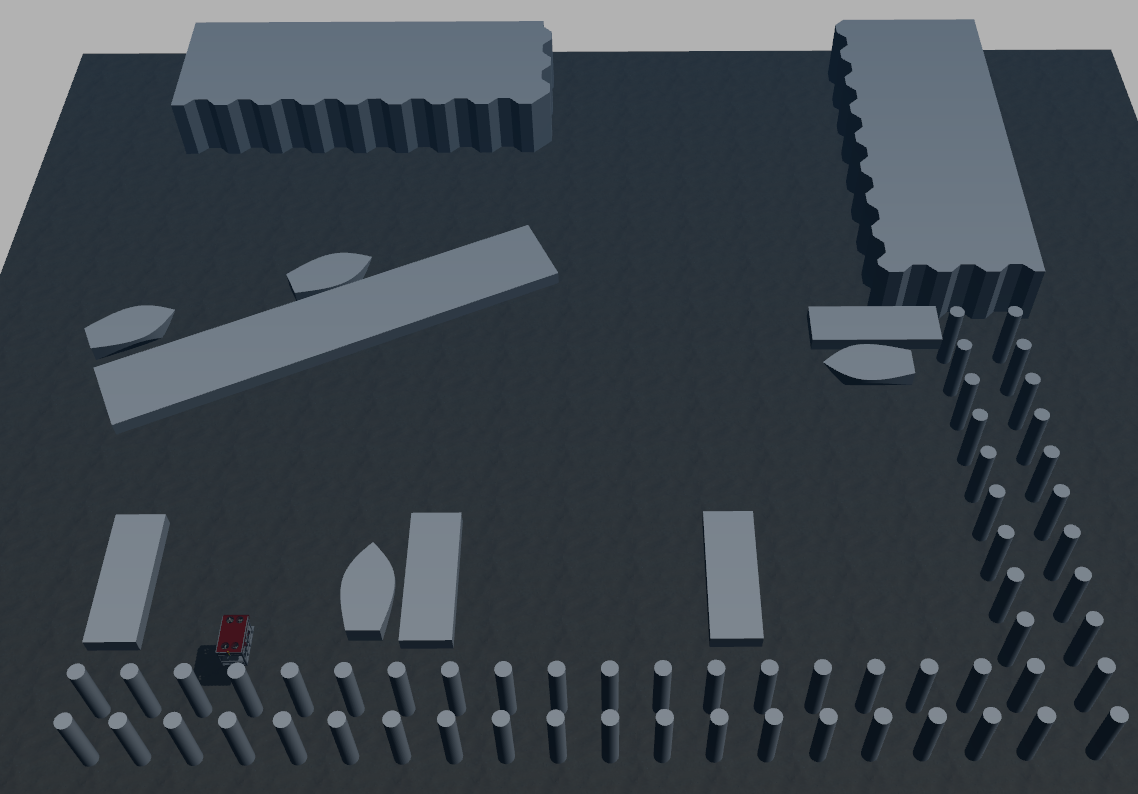}\label{fig:marina_2_pic}}%
  \subfloat[\textbf{Simulated Plane} ]{\includegraphics[height= 3cm]{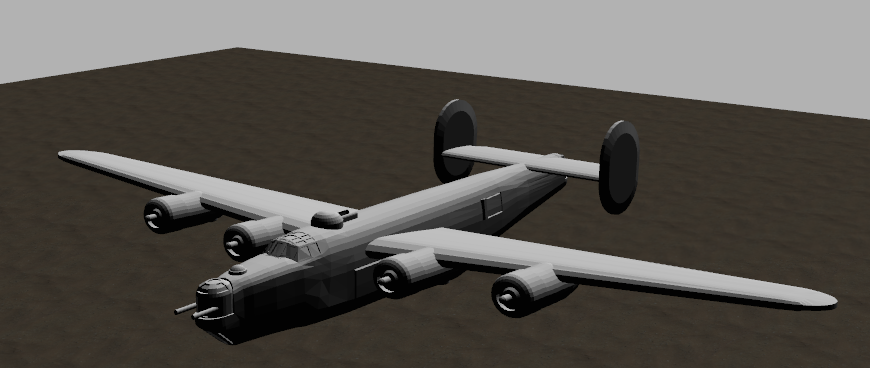}\label{fig:plane_pic}}\\
  \subfloat[\textbf{Simulated harbor}]{\includegraphics[width=0.7\linewidth]{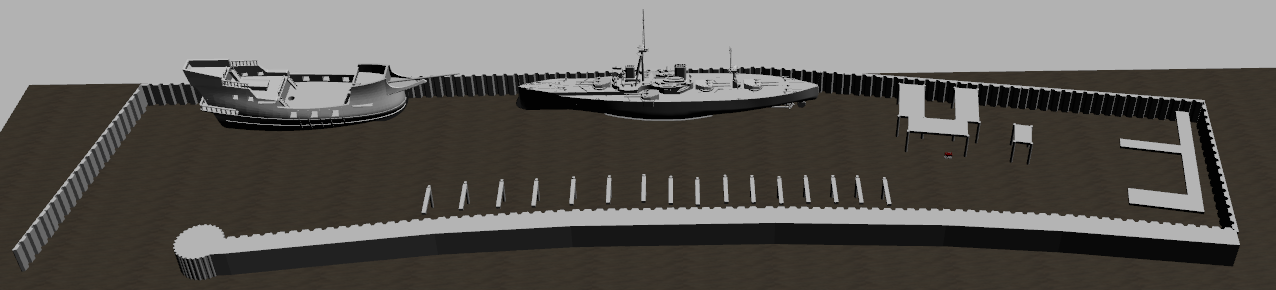}\label{fig:harbor_pic}}\\
\caption{\textbf{Simulation Environments.}}
\label{fig:simulation_pics}
\end{figure*}

In this section, we study our three mapping systems using simulated underwater environments. We utilize Gazebo \cite{gazebo} with UUV Simulator \cite{uuv-sim} to simulate the environment, vehicle, and sensors, including the wide aperture multi-beam imaging sonar \cite{sonar-sim}. Simulated environments are selected to capture variability in complex structures like ship hulls and aircraft and the number of repeating simple objects, such as round pier pilings. The four environments are described below:
\begin{enumerate}
    \item \textbf{Simulated Marina 1:} A marina environment consisting of floating docks and circular pier pilings. Note the heavy presence of repeating objects; the circular pier pilings. Designed to be similar in appearance to SUNY Maritime college's marina in The Bronx, NY. The simulation environment is shown in Figure \ref{fig:marina_1_pic}.
    \item \textbf{Simulated Marina 2:} A marina environment consisting of floating docks, small boats, circular pier pilings, and corrugated seawall. Again, note the heavy presence of repeating objects. Shown in Figure \ref{fig:marina_2_pic}.
    \item \textbf{Simulated Ships harbor:} A harbor environment with floating docks, a long corrugated seawall, and two large vessels. One large sailing ship and a WWI-era cruiser. This environment only has some repeating structures. Designed to be similar in appearance to Penn's Landing Marina in Philadelphia, PA. The simulation environment is shown in Figure \ref{fig:harbor_pic}.
    \item \textbf{Simulated Aircraft Site:} A single large aircraft on the seafloor. Note that there are no repeating structures. Moreover, this is the most complex geometry we have examined to date. The simulation environment is shown in Figure \ref{fig:plane_pic}.
\end{enumerate}
Since our systems utilize discrete SLAM keyframes in one way or another, which are added by distance/rotation threshold, we vary this distance and rotation, denoted as keyframe density. It is critical to vary keyframe density since this directly influences mapping coverage. In our work, we test combinations of keyframe distance and rotation. Keyframe distances of [1,2,3,4,5] meters and keyframe rotations of [30,60,90] degrees are examined. Note that we test all 15 combinations of distance and rotation.  

\begin{figure}[t]%
 \centering
 \subfloat[\textbf{Simulated Marina 1} ]{\includegraphics[width=0.68\columnwidth]{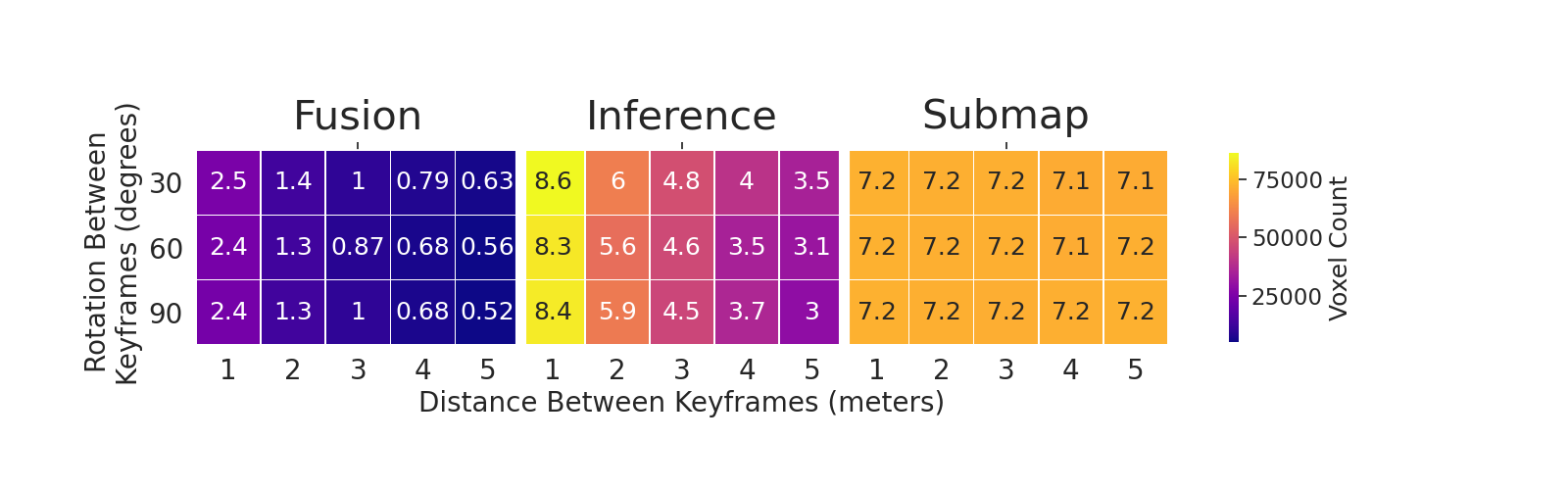}\label{fig:marina_1_heat}}\\
 \subfloat[\textbf{Simulated Marina 2} ]{\includegraphics[width=0.68\columnwidth]{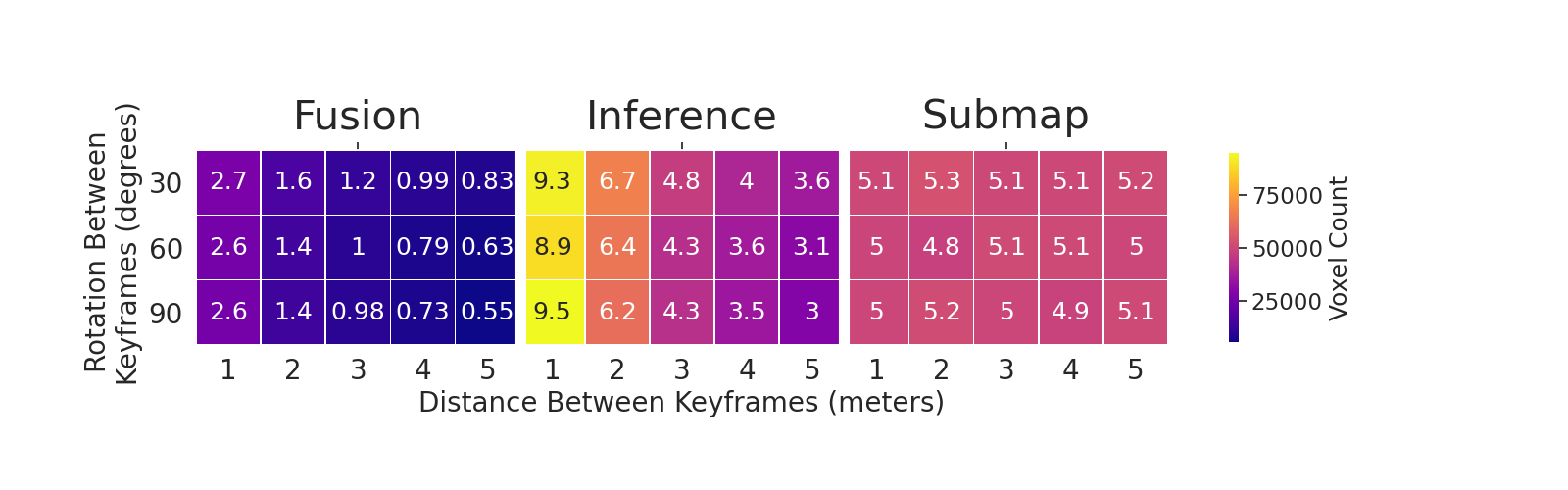}\label{fig:marina_2_heat}}\\
  \subfloat[\textbf{Simulated harbor}]{\includegraphics[width=0.68\columnwidth]{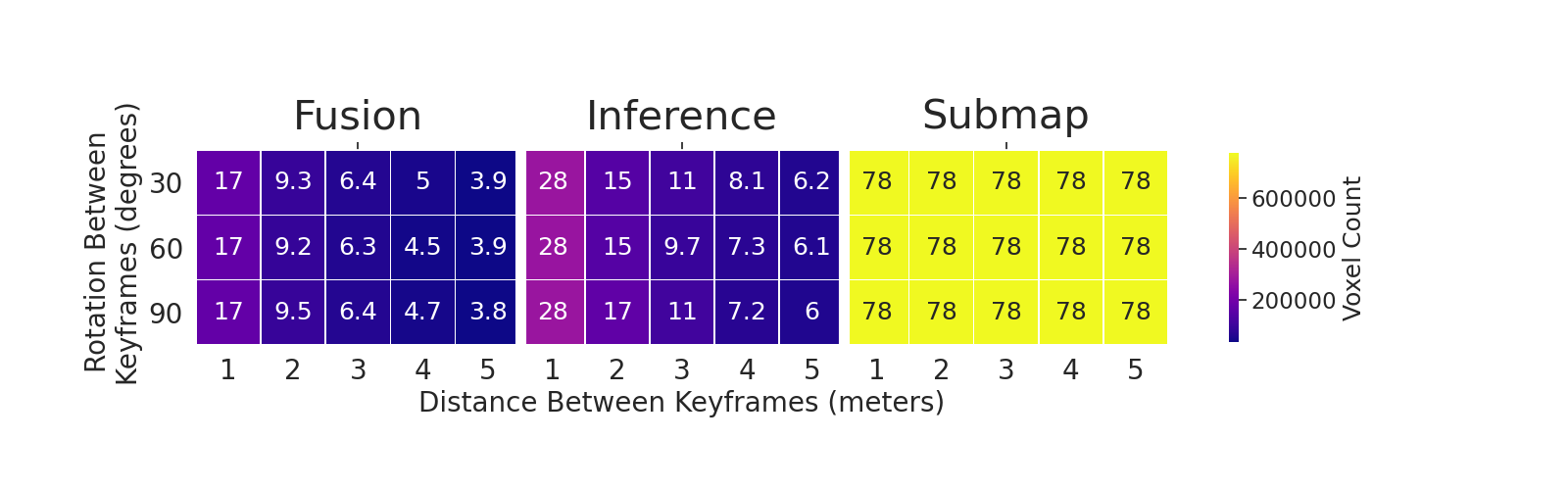}\label{fig:harbor_heat}}\\
 \subfloat[\textbf{Simulated Plane} ]{\includegraphics[width=0.68\columnwidth]{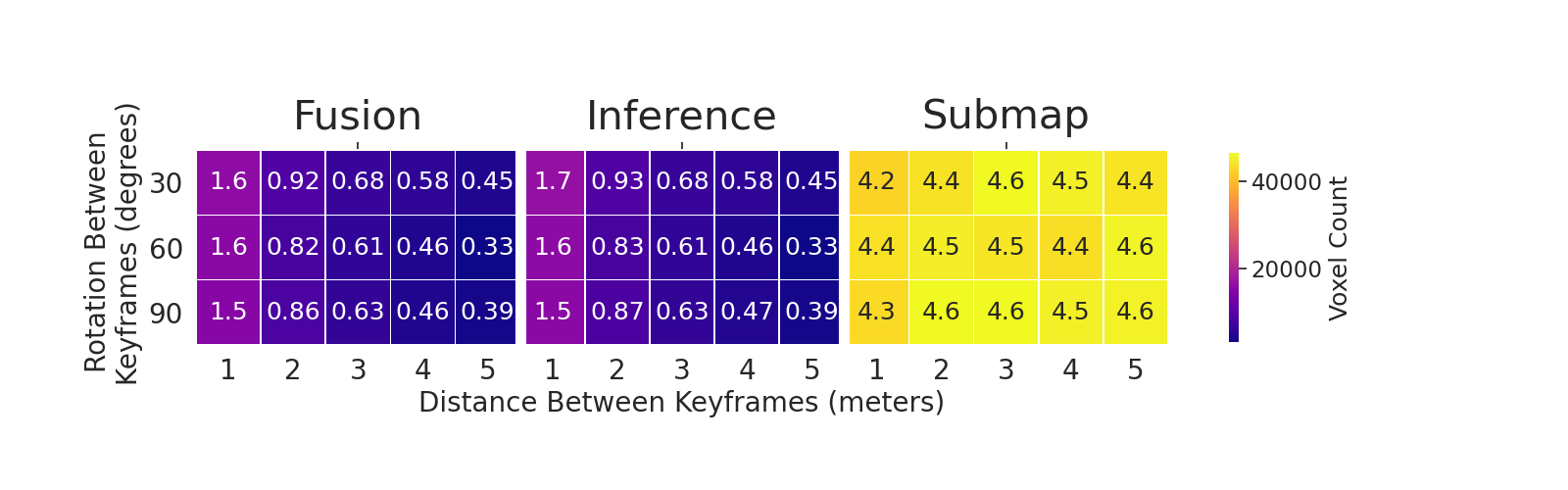}\label{fig:plane_heat}}\\
\caption{\textbf{Simulation coverage results.} Coverage in both table and color format. Each cell reports voxel count, with color mapping from blue to orange as low to high coverage. The vertical axis shows the varying keyframe rotations in degrees, with the horizontal axis showing the Euclidean distance between keyframes in meters. Each system type is shown here, with sonar fusion mapping at left, inference based mapping at center and submapping at right. Subfigure (a) shows the results for simulated marina 1, (b) shows the results for simulated marina 2, (c) shows the results for the simulated harbor and (d) shows the results for the simulated plane. Note, the voxel count is colored according to the scale bar given at the right of each subfigure.}
\end{figure}

To study the performance of each method, we consider several metrics. Firstly we define a coverage metric, the total number of voxels occupied when the point cloud map is discretized. We use voxel count as a coverage metric to avoid double-counting redundant data in the same location. Point cloud maps are voxelized using a 0.1m voxel size. Next, we can quantify point cloud map error using the environment CAD model. Here we consider the absolute distance between a point in the map and the CAD model. Accuracy is reported using mean-absolute-error (MAE) and root-mean-squared-error (RMSE) to characterize the error distribution's mean and size. 

\subsubsection{Simulated Marina 1}
Firstly we consider simulated marina 1. Recall that this environment contains floating docks and many repeating circular pier pilings, shown in Figure \ref{fig:marina_1_pic}.  Figure \ref{fig:marina_1_heat} shows a summary of coverage, measured in voxel count. Qualitative sonar fusion results are shown in Figure \ref{fig:marina_1_fusion_cloud}, inference based mapping results are shown in Figure \ref{fig:marina_1_infer_cloud}, and submapping is shown in Figure \ref{fig:marina_1_submap_cloud}. The trajectory followed by the robot is shown in Fig.
\ref{fig:marina_1_traj}. For sonar fusion mapping, we note the lowest coverage of the three methods, with coverage increasing as keyframe density increases, left and up in Figure \ref{fig:marina_1_heat}. When considering inference based mapping, the center of Figure \ref{fig:marina_1_heat}, we again note increased coverage with increased keyframe density and generally improved coverage compared to sonar fusion mapping, due to the prevalence of simple, repeating objects. However, we also note that inference based mapping has higher coverage than submapping when the distance between keyframes is 1 meter. When considering submapping, on the right of Figure \ref{fig:marina_1_heat}, coverage is generally flat across keyframe density, and coverage is higher than other methods, except when the inference based mapping system is applied in the densest keyframe category.

Again, we consider mapping error by comparing the point cloud map to the simulation environment CAD file. Mapping error is reported in Table \ref{error}. We note that mean mapping error is similar across all three methods in simulated marina 1, with submapping having a slightly smaller RMSE. 

\subsubsection{Simulated Marina 2} 
\begin{figure*}[t]%
 \centering

    \subfloat[\textbf{Marina 1 Sonar Fusion Mapping} ]{\includegraphics[height= 3cm]{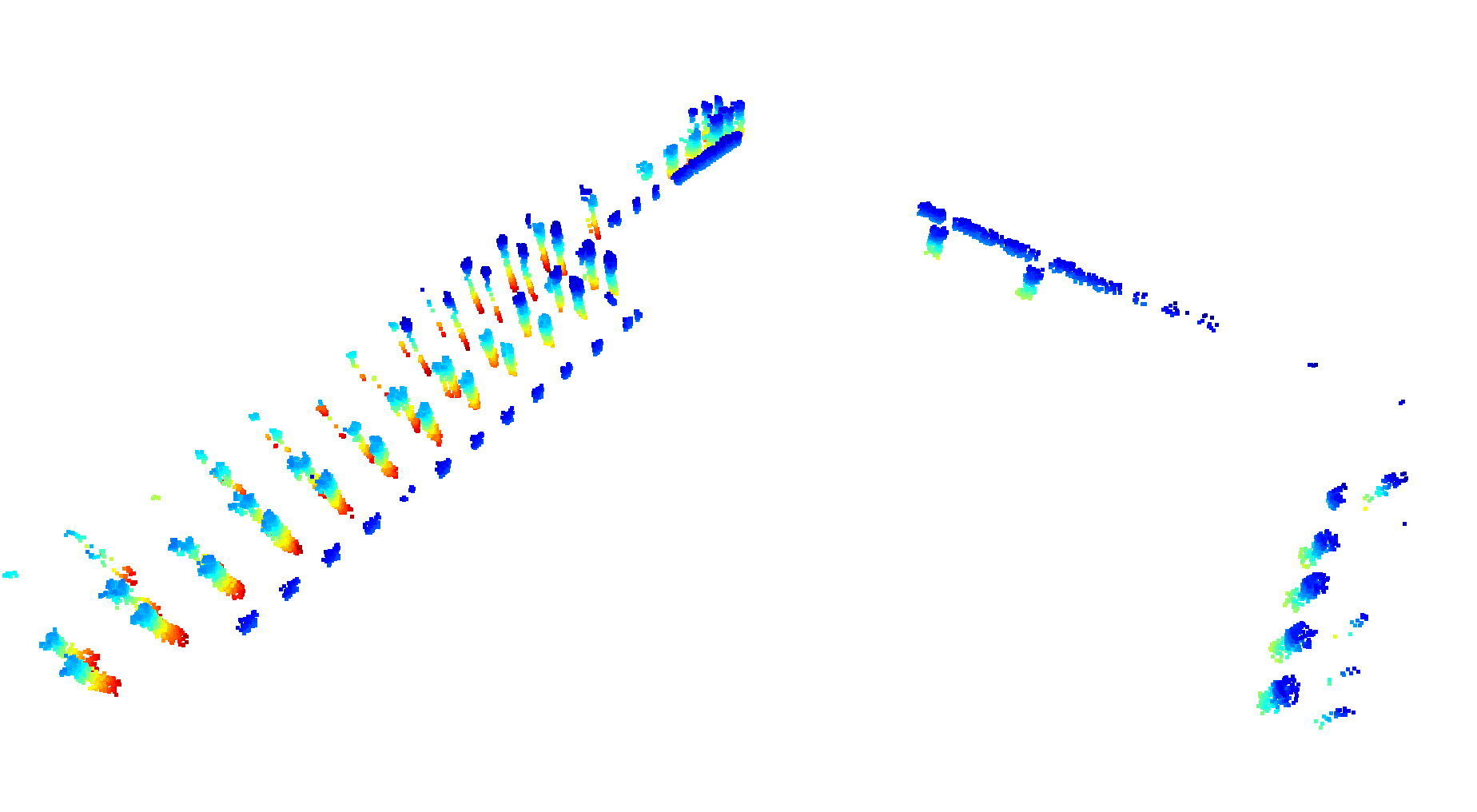}\label{fig:marina_1_fusion_cloud}}%
 \subfloat[\textbf{Marina 1 Inference Based Mapping} ]{\includegraphics[height= 3cm]{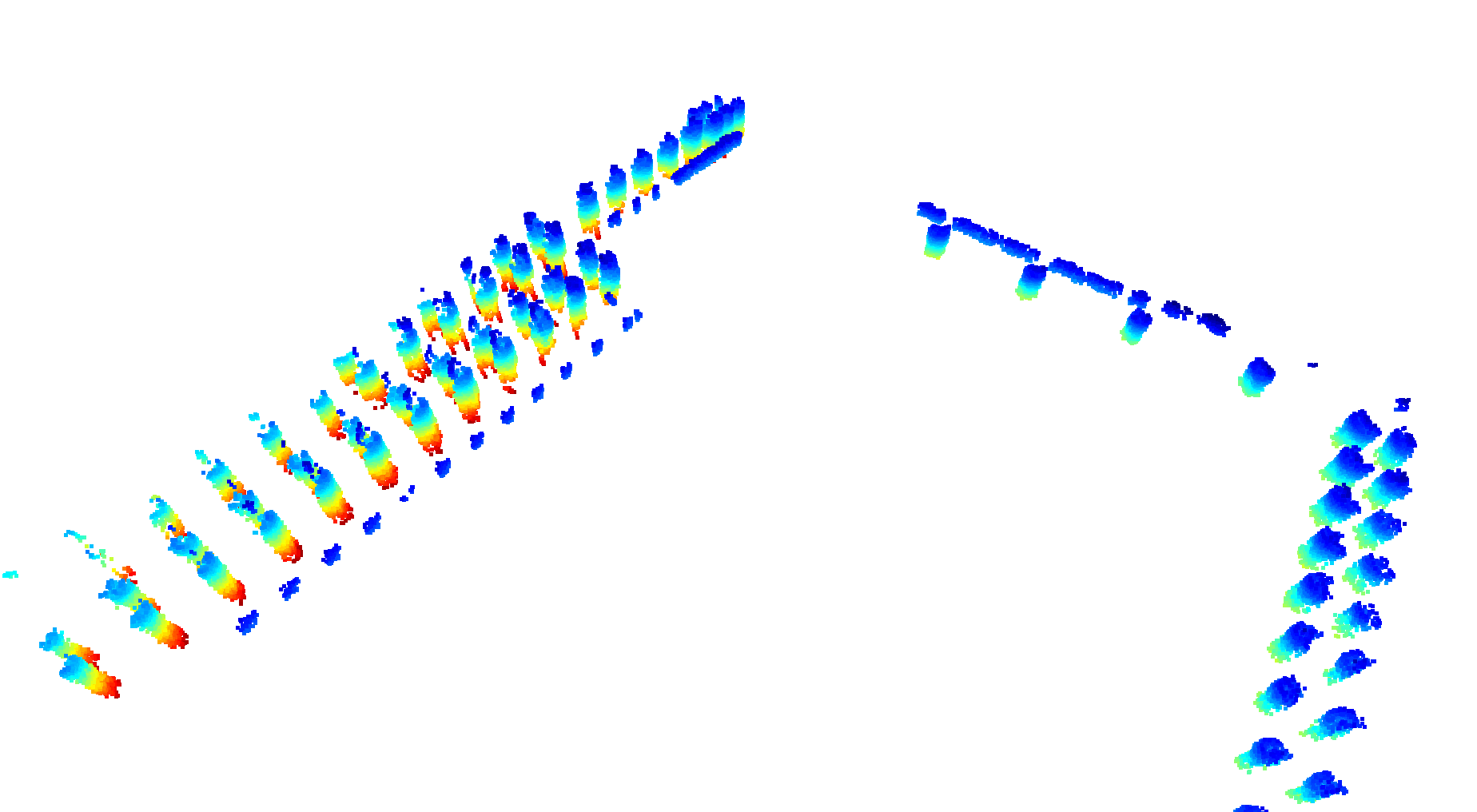}\label{fig:marina_1_infer_cloud}}%
  \subfloat[\textbf{Marina 1 Submapping} ]{\includegraphics[height= 3cm]{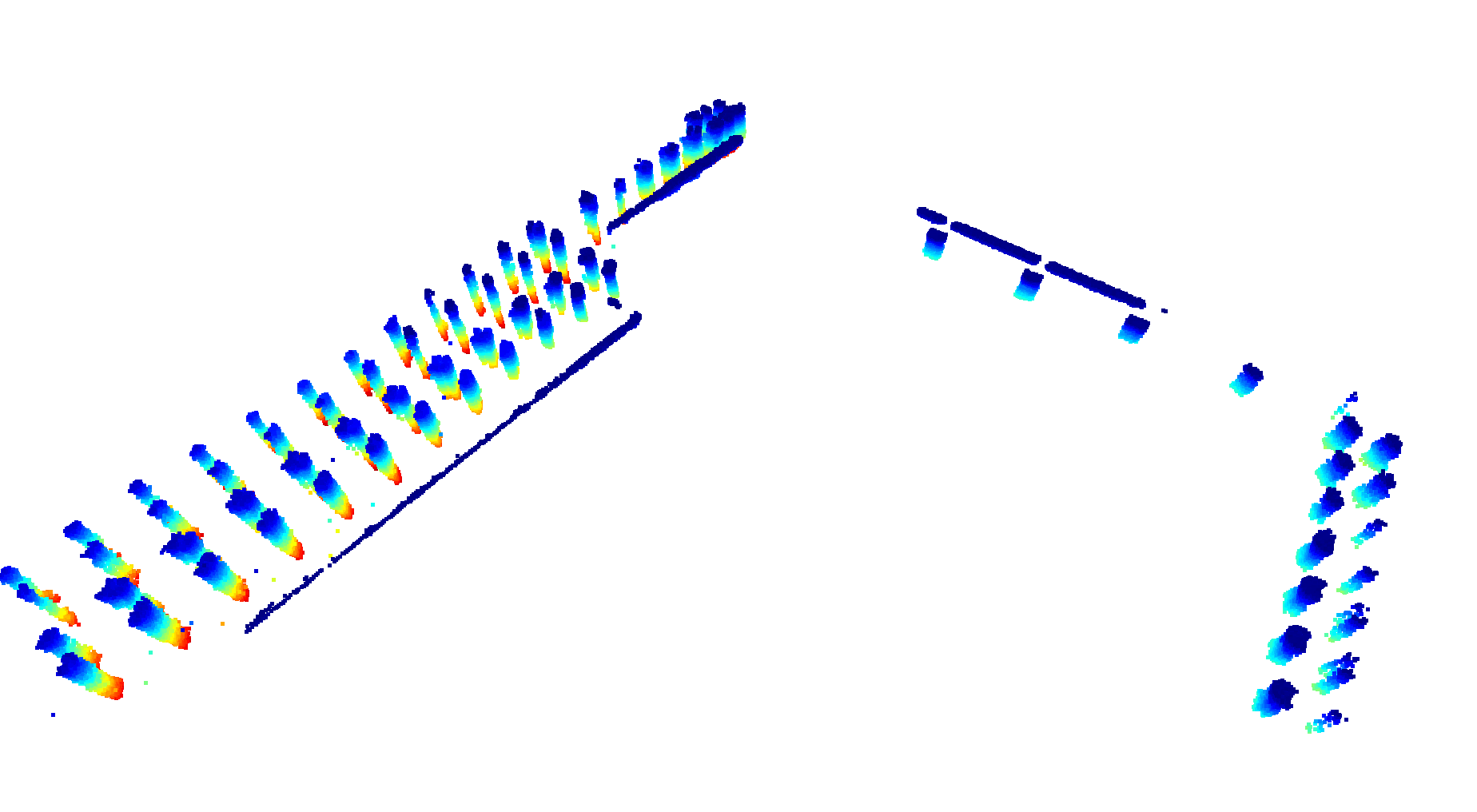}\label{fig:marina_1_submap_cloud}}\\

  \subfloat[\textbf{Marina 2 Sonar Fusion Mapping} ]{\includegraphics[height= 3cm]{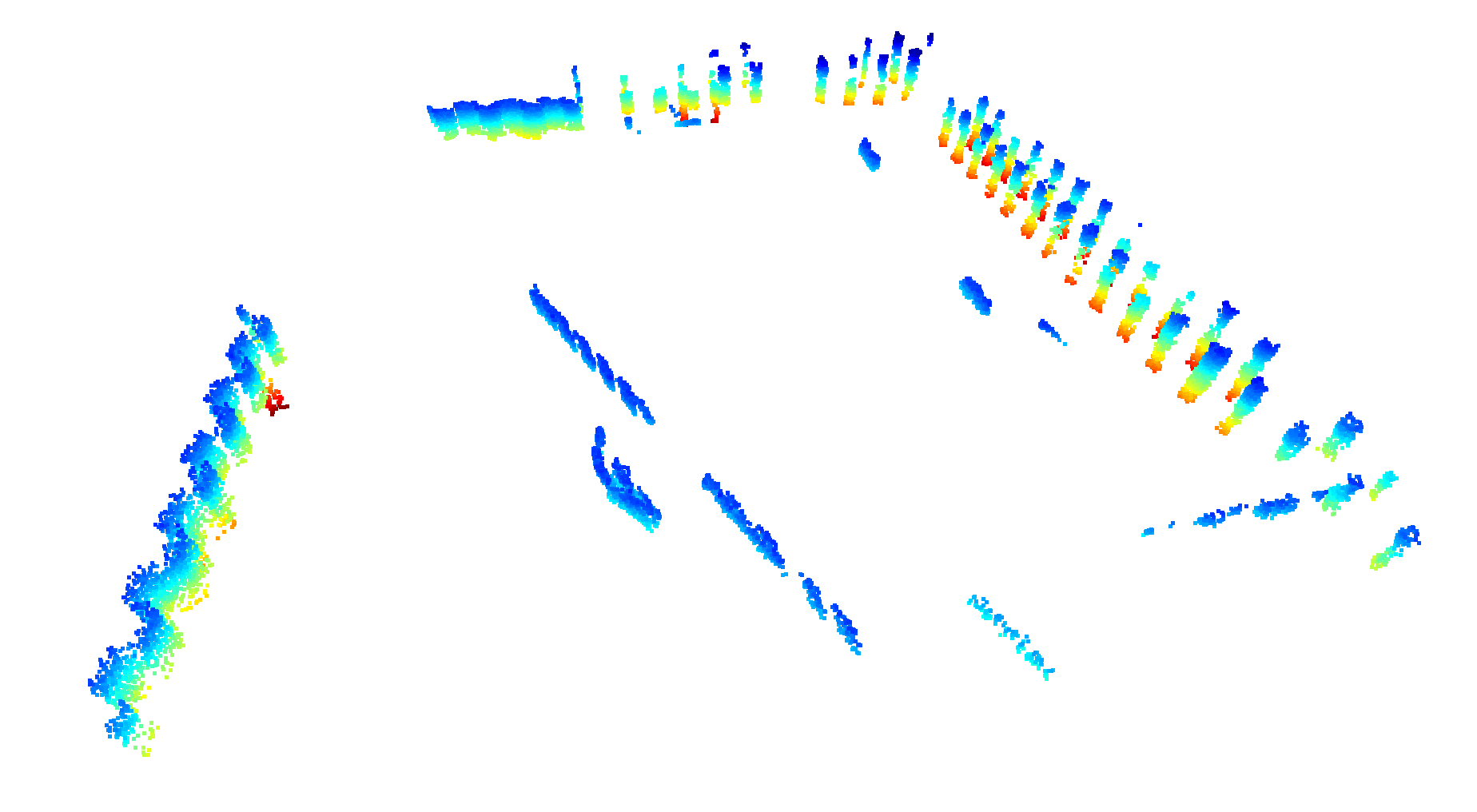}\label{fig:marina_2_fusion_cloud}}%
 \subfloat[\textbf{Marina 2 Inference Based Mapping} ]{\includegraphics[height= 3cm]{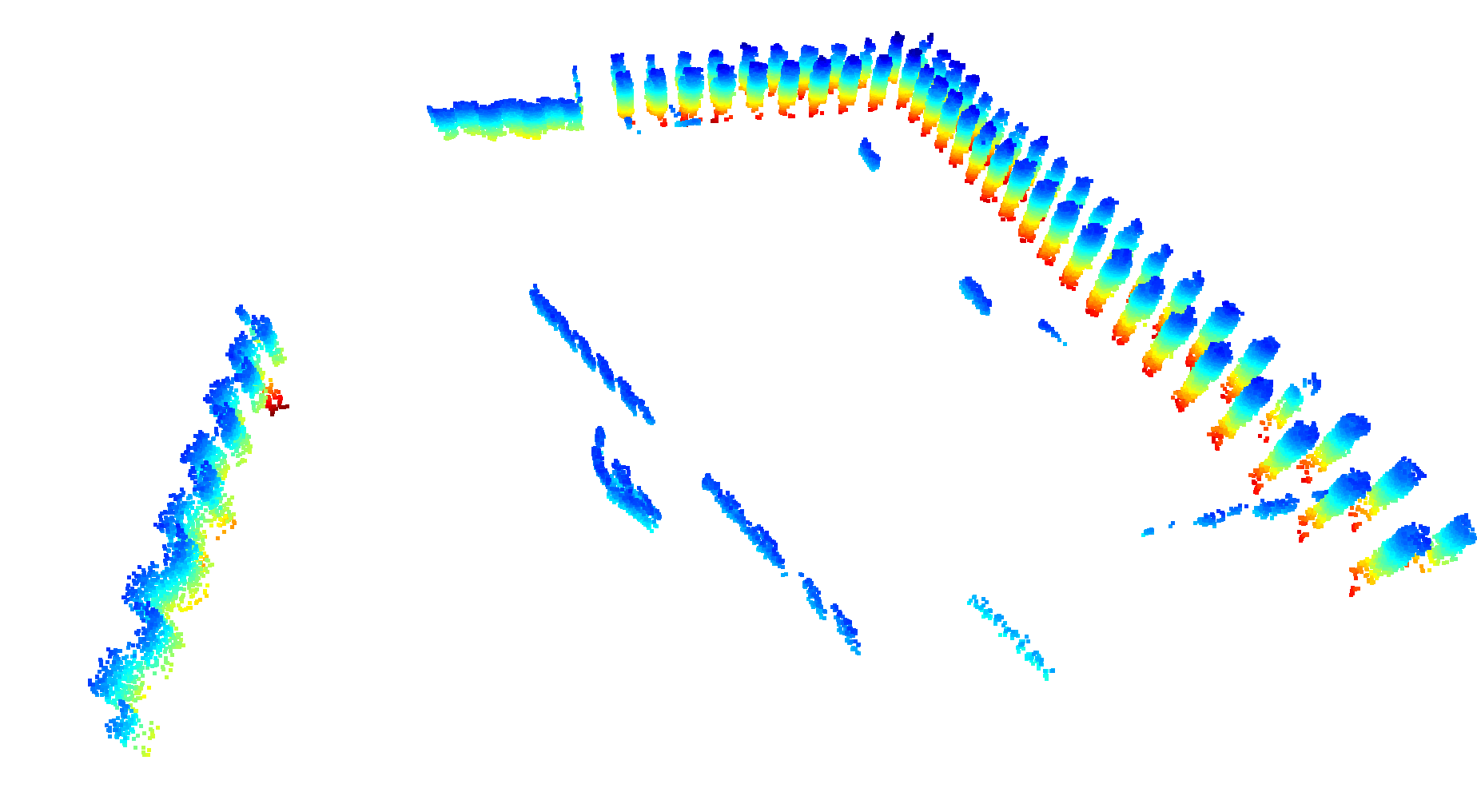}\label{fig:marina_2_infer_cloud}}%
  \subfloat[\textbf{Marina 2 Submapping} ]{\includegraphics[height= 3cm]{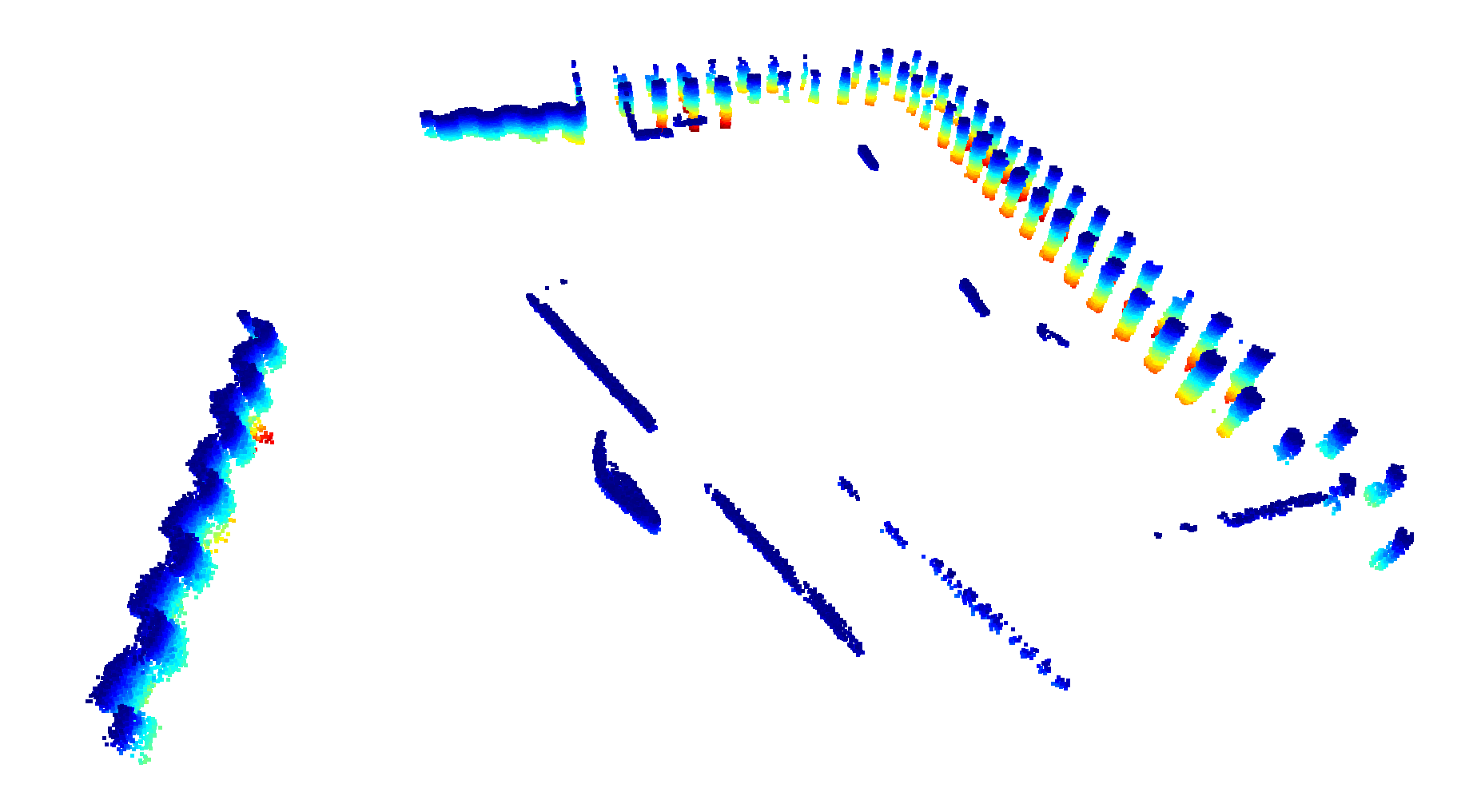}\label{fig:marina_2_submap_cloud}}\\
  
 \subfloat[\textbf{Plane Sonar Fusion Mapping} ]{\includegraphics[height= 3cm]{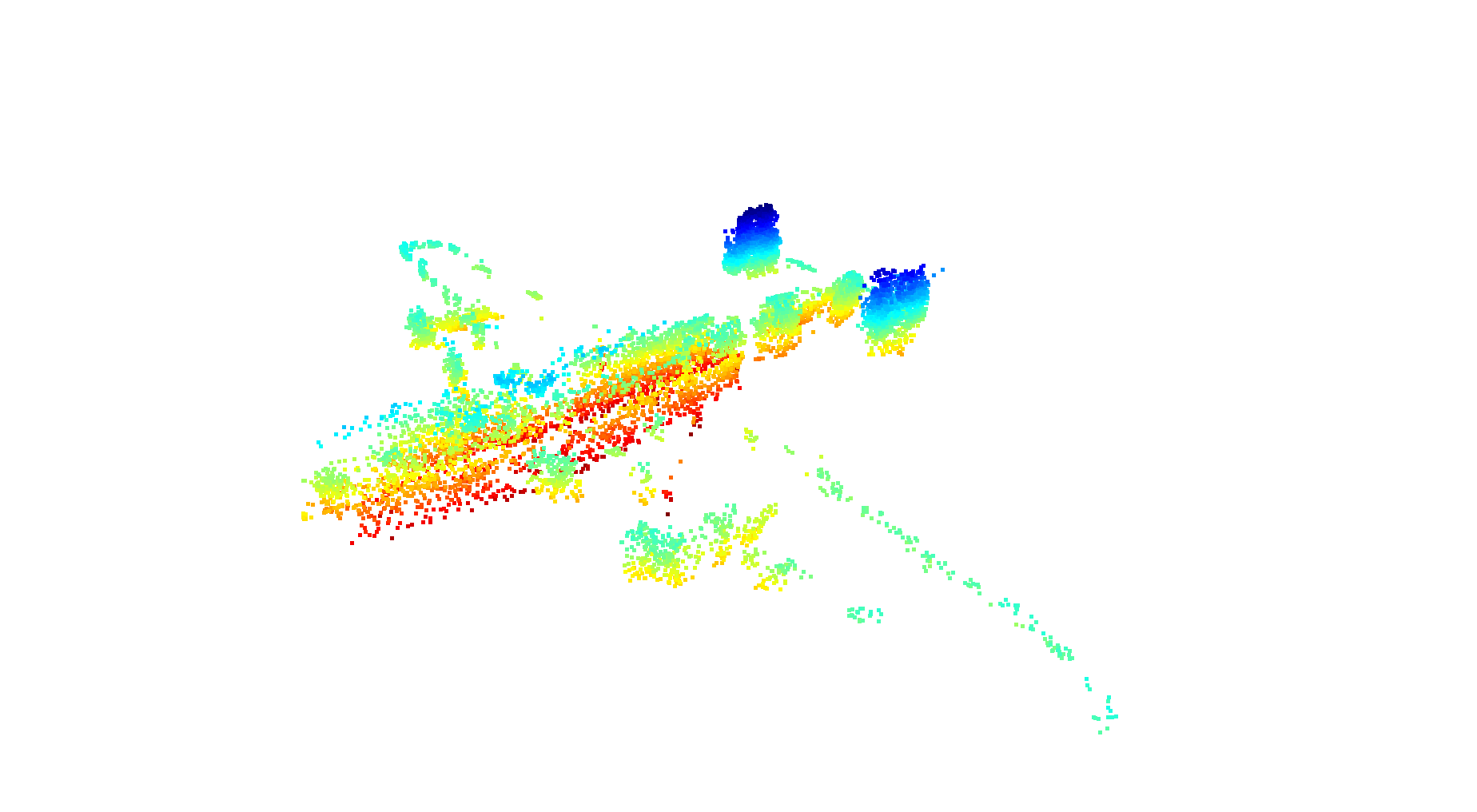}\label{fig:plane_fusion_cloud}}%
 \subfloat[\textbf{Plane Inference Based Mapping} ]{\includegraphics[height= 3cm]{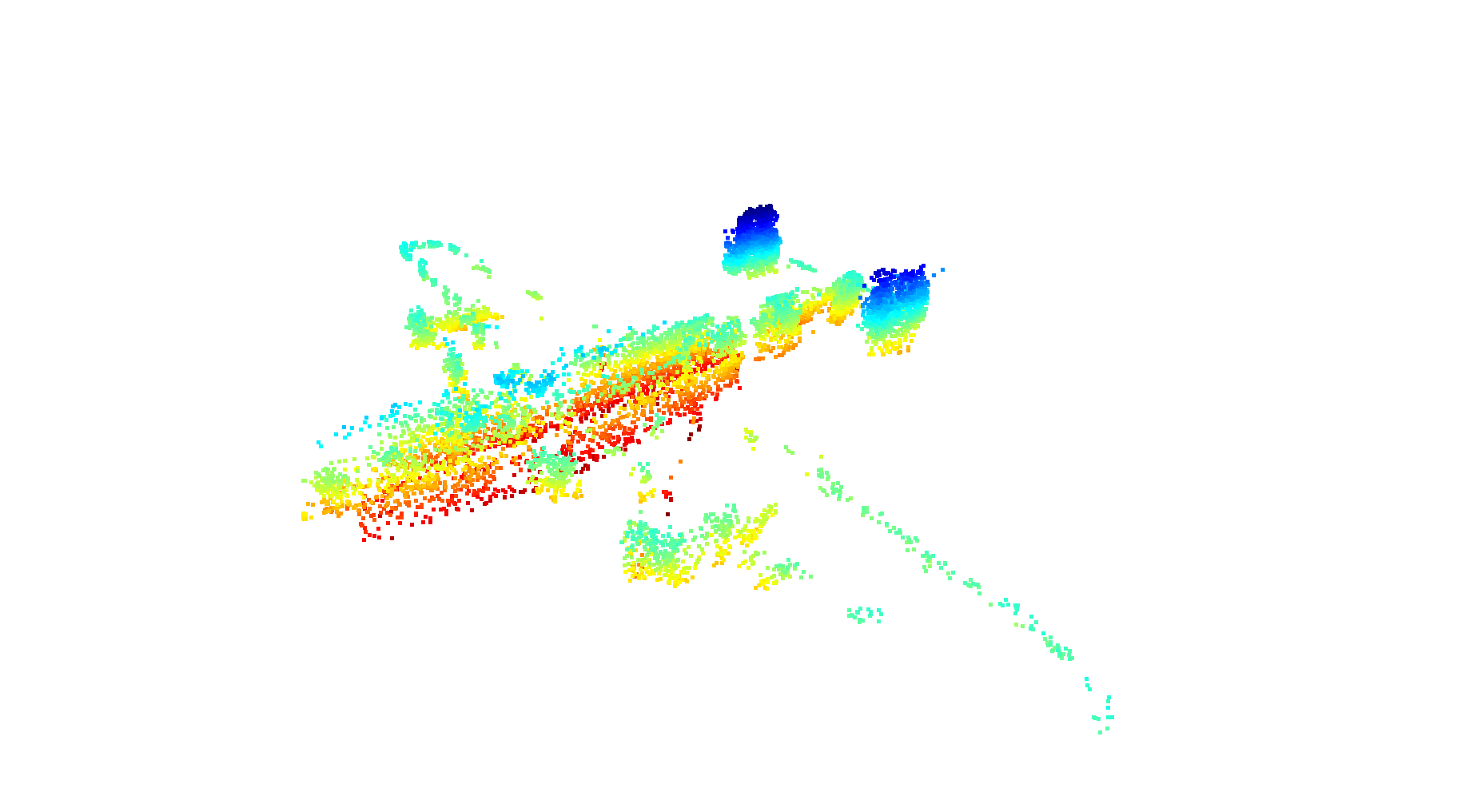}\label{fig:plane_infer_cloud}}%
  \subfloat[\textbf{Plane Submapping} ]{\includegraphics[height= 3cm]{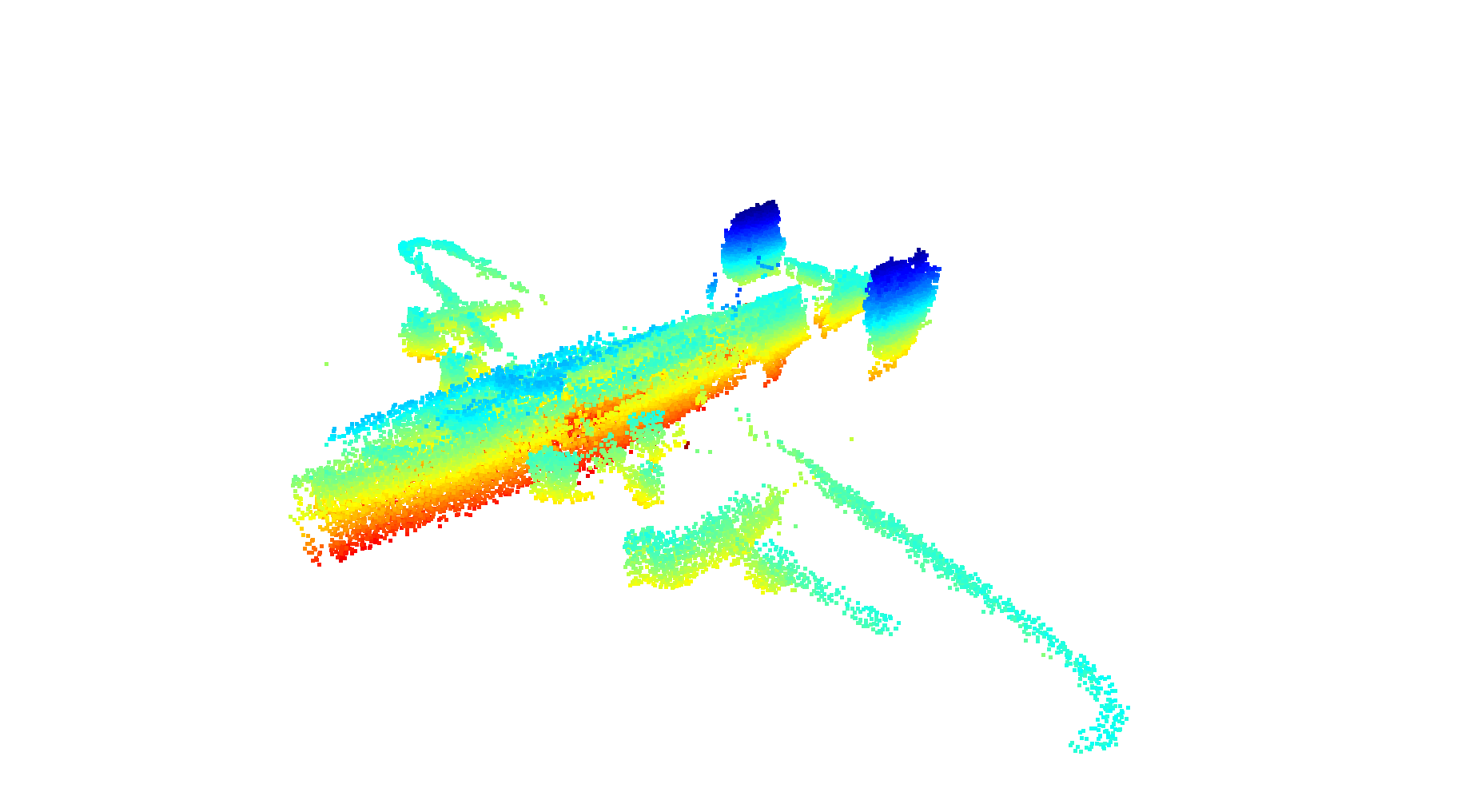}\label{fig:plane_submap_cloud}}\\

 \subfloat[\textbf{harbor Sonar Fusion Mapping} ]{\includegraphics[height= 3cm]{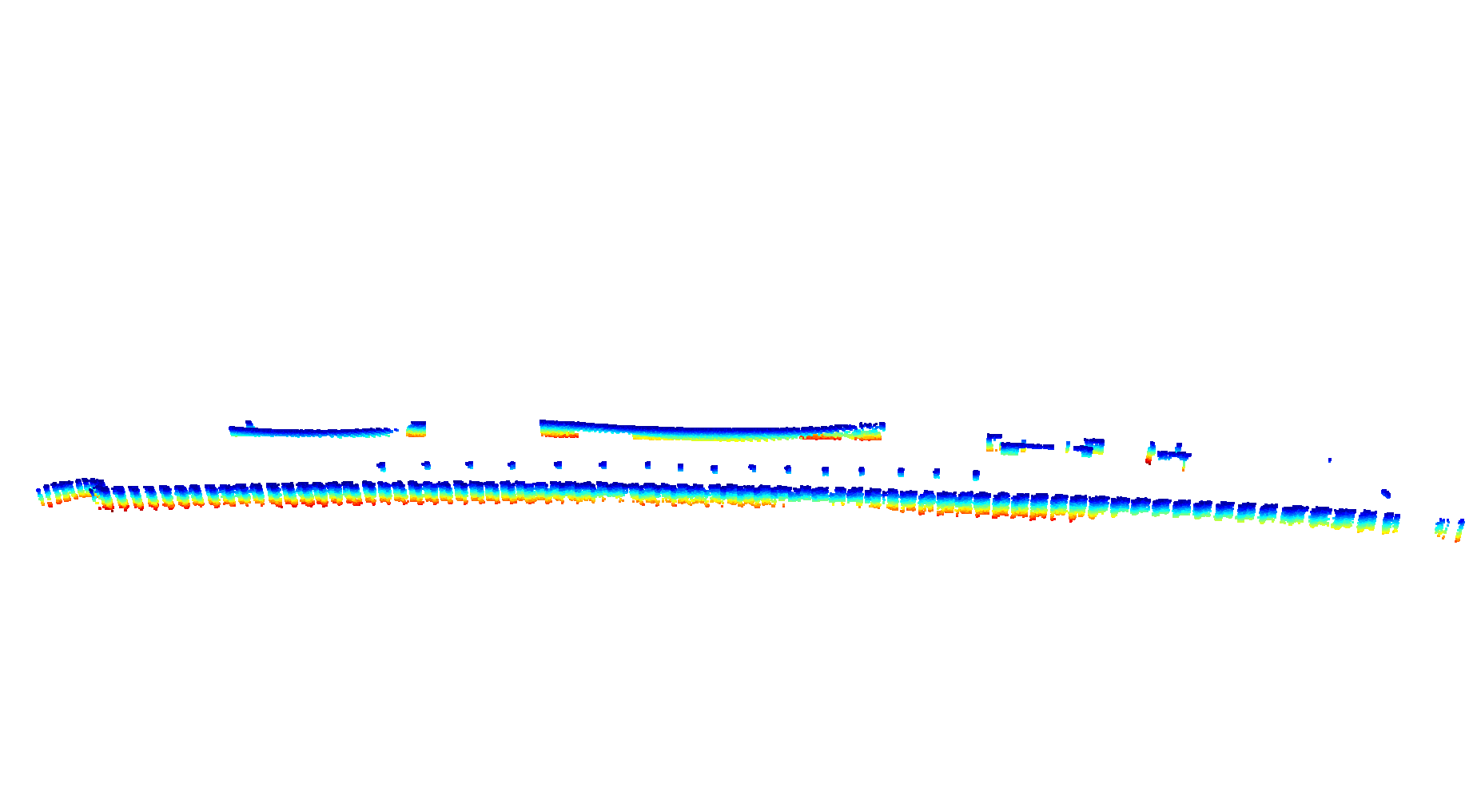}\label{fig:harbor_fusion_cloud}}%
 \subfloat[\textbf{harbor Inference Based Mapping} ]{\includegraphics[height= 3cm]{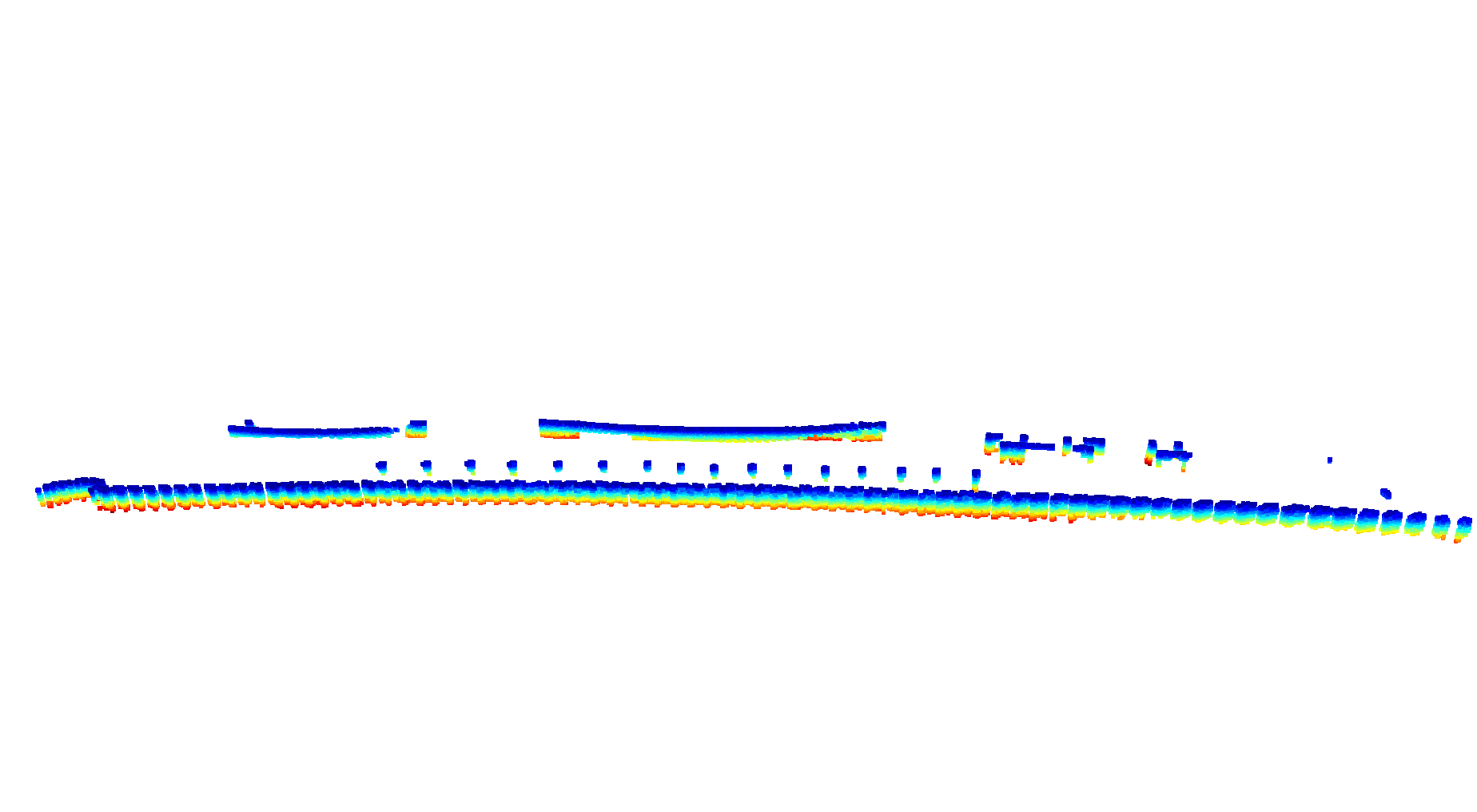}\label{fig:harbor_infer_cloud}}%
  \subfloat[\textbf{harbor Submapping} ]{\includegraphics[height= 3cm]{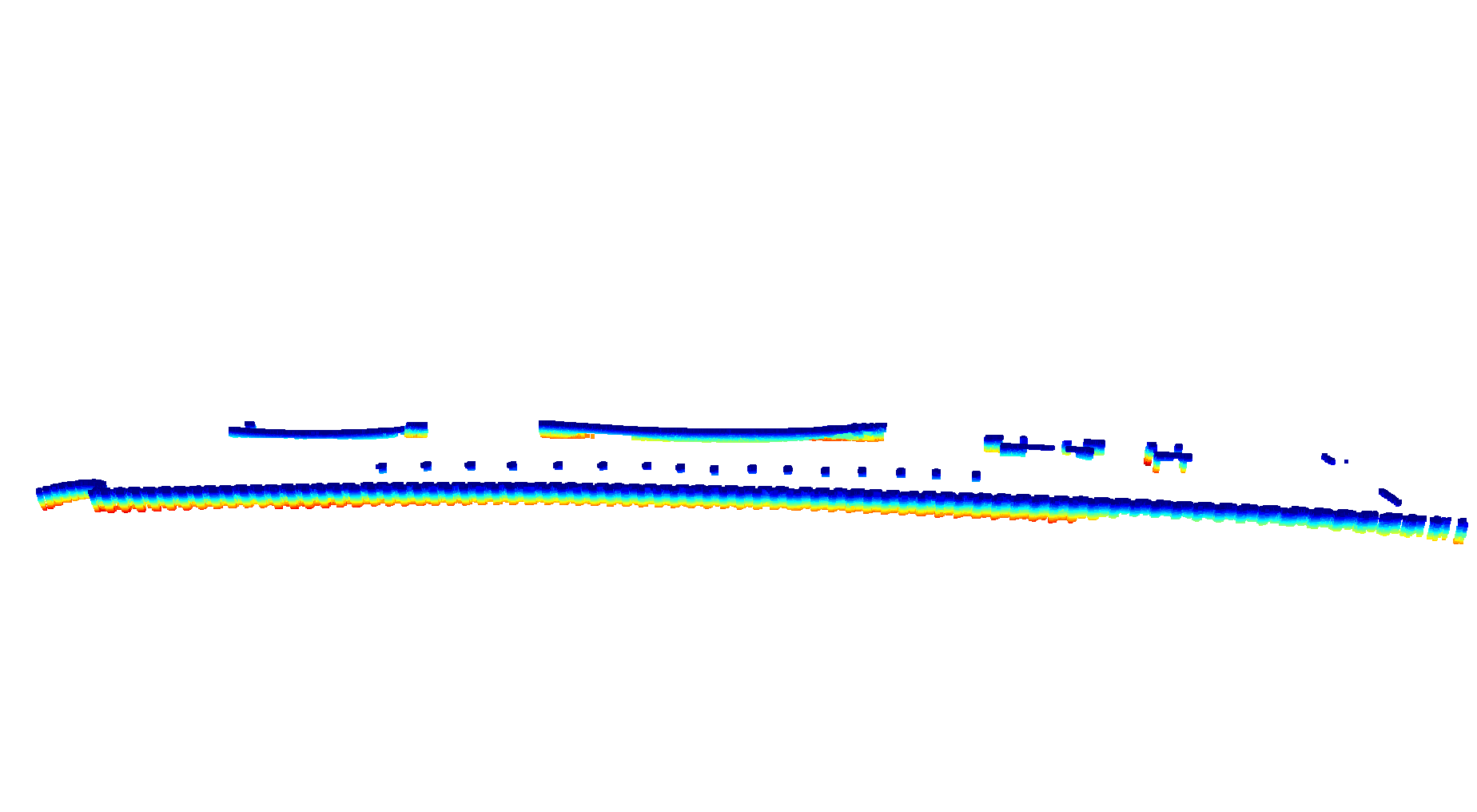}\label{fig:harbor_submap_cloud}}\\
  
\caption{\textbf{Simulation Environment Mapping Results.}}
\end{figure*}

Next, we consider simulated marina 2, shown in Figure \ref{fig:marina_1_pic}. Recall that this environment contains floating docks, small boats, many repeated circular pilings, and corrugated seawalls.   Coverage results for simulated marina 2 are shown in Figure \ref{fig:marina_2_heat}. Qualitative results are shown in Figures \ref{fig:marina_2_fusion_cloud}, \ref{fig:marina_2_infer_cloud} and \ref{fig:marina_2_submap_cloud}. The trajectory followed by the robot is shown in Fig. \ref{fig:marina_2_traj}. Again we see a similar trend; sonar fusion mapping shows the least coverage, with inference based mapping improving due to the repeating objects in the environment. However, inference based mapping only improves coverage over submapping when keyframes are two or less meters apart. Submapping again shows a generally flat performance in all keyframe densities. When considering mapping accuracy, we note comparable MAE, with slight RMSE change when moving to submapping. Simulated marina 2 accuracy results are shown in Table \ref{error}.

\subsubsection{Simulated Ships harbor} 
This simulated ship's harbor contains only a few repeating objects, floating docks, a long corrugated seawall, and two ships, shown in Figure \ref{fig:harbor_pic}.   Qualitative results are shown in Figures \ref{fig:harbor_fusion_cloud}, \ref{fig:harbor_infer_cloud} and \ref{fig:harbor_submap_cloud}. The trajectory followed by the robot is shown in Fig. \ref{fig:harbour_traj}. When considering the coverage in this environment, shown in Figure \ref{fig:harbor_heat}, we note a smaller coverage gap between sonar fusion mapping and inference-based mapping due to the reduced prevalence of simple repeating objects for the inference based mapping system to leverage. Additionally, we note that submapping outperforms inference based mapping in all keyframe density cases. The performance difference between submapping and inference based mapping is likely a combination to two key reasons—first, the lack of repeating objects in the environment. Second, the complex geometry of ship hulls is difficult to map using only keyframes; submapping uses the data \textit{at and between} the keyframes, enabling denser mapping of complex structures. When considering the simulated harbor environment error reported in Table \ref{error}  we note similar means, with submapping having a larger RMSE.

\subsubsection{Simulated Aircraft Site} 
\begin{figure*}[t]%
 \centering

    \subfloat[\textbf{Marina 1 Trajectory} ]{\includegraphics[height= 4cm]{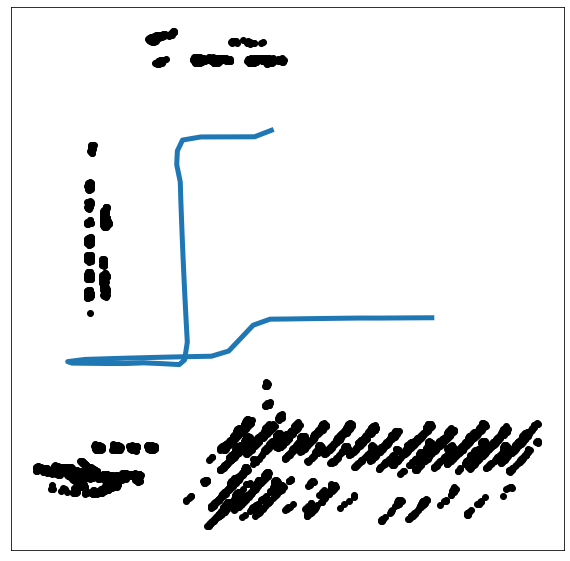}\label{fig:marina_1_traj}}
  \subfloat[\textbf{Marina 2 Trajectory} ]{\includegraphics[height= 4cm]{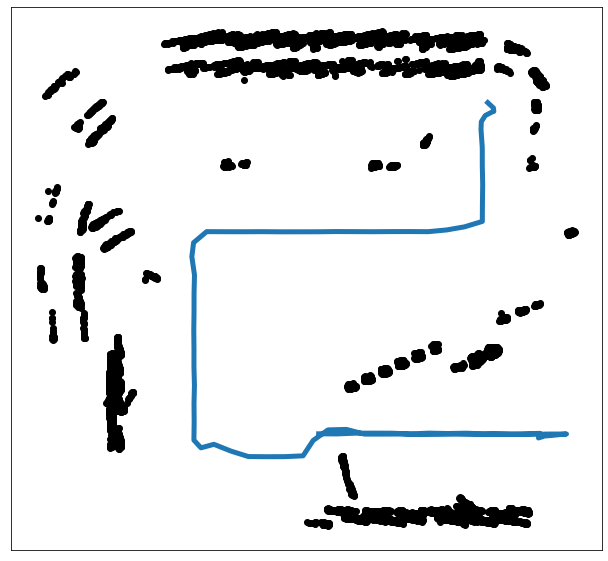}\label{fig:marina_2_traj}}
  \subfloat[\textbf{Plane Trajectory} ]{\includegraphics[height= 4cm]{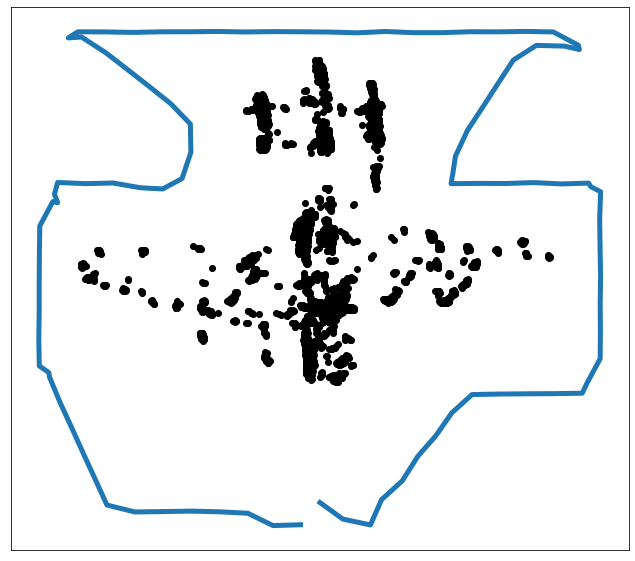}\label{fig:plane_traj}}\\

  \subfloat[\textbf{Harbor Trajectory} ]{\includegraphics[angle=90,height = 3cm]{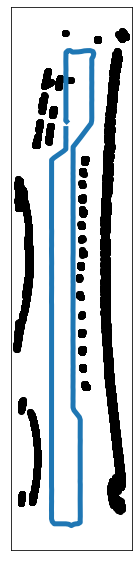}\label{fig:harbour_traj}}\\
  
\caption{\textbf{Simulation Environment Robot Trajectories.} Blue lines indicate the robot path, black points indicate the sonar observations.}
\label{fig:sim_trajectories}
\end{figure*}

This environment consists of one large structure, a WWII-era B-24 Liberator sitting on the seafloor, shown in Figure \ref{fig:plane_pic}. We use this structure as it presents highly complex geometry not previously analyzed with our mapping systems. Further, it does contain \textit{any} repeating objects, only one large structure, the aircraft. This specific aircraft is selected due to its distinctive dual tail arrangement, which, if mapped correctly, will be easily recognizable. Qualitative results are shown in Figures \ref{fig:plane_fusion_cloud}, \ref{fig:plane_infer_cloud} and \ref{fig:plane_submap_cloud}. Coverage results are shown in Figure \ref{fig:plane_heat}. The trajectory followed by the robot is shown in Fig. \ref{fig:plane_traj}. Due to the total lack of repeating objects in this environment, there is a negligible coverage difference comparing the inference-based mapping system to the sonar fusion map. When considering the submapping system, Figure \ref{fig:plane_heat} shows a coverage improvement compared to the other two systems. Most importantly though, when considering the qualitative results in Figure \ref{fig:plane_submap_cloud}, there is a more complete representation of the aircraft's main wing and tail section. Regarding map accuracy, shown in Table \ref{error}, we note similar means, with the submapping RMSE slightly larger. 

\begin{table*}[t]
\vspace{-0mm}
\centering
\begin{tabular}{ccccccccc}
\toprule
 & \multicolumn{7}{c}{Environment} \\
 & \multicolumn{2}{c}{Marina 1} & \multicolumn{2}{c}{Marina 2} & \multicolumn{2}{c}{Harbor} & \multicolumn{2}{c}{Plane}\\
\midrule
 System & MAE & RMSE & MAE & RMSE & MAE & RMSE & MAE & RMSE \\
 \midrule
Sonar Fusion Mapping     & 0.16 & 0.21 & 0.19 & 0.25 & 0.21 & 0.30 & 0.17 & 0.24 \\
Inference Based Mapping  & 0.16 & 0.20 & 0.17 & 0.22 & 0.22 & 0.30 & 0.17 & 0.24 \\
Submapping               & 0.14 & 0.17 & 0.16 & 0.19 & 0.26 & 0.37 & 0.21 & 0.28 \\
\toprule
\end{tabular}
\caption{\textbf{Simulated mapping accuracy in meters.} We report aggregated results for each system in each environment. MAE and RMSE are both reported in meters. }
\label{error}
\end{table*}

\begin{table*}[h]
\vspace{-0mm}
\centering
\begin{tabular}{ccccccccc}
\toprule
 & \multicolumn{7}{c}{Environment} \\
 & \multicolumn{2}{c}{Marina 1} & \multicolumn{2}{c}{Marina 2} & \multicolumn{2}{c}{Harbor} & \multicolumn{2}{c}{Plane}\\
\midrule
 System & Mean & SD & Mean & SD & Mean & SD & Mean & SD \\
 \midrule
Orthogonal Sensor Fusion  & 0.088 & 0.018 & 0.080 & 0.011 & 0.086 & 0.014 & 0.075 & 0.010 \\
Mapping Inference         & 0.617 & 0.267 & 0.529 & 0.387 & 0.277 & 0.172 & 0.233 & 0.193 \\
Submap Construction       & 0.005 & 0.009 & 0.001 & 0.003 & 0.006 & 0.010 & 0.001 & 0.003 \\
\toprule
\end{tabular}
\caption{\textbf{Runtime in seconds.} We report mean and standard deviation (SD) runtime for all experiments for a particular system in each environment. Orthogonal sensor fusion refers to the process of fusing a single sonar image pair. Mapping inference is the time required to apply the method in Section \ref{infer_map} to a single keyframe. Submap construction is the time required to assemble a submap from a single timestep once all the data is collected. }
\label{runtime}
\end{table*}

\subsubsection{Summary of Simulation Results}
The results of the simulation study can be summarized as follows. Firstly the sonar fusion mapping system provides accurate maps but potentially low coverage, especially when keyframe density is sparse. However, the sonar fusion mapping system has the fewest system requirements of the three compared methods, requiring no trained semantic model, no repeating objects to enhance coverage, and no short term dead reckoning system. Next, the inference based mapping system makes coverage improvements in three out of four environments compared to sonar fusion mapping. We note that this increased mapping coverage is due to the presence of simple repeating objects to perform object-specific inference using a trained semantic labeling model. Lastly, submapping only outperforms inference based mapping when there are few, if any, repeating objects in the environment. However, submapping recovers reasonably accurate, high-coverage maps in cases where the structures are complex. It is essential to note the value of submapping when potentially deploying an autonomous system to an unknown environment. Dense, 3D maps can be recovered with \textit{no prior information} regarding the environment. Moreover, in the case of complex 3D geometry, parts of the structure can be mapped that the other systems in this paper miss, such as the wing of the airplane shown in Figure \ref{fig:plane_submap_cloud}.

When considering runtime, reported in Table \ref{runtime}, we provide both mean and standard deviation (SD) in seconds. In the row marked ``orthogonal sensor fusion'' in Table \ref{runtime}, we provide the time required to fuse a single pair of sonar images, not build a map. We note that this runtime is sufficient to keep pace with the simulated sonar sensor, which runs at 5Hz (generating new images every 0.2 seconds). The row marked ``Mapping inference'' in Table \ref{runtime} records the time required to apply the inference based mapping system in section \ref{infer_map} to a single keyframe. We note that runtime for this category is slower when repeating objects are present. Moreover, a key detail some readers may note is that we do not apply the inference systems to data between keyframes. While the runtime is sufficient to keep up with the addition of keyframes, applying this method to the $N$ sonar image pairs between keyframes would be impractical. Lastly, when considering the runtime of the ``submap construction'' row in Table \ref{runtime}, this is the required time to construct a submap once all the data is collected between keyframes. We note that this process is low in runtime and does not add undue computation to our system.

\subsection{Real World Study}
\begin{figure*}[t]%
 \centering
 \subfloat[]{\includegraphics[height=5cm]{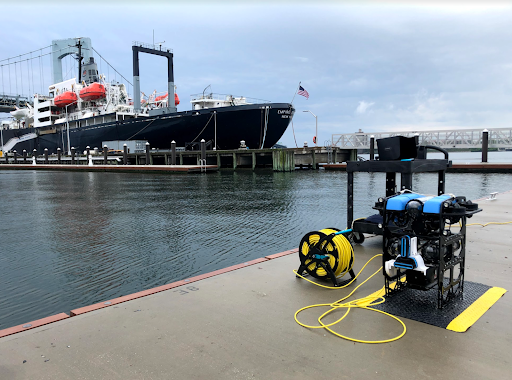}\label{fig:suny_img}}%
 \subfloat[]{\includegraphics[height=5cm]{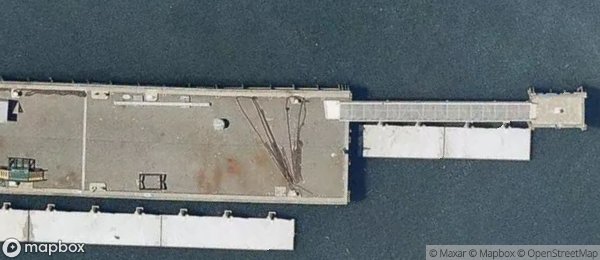}\label{fig:suny_sat}}\\
  \subfloat[]{\includegraphics[height=3.5cm]{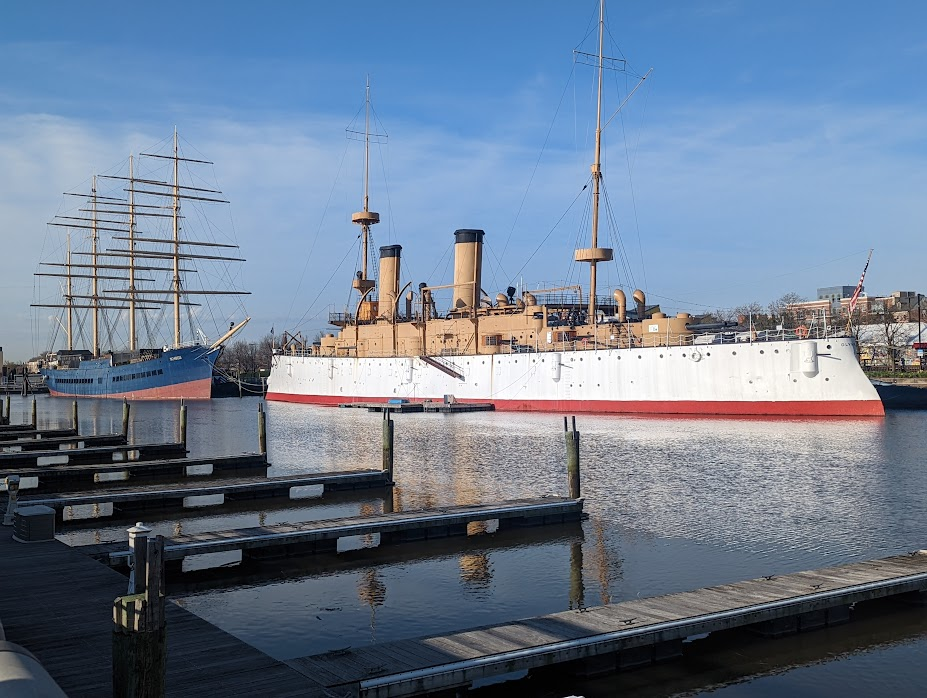}\label{fig:penn_img}}%
 \subfloat[]{\includegraphics[height=3.5cm]{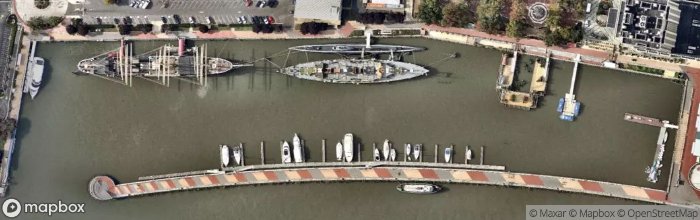}\label{fig:penn_sat}}\\
   \subfloat[]{\includegraphics[height=3.5cm]{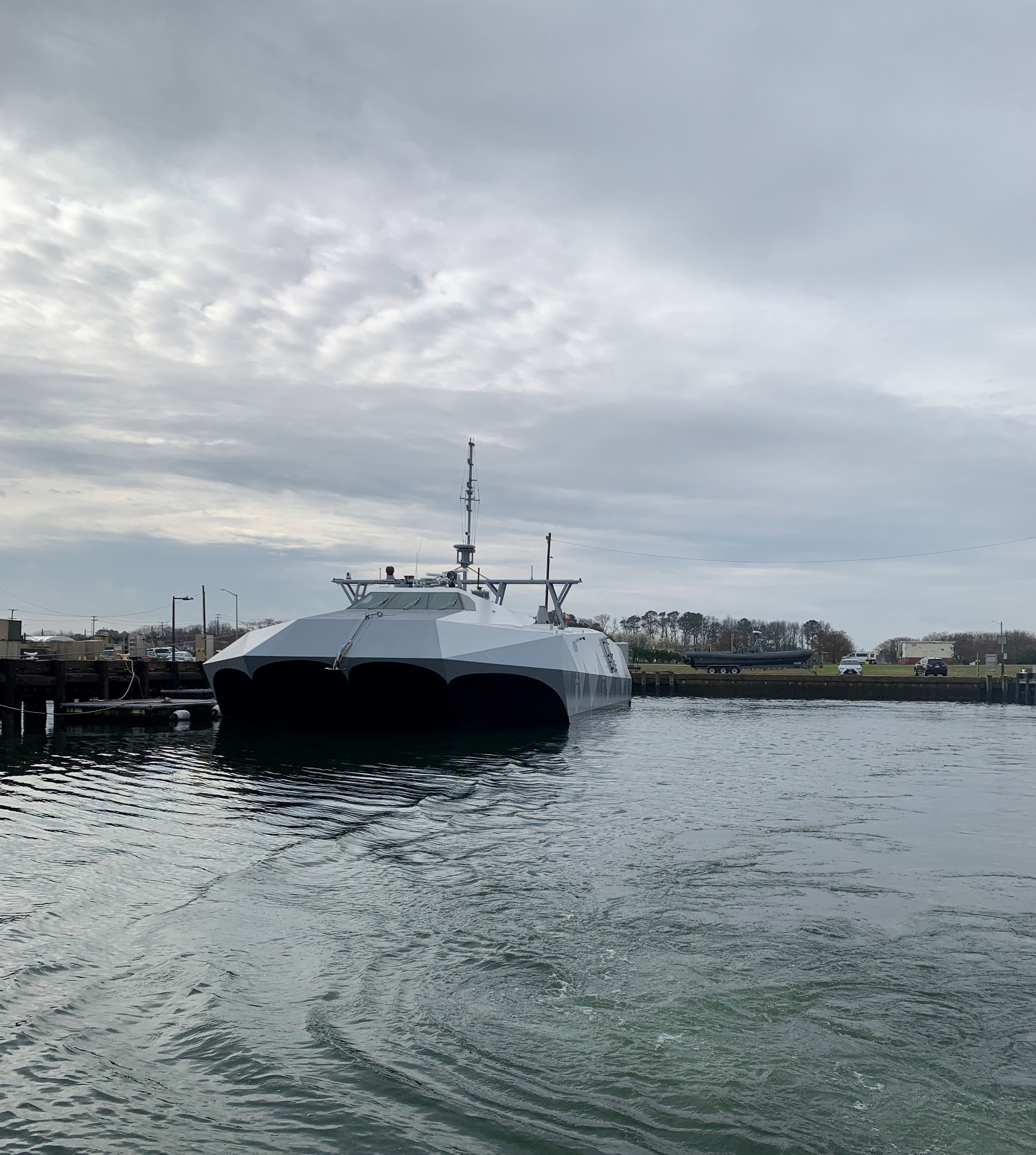}\label{fig:norfolk_img}}%
 \subfloat[]{\includegraphics[height=3.5cm]{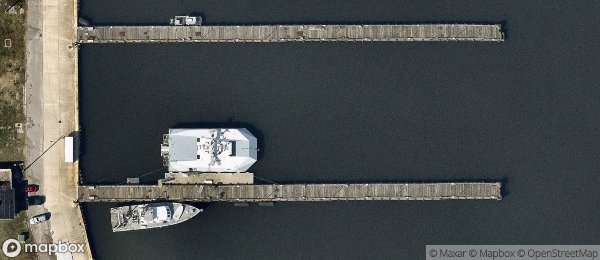}\label{fig:norfolk_sat}}\\
 \caption{\textbf{Real world environments.} (a) shows our experimental setup at SUNY Maritime college. (b) shows a satellite image of (a). (c) shows a ground level view of Penn's landing with (d) showing a satellite view. (e) shows a ground level view of the marina in Norfolk, VA with (f) showing a satellite image view. }%
 \label{some-label}%
\end{figure*}

This section will showcase three experiments with real-world data collected with our robot. In this real-world case study, ground truth data is unavailable, so we will analyze only coverage and runtime using the same means as the simulation study. The three environments are (1) SUNY Maritime College's marina in The Bronx, NY, (2) Penn's Landing marina in Philadelphia, PA, and (3) a marina adjacent to Norfolk Naval base in Norfolk, VA. SUNY Maritime is shown in Figure \ref{fig:suny_img} and from satellite view in Figure \ref{fig:suny_sat}; note the heavy presence of circular pilings, repeating simple objects. Penn's landing is shown in Figure \ref{fig:penn_img} and from satellite view in Figure \ref{fig:penn_sat}; note the presence of several large complex structures and few repeating objects. Norfolk is shown in Figure \ref{fig:norfolk_img} with satellite view show in Figure \ref{fig:norfolk_sat}. Note that this environment has two long edge structures, the concrete wall on the left of the marina and the long floating dock on the bottom, with the rest of the structures composed of repeating pier pilings. 

\begin{figure}[t]%
 \centering
 \subfloat[\textbf{SUNY Maritime College} ]{\includegraphics[width=0.68\columnwidth]{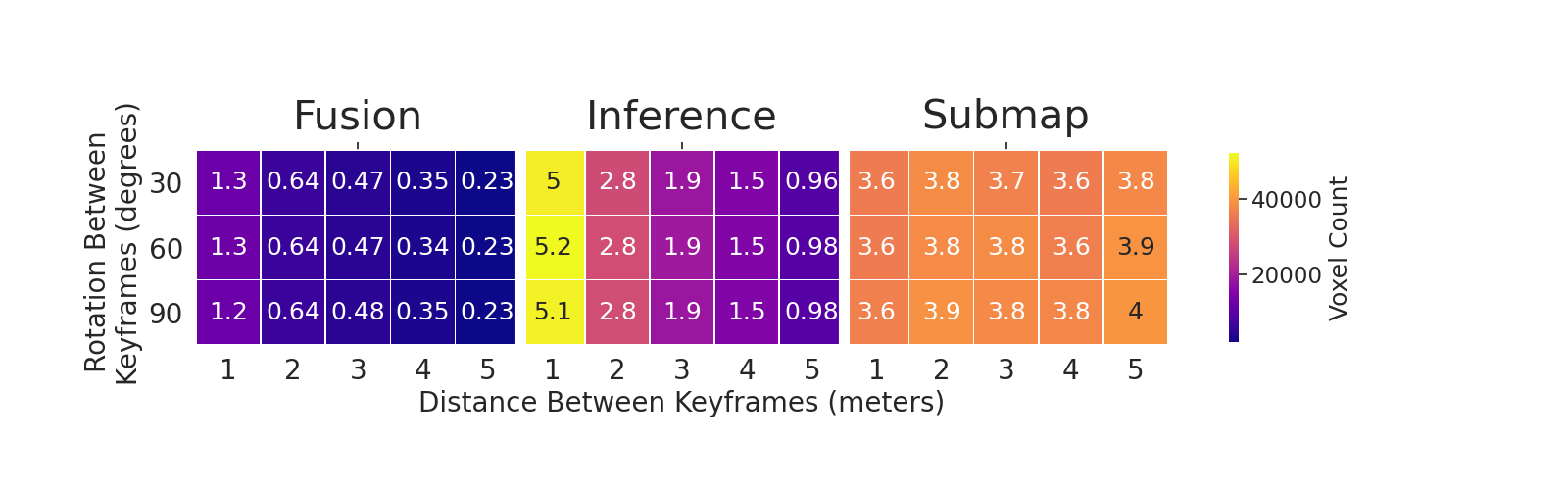}\label{fig:suny_heat}}\\
 \subfloat[\textbf{Penn's Landing} ]{\includegraphics[width=0.68\columnwidth]{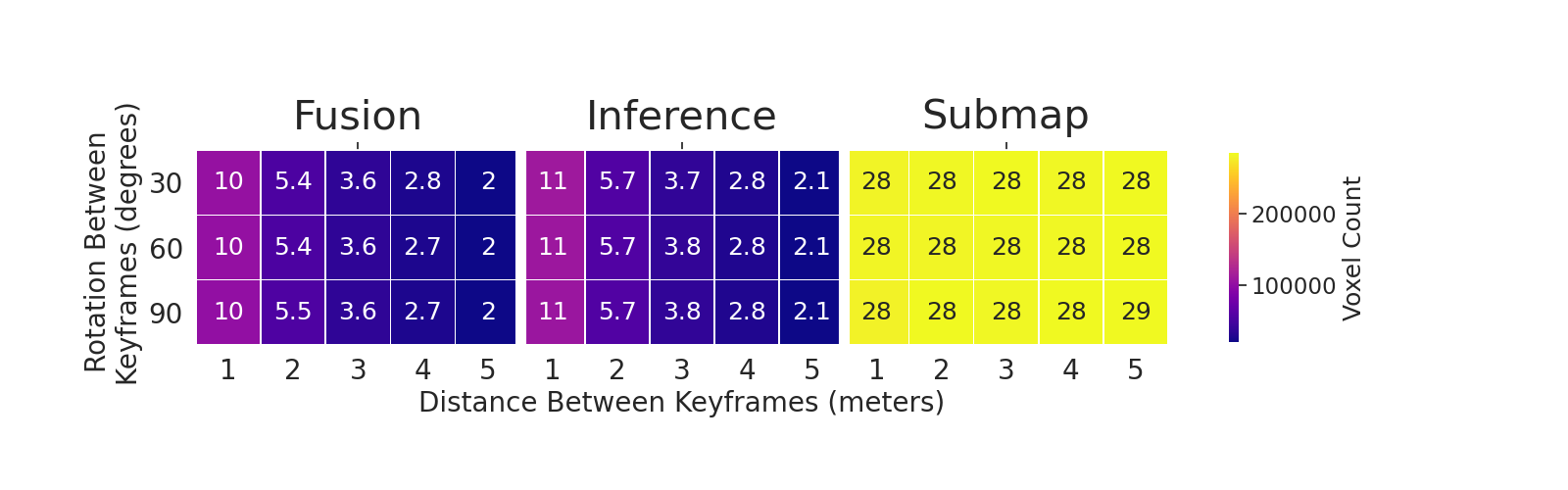}\label{fig:penn_heat}}\\
  \subfloat[\textbf{Norfolk} ]{\includegraphics[width=0.68\columnwidth]{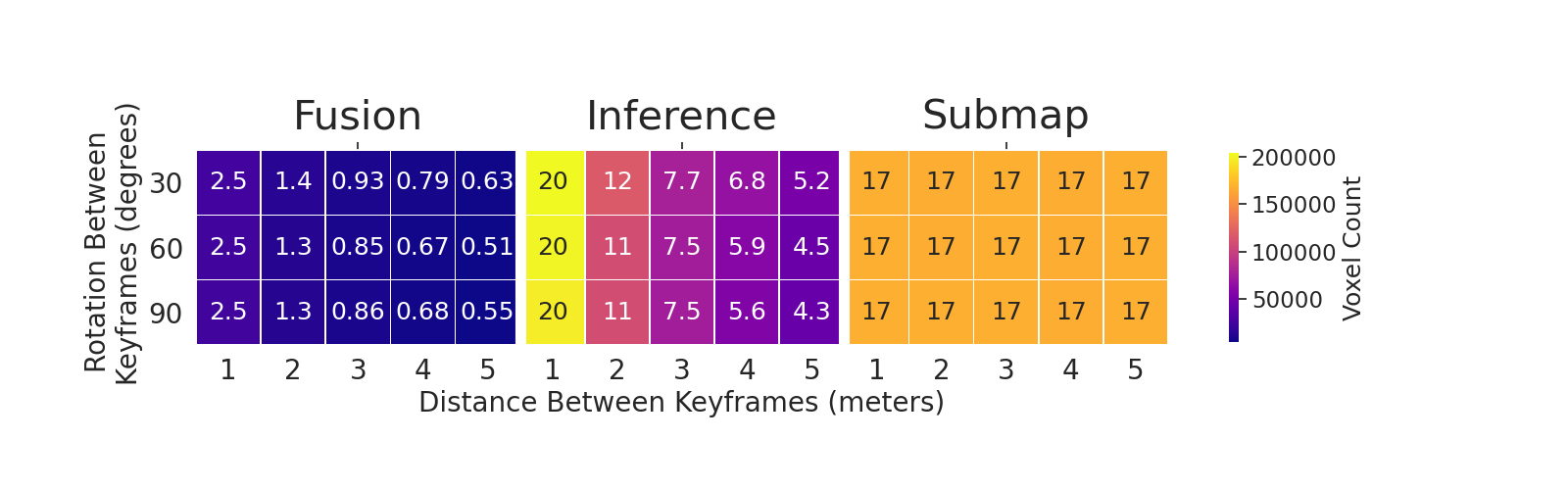}\label{fig:norfolk_heat}}
\caption{\textbf{Real world coverage results.} Coverage in both table and color format. Each cell reports voxel count, with color mapping from blue to orange as low to high coverage. The vertical axis shows the varying keyframe rotations in degrees with the horizontal axis showing Euclidean distance between keyframes in meters. Each system type is shown in this figure, with sonar fusion mapping at left, inference based mapping at center and submapping at right. Note, the voxel count is colored according to the scale bar given at the right of each subfigure.}
\end{figure}

\subsubsection{SUNY Maritime Results}
At SUNY Maritime college, we see a coverage trend consistent with our simulation experiments, shown in Figure \ref{fig:suny_heat}.   Qualitative results are shown in Figures \ref{fig:suny_fusion_cloud}, \ref{fig:suny_infer_cloud} and \ref{fig:suny_submap_cloud}. The estimated trajectory followed by the robot is shown in Fig. \ref{fig:suny_traj}. Inference based mapping makes significant coverage improvements over fusion based mapping. Moreover, submapping shows good coverage, better than fusion based mapping. However, inference based mapping shows better coverage results than submapping with its densest set of keyframes. Note that SUNY Maritime has very few complex structures and is characterized by many repeating pier pilings, an excellent use case for inference based mapping. 

\subsubsection{Penns Landing Results}
Penn's Landing has very few repeating objects but more complex structures, and our trajectories include observations of a ship hull and barge. Qualitative results are shown in Figures \ref{fig:penn_fusion_cloud}, \ref{fig:penn_infer_cloud} and \ref{fig:penn_submap_cloud}. Figure \ref{fig:penn_heat} shows coverage results from Penn's Landing. The estimated trajectory followed by the robot is shown in Fig.  \ref{fig:penn_traj}. We note the small performance improvement in inference based mapping compared to fusion based mapping, likely due to the small number of repeating objects. Submapping shows the highest coverage in this environment and is flat across varying keyframe densities. Note the dense point clouds recovered from the barge and ship in Figure \ref{fig:penn_submap_cloud}. 

\subsubsection{Norfolk Results}
\begin{figure*}[t]%
 \centering

    \subfloat[\textbf{SUNY Maritime Sonar Fusion Map} ]{\includegraphics[height= 3cm]{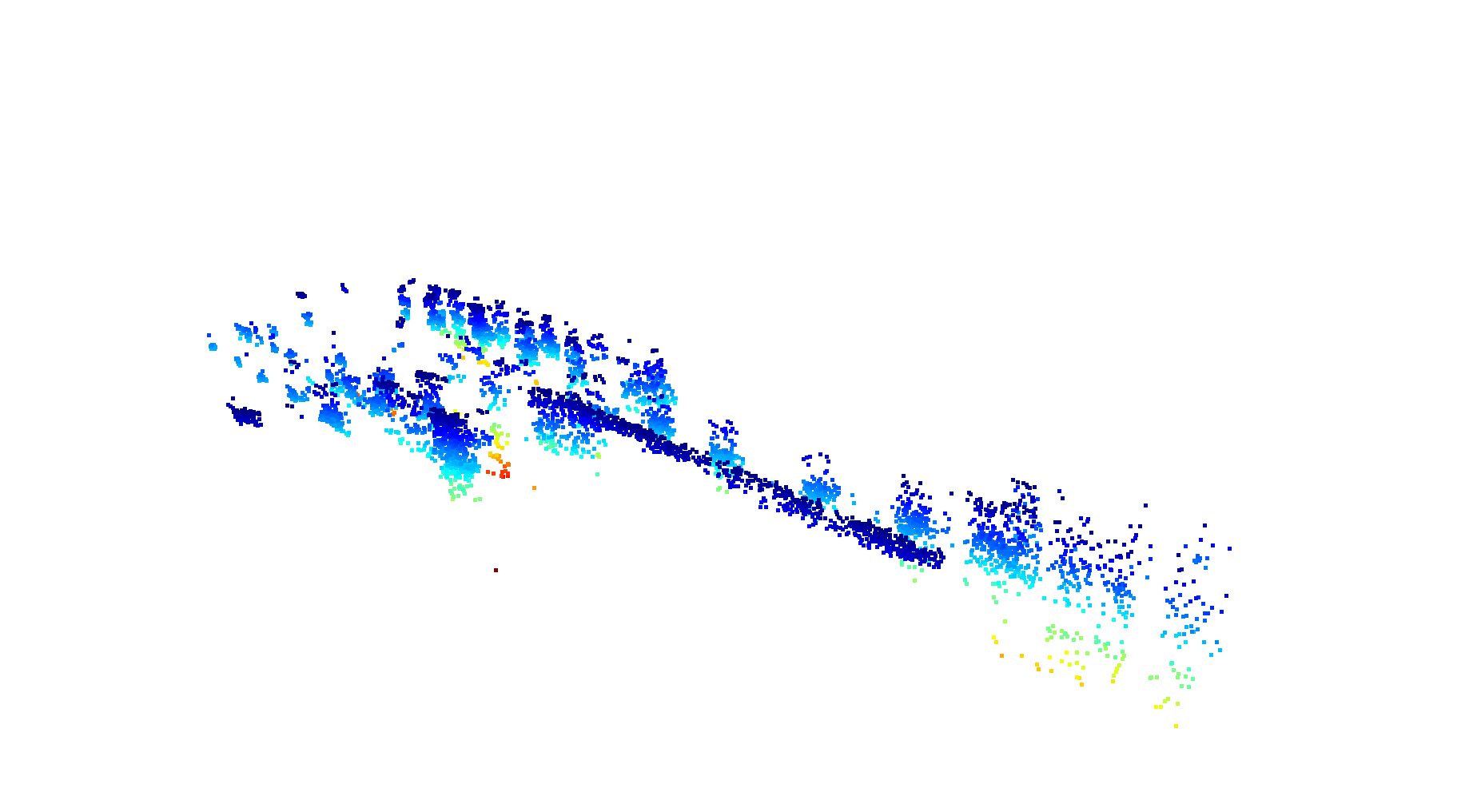}\label{fig:suny_fusion_cloud}}%
 \subfloat[\textbf{SUNY Maritime Inference Map} ]{\includegraphics[height= 3cm]{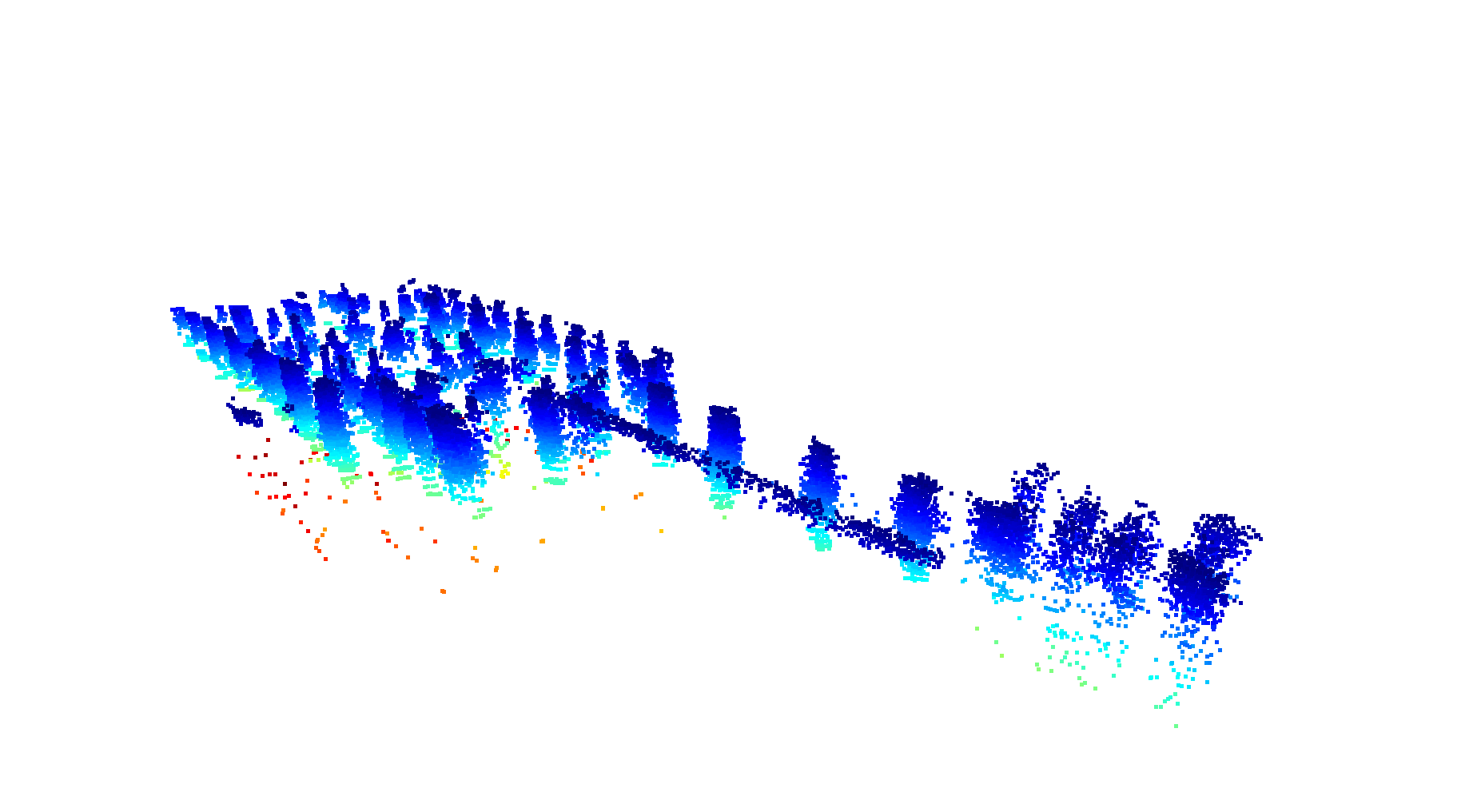}\label{fig:suny_infer_cloud}}%
  \subfloat[\textbf{SUNY Maritime Submapping} ]{\includegraphics[height= 3cm]{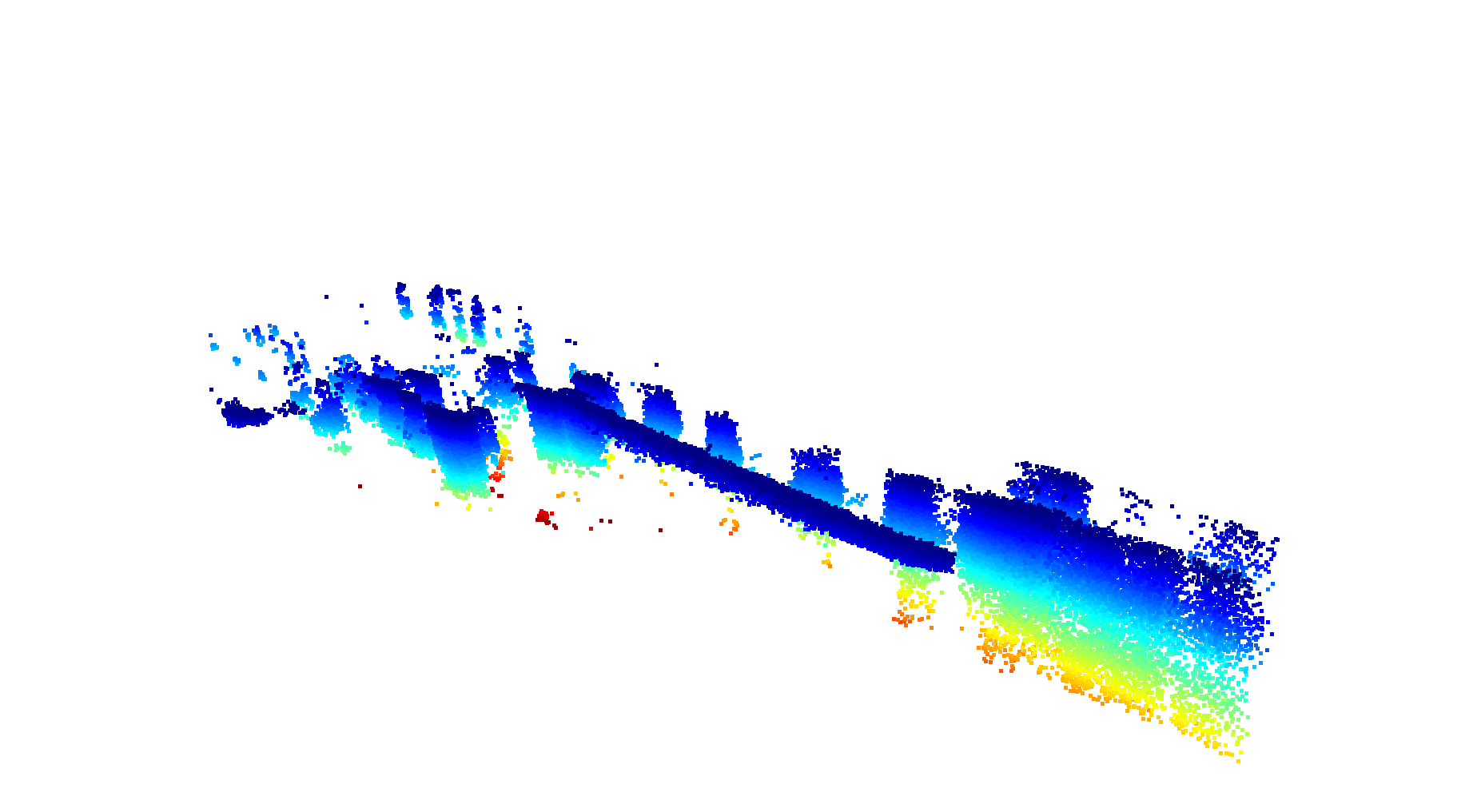}\label{fig:suny_submap_cloud}}\\

  \subfloat[\textbf{Penn's Landing Sonar Fusion Map} ]{\includegraphics[height= 3cm]{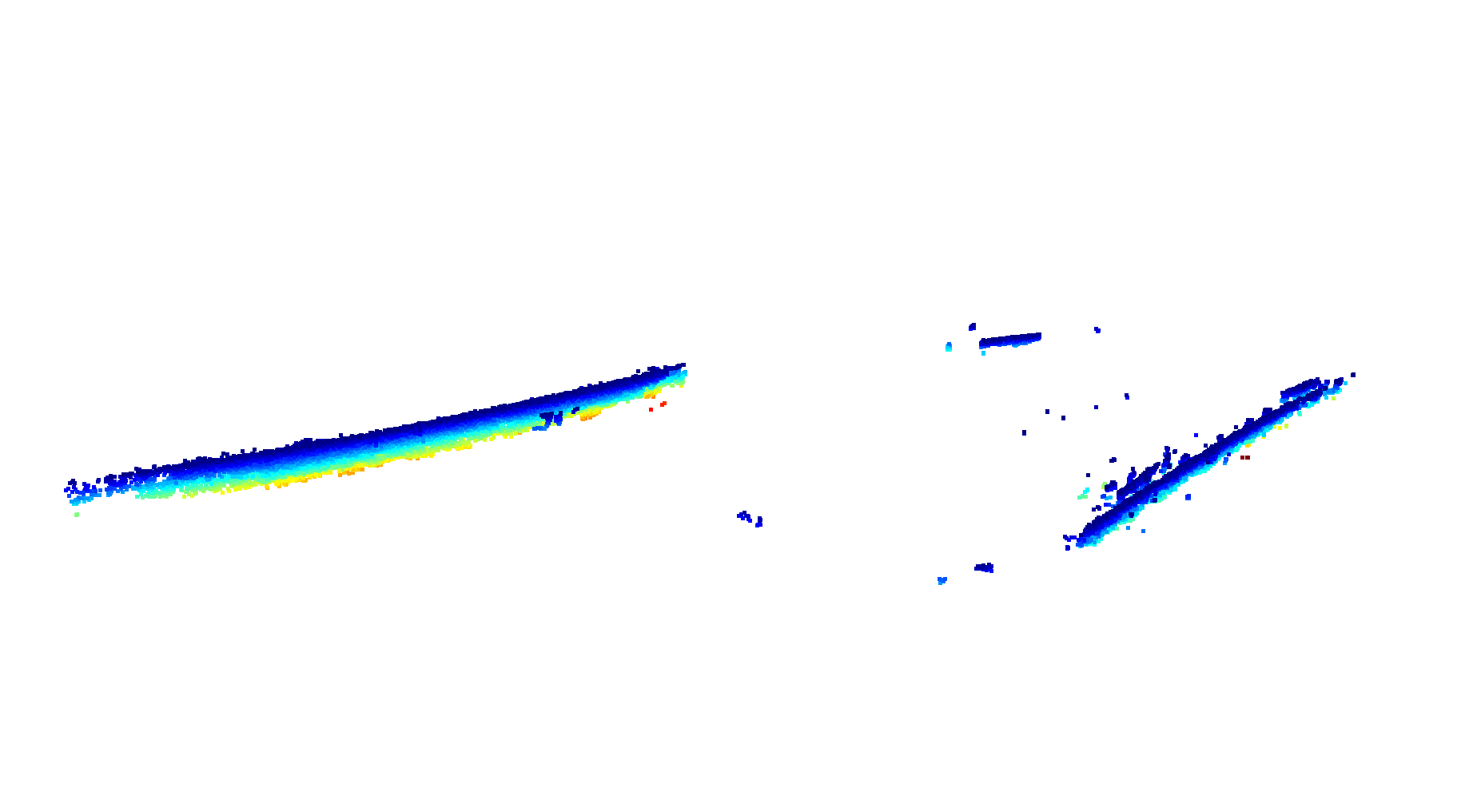}\label{fig:penn_fusion_cloud}}%
 \subfloat[\textbf{Penn's Landing  Inference Map} ]{\includegraphics[height= 3cm]{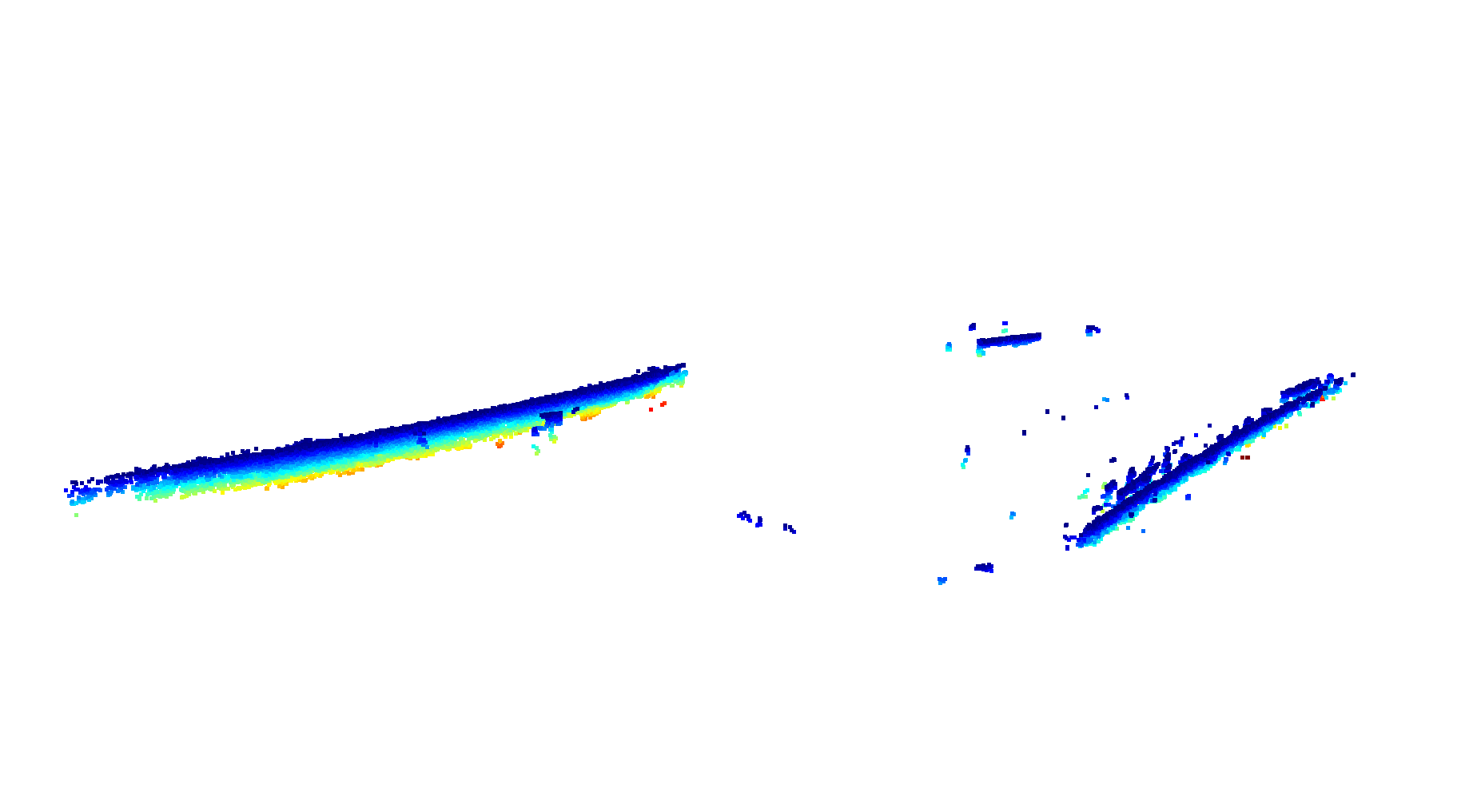}\label{fig:penn_infer_cloud}}%
  \subfloat[\textbf{Penn's Landing Submapping} ]{\includegraphics[height= 3cm]{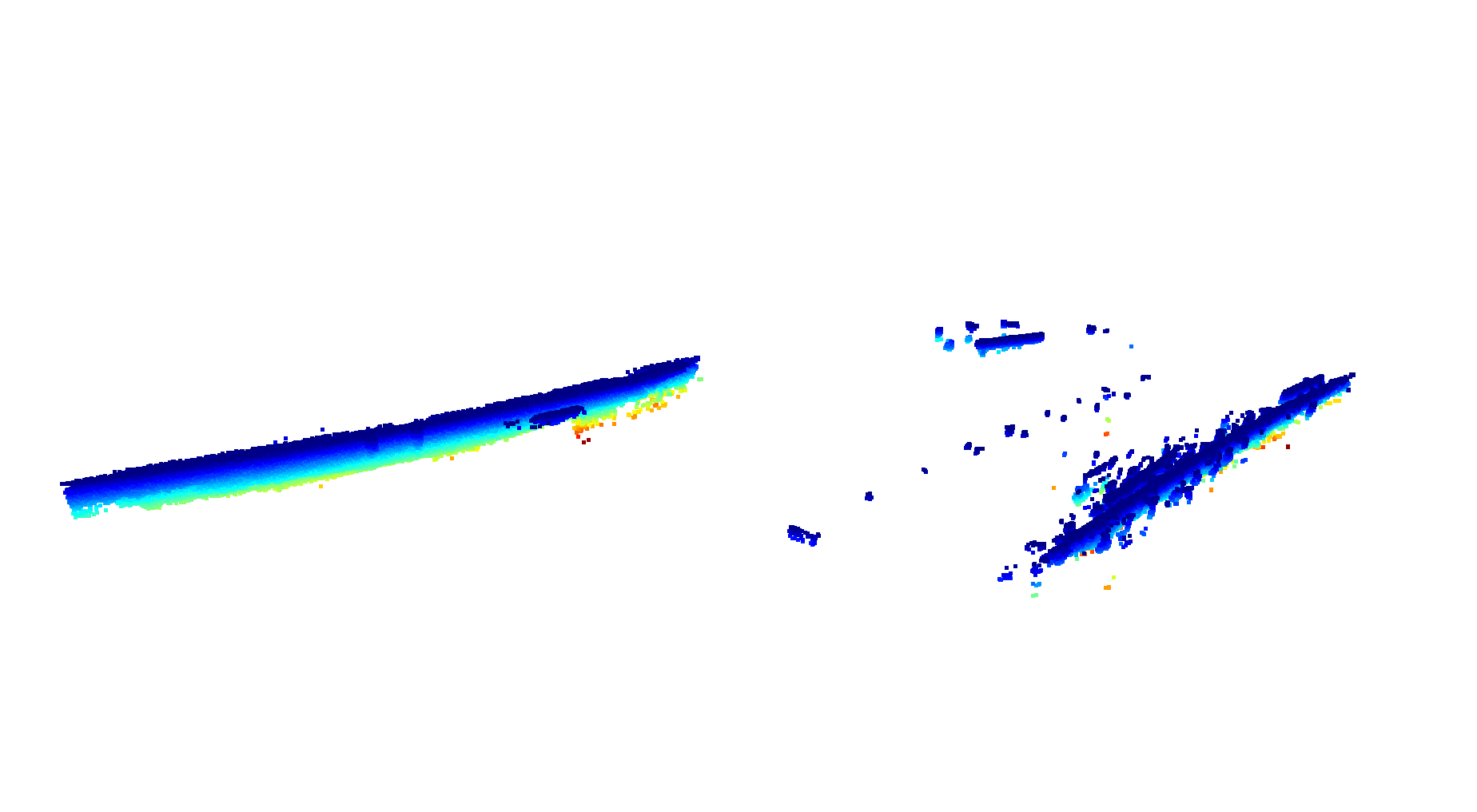}\label{fig:penn_submap_cloud}}\\

  \subfloat[\textbf{Norfolk Sonar Fusion Map} ]{\includegraphics[height= 3cm]{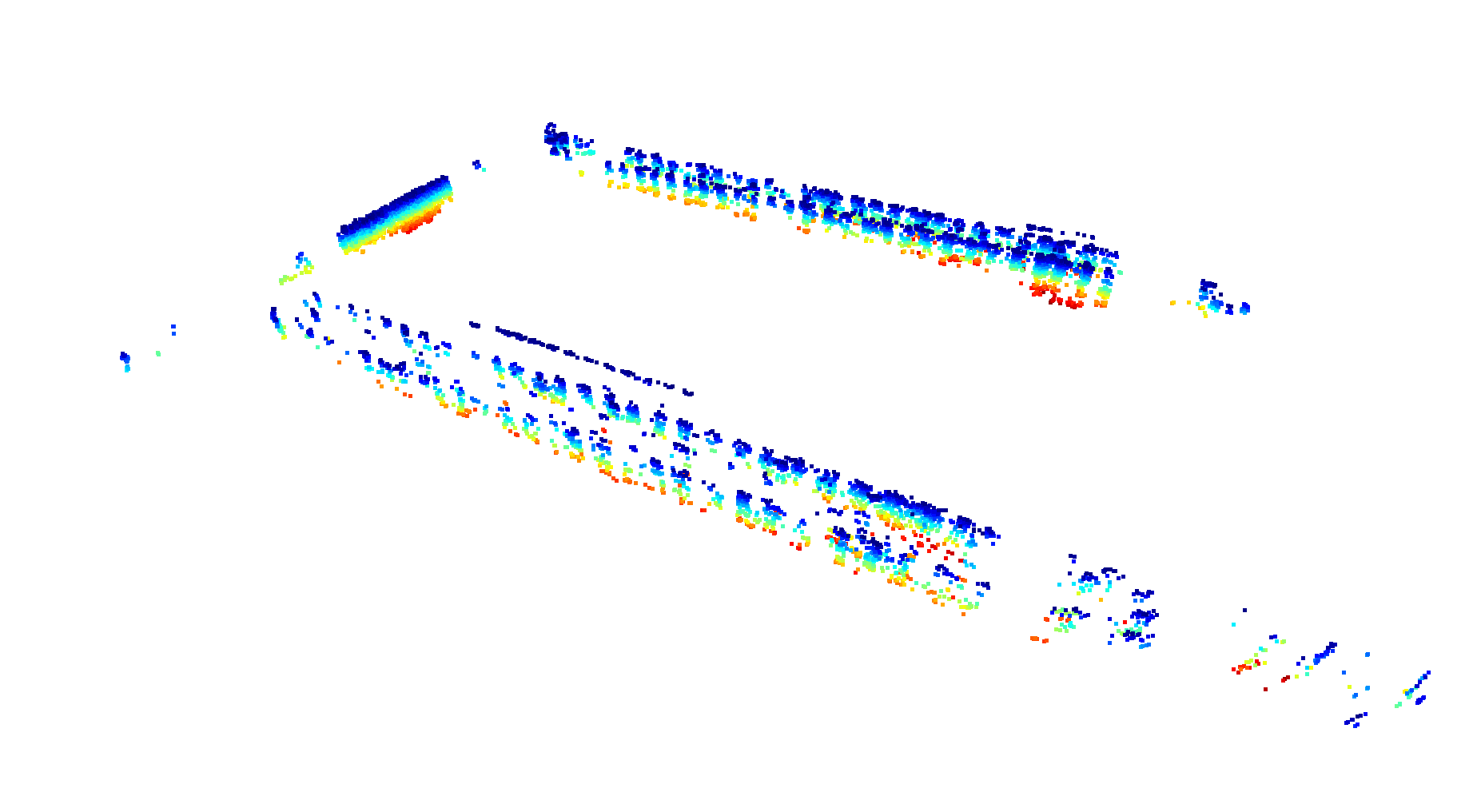}\label{fig:nor_fusion_cloud}}%
 \subfloat[\textbf{Norfolk  Inference Map} ]{\includegraphics[height= 3cm]{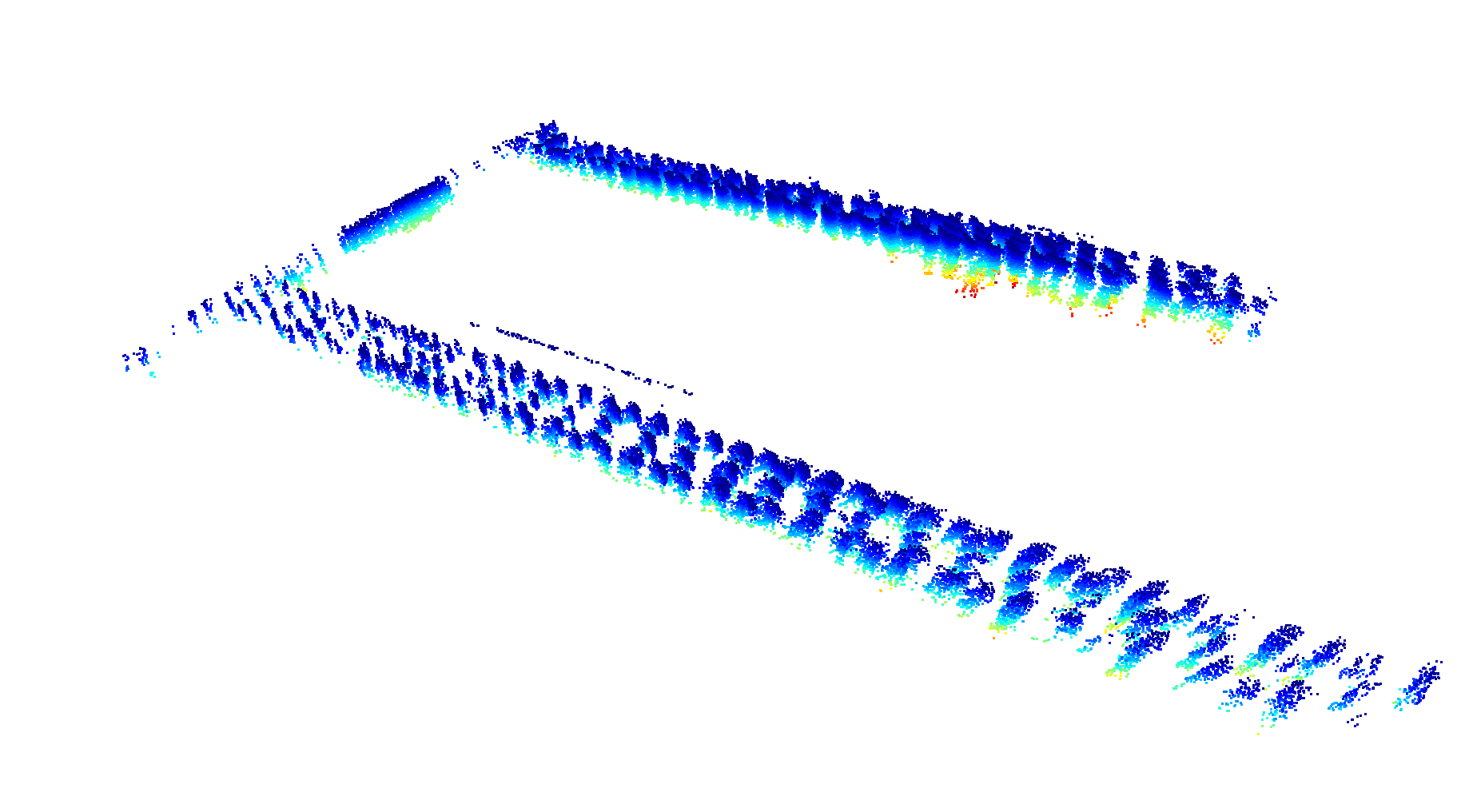}\label{fig:nor_infer_cloud}}%
  \subfloat[\textbf{Norfolk Submapping} ]{\includegraphics[height= 3cm]{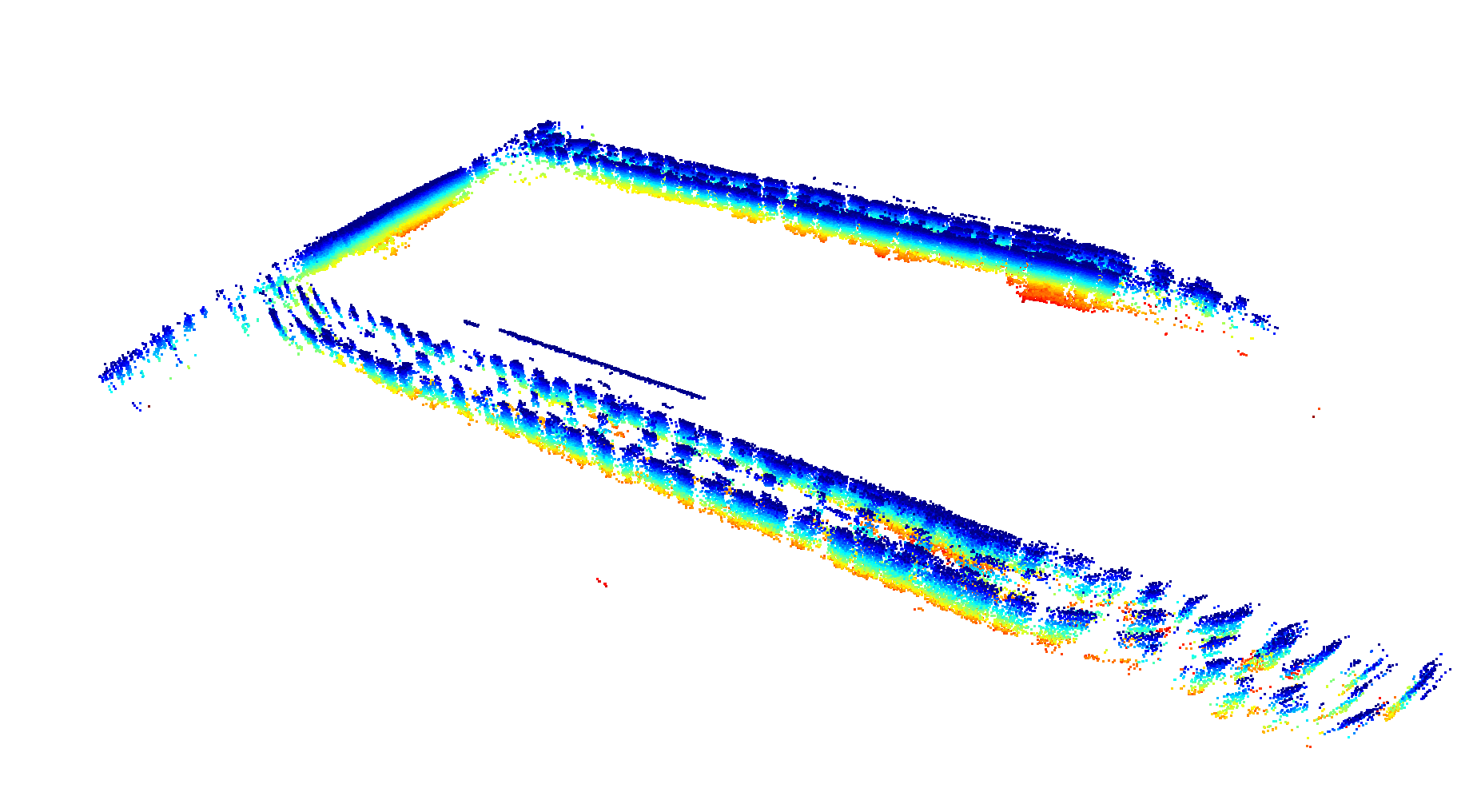}\label{fig:nor_submap_cloud}}\\
  
\caption{\textbf{Real World Results.}}
\end{figure*}

\begin{figure*}[t]%
 \centering

    \subfloat[\textbf{SUNY Maritime Trajectory} ]{\includegraphics[height= 4cm]{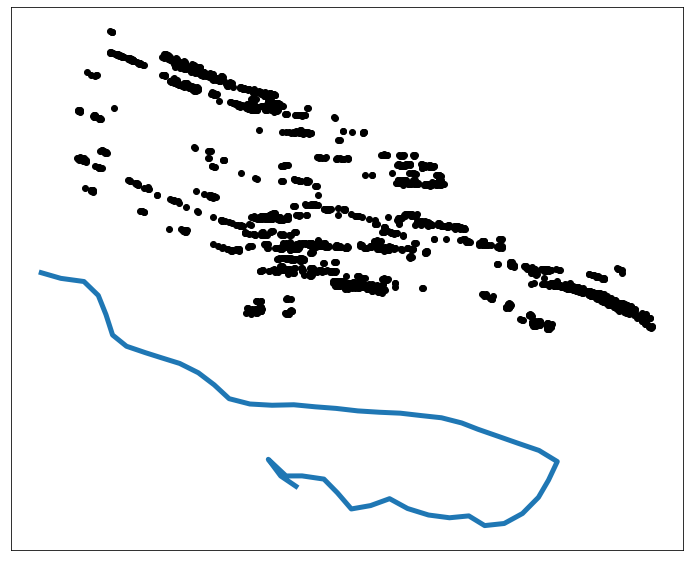}\label{fig:suny_traj}}
  \subfloat[\textbf{Penn's Landing Trajectory} ]{\includegraphics[height= 4cm]{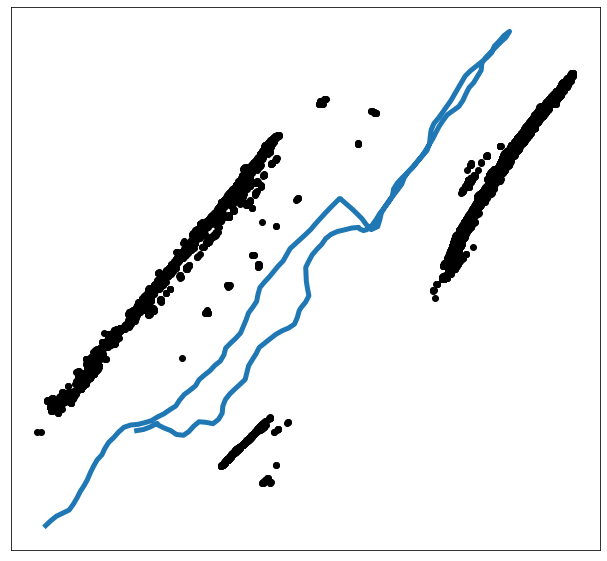}\label{fig:penn_traj}}
  \subfloat[\textbf{Norfolk Trajectory} ]{\includegraphics[height= 4cm]{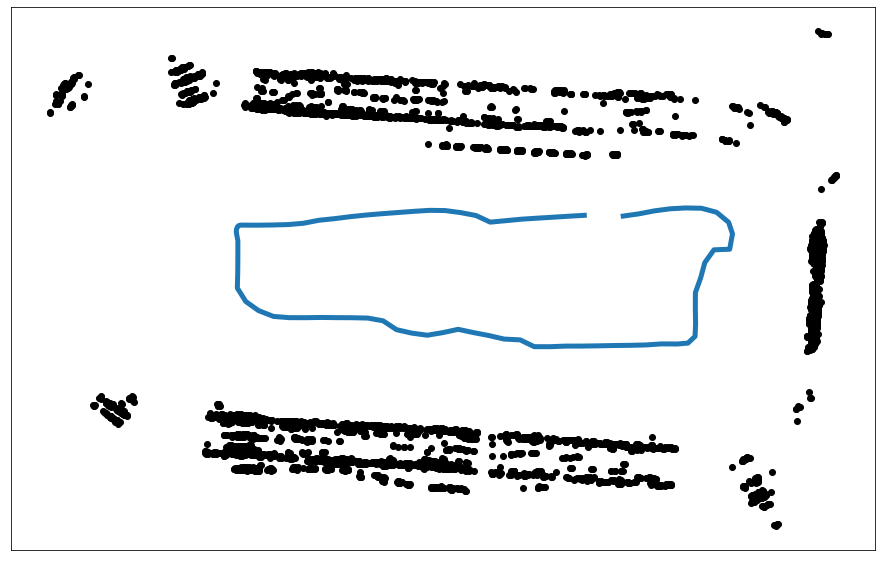}\label{fig:norfolk_traj}}\\
  
\caption{\textbf{Real Environment Robot Trajectories.} Blue lines indicate the robot path, black points indicate the sonar observations.}
\end{figure*}

Norfolk shows a similar trend to SUNY Maritime college and our simulation experiments, as shown in Figure \ref{fig:norfolk_heat}. The presence of repeating objects means inference based mapping is able to recover the most dense version of the map, but submapping is consistent across all parameterizations. We especially note the increase in coverage on the concrete wall at the back of the map when using submapping, shown in Figure \ref{fig:nor_submap_cloud}. The estimated trajectory followed by the robot is shown in Fig. \ref{fig:norfolk_traj}. We do note some rotational distortion, likely due to gyroscope error. 

\begin{table}[t]
\vspace{-0mm}
\centering
\begin{tabular}{ccccc}
\toprule
 & \multicolumn{4}{c}{Environment} \\
 & \multicolumn{2}{c}{SUNY Maritime} & \multicolumn{2}{c}{Penn's Landing}\\
\midrule
 System & Mean & SD & Mean & SD \\
 \midrule
Orthogonal Sensor Fusion  &  0.172 & 0.030 & 0.207 & 0.024\\
Mapping Inference         &  0.947 & 0.251 & 0.405 & 0.253\\
Submap Construction       &  0.003 & 0.005 & 0.003 & 0.003\\
\toprule
\end{tabular}
\caption{\textbf{Real world runtime in seconds.} We report mean and standard deviation (SD) runtime for all experiments for a particular system in each environment. Orthogonal sensor fusion refers to the proccess of fusing a single sonar image pair. Mapping inference is the time required to apply the method in section \ref{infer_map} to a single keyframe. Submap construction is the time required to assemble a submap from a single timestep once all the data is collected. }
\label{runtime_real}
\end{table}

\subsubsection{Real World Runtime}
Table \ref{runtime_real} shows the runtime for these real-world datasets. Submap construction again does not add undue computational burden to our system, and performing inference on keyframes takes time, making it clear that inference should only be applied to keyframes. Orthogonal sensor fusion time runs faster or near the sonar refresh rate of 5Hz. 

\subsubsection{Real World Results Summary}
These real-world datasets make the results of this study clear; when repeating objects are present, and a semantic model can be provided to leverage those objects, inference based mapping provides excellent coverage. However, when a model is unavailable or when these repeating objects are not prevalent, submapping provides better coverage, especially when considering complex objects like ship hulls. However, sonar fusion mapping provides adequate results when neither a model nor a short-term dead reckoning system is available.

\section{Conclusions}
\label{conc}
In this paper, we have presented an extension of our existing 3D underwater sonar mapping system. We have proposed utilizing the data available between SLAM keyframes by building submaps, moving toward a capability for densely mapping any underwater environment using an orthogonal imaging sonar fusion system. We have integrated this submapping system with our existing, open-source SLAM system\footnote{\url{https://github.com/jake3991/sonar-SLAM}} and demonstrated its utility in two real-world environments, including building a realistic ship map. We compared this new submapping system to our previous work with orthogonal imaging sonars. We have shown that when the environment contains many simple repeating objects and a trained model is available, inference based mapping is the tool of choice. However, when reliable short-term dead reckoning is available, and structures are highly complex, such as aircraft, submapping provides the superior option. We have also shown that our most basic system is adept at building reasonable maps with few system requirements. 

However, there remain some open questions. Chief among them is how to combine the value of submapping with that of inference. If an environment has both complex structures \textit{and} simple repeating objects, how can the runtime limitations of the inference system be overcome? We aim to explore this question in future work.

\section*{Acknowledgments}
This research was supported in part by ONR Grant N00014-21-1-2161, USDA-NIFA Grant 2021-67022-35977, NSF Grant IIS-1652064, and by a grant from Schlumberger Technology Corporation.

\ifCLASSOPTIONcaptionsoff
  \newpage
\fi

%
\bibliographystyle{IEEEtran}

\vspace{-10mm}

 \begin{IEEEbiography}[{\includegraphics[width=1in,height=1.25in,clip,keepaspectratio]{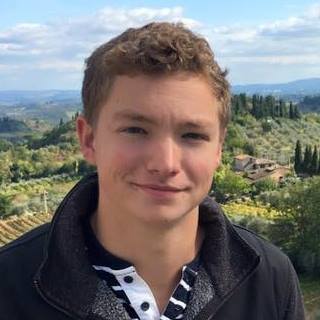}}]{John McConnell}  is an Assistant Professor at the United States Naval Academy, Annapolis, MD, USA. He received a Bachelor's Degree in Mechanical Engineering from SUNY Maritime College, The Bronx, NY, USA, in 2015, a Master of Engineering in Mechanical Engineering from Stevens Institute of Technology, Hoboken, NJ, USA, in 2020, and a Ph.D. in Mechanical Engineering from the same institution in 2023. Prior to joining Stevens, he worked in engineering roles with the American Bureau of Shipping, ExxonMobil, and Duro UAS.
 \end{IEEEbiography}

 \begin{IEEEbiography}[{\includegraphics[width=1in,height=1.25in,clip,keepaspectratio]{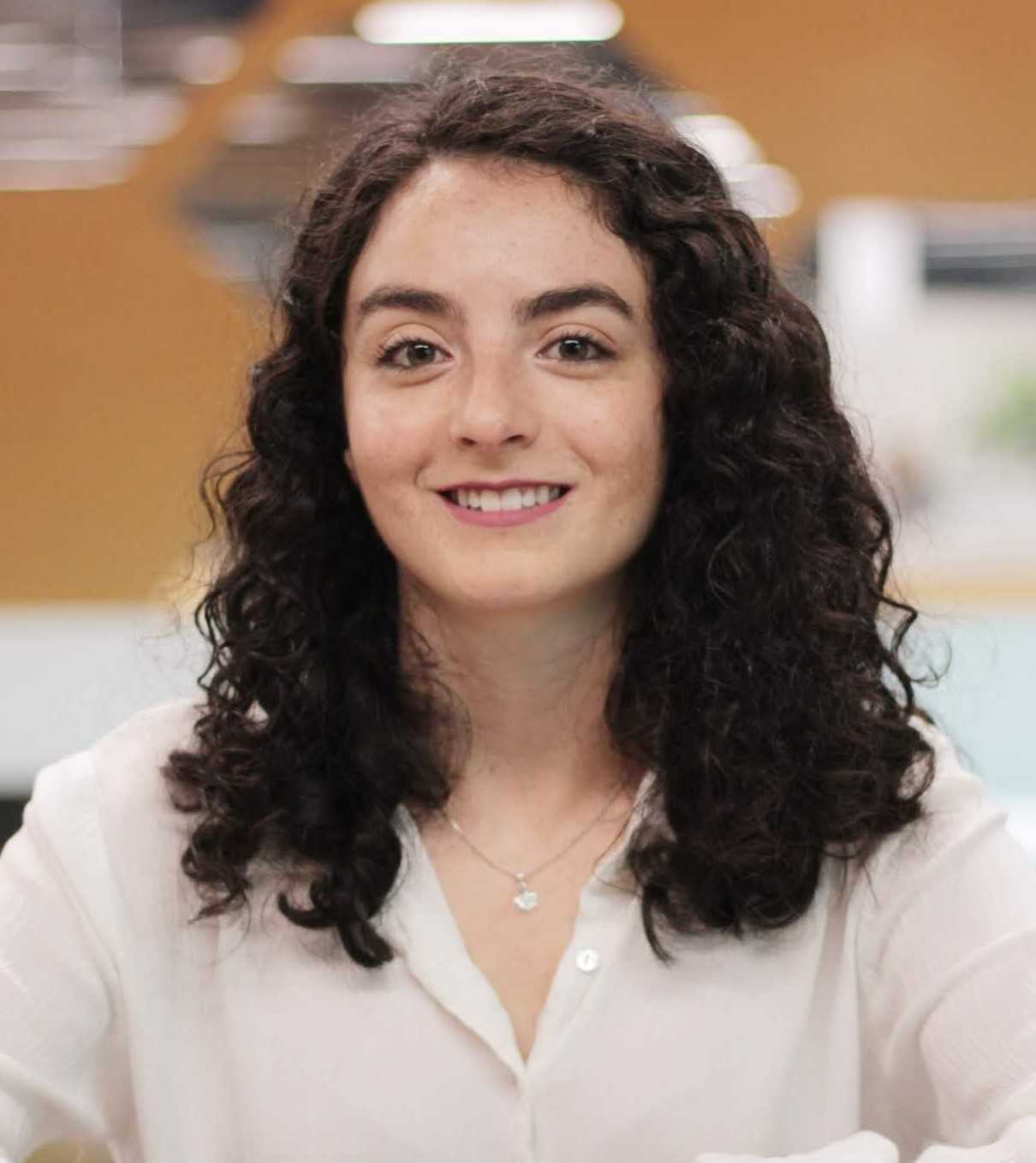}}]{Ivana Collado-Gonzalez}
 received the bachelor’s degree in mechatronics engineering from Tecnologico
de Monterrey, Monterrey, Mexico, in 2021, and the
Master of Engineering in robotics in 2023 from the
Stevens Institute of Technology, Hoboken, NJ, USA,
where she is currently working toward the Ph.D. degree in mechanical engineering with the Department
of Mechanical Engineering.
Prior to joining Stevens, she worked with Xlab
Protexa R\&D, on the development of an autonomous
mobile robot.
 \end{IEEEbiography}

 \begin{IEEEbiography}[{\includegraphics[width=1in,height=1.25in,clip,keepaspectratio]{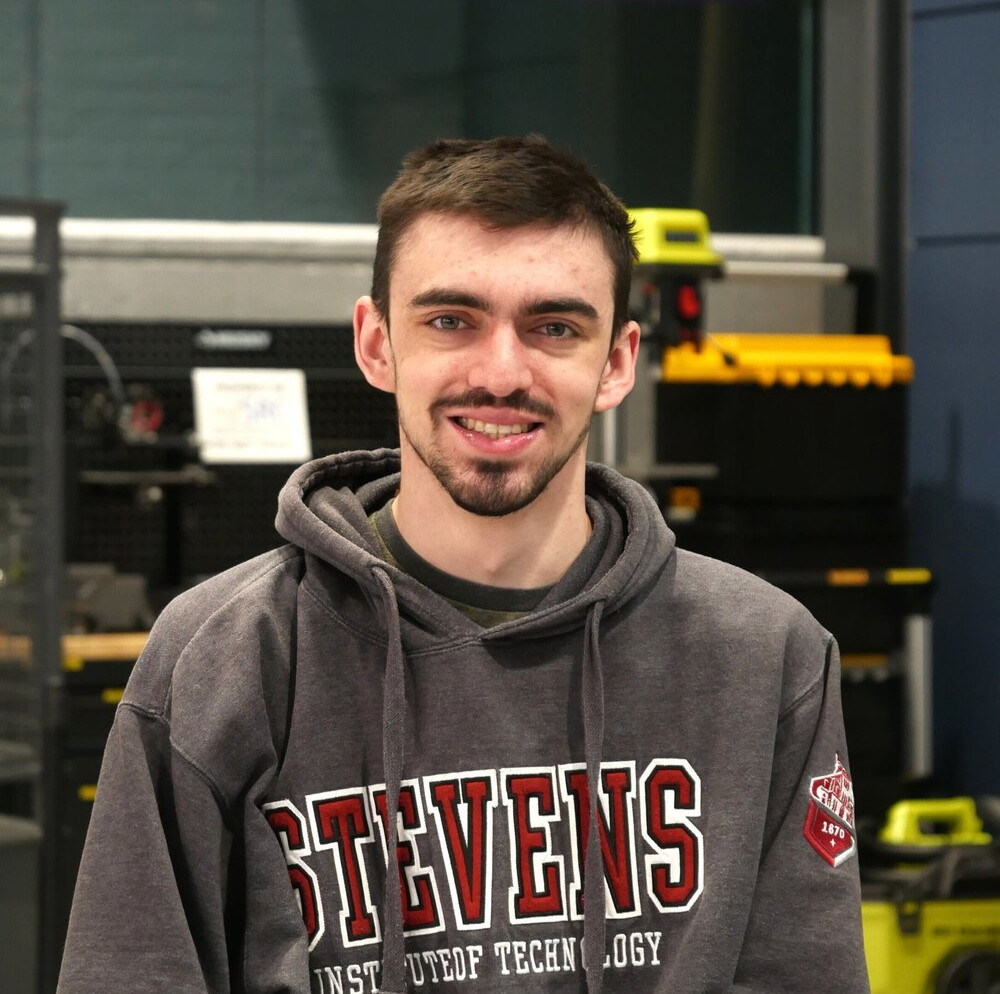}}]{Paul Szenher}
received a Bachelor of Engineering in Computer Engineering from Stevens Institute of Technology, Hoboken, NJ, USA, in 2020, and a Master of Science in Computer Science from the same institution in 2023. He is currently a Ph.D. student in the 
Department of Mechanical Engineering, Stevens Institute of Technology.
 \end{IEEEbiography}


 \begin{IEEEbiography}[{\includegraphics[width=1in,height=1.25in,clip,keepaspectratio]{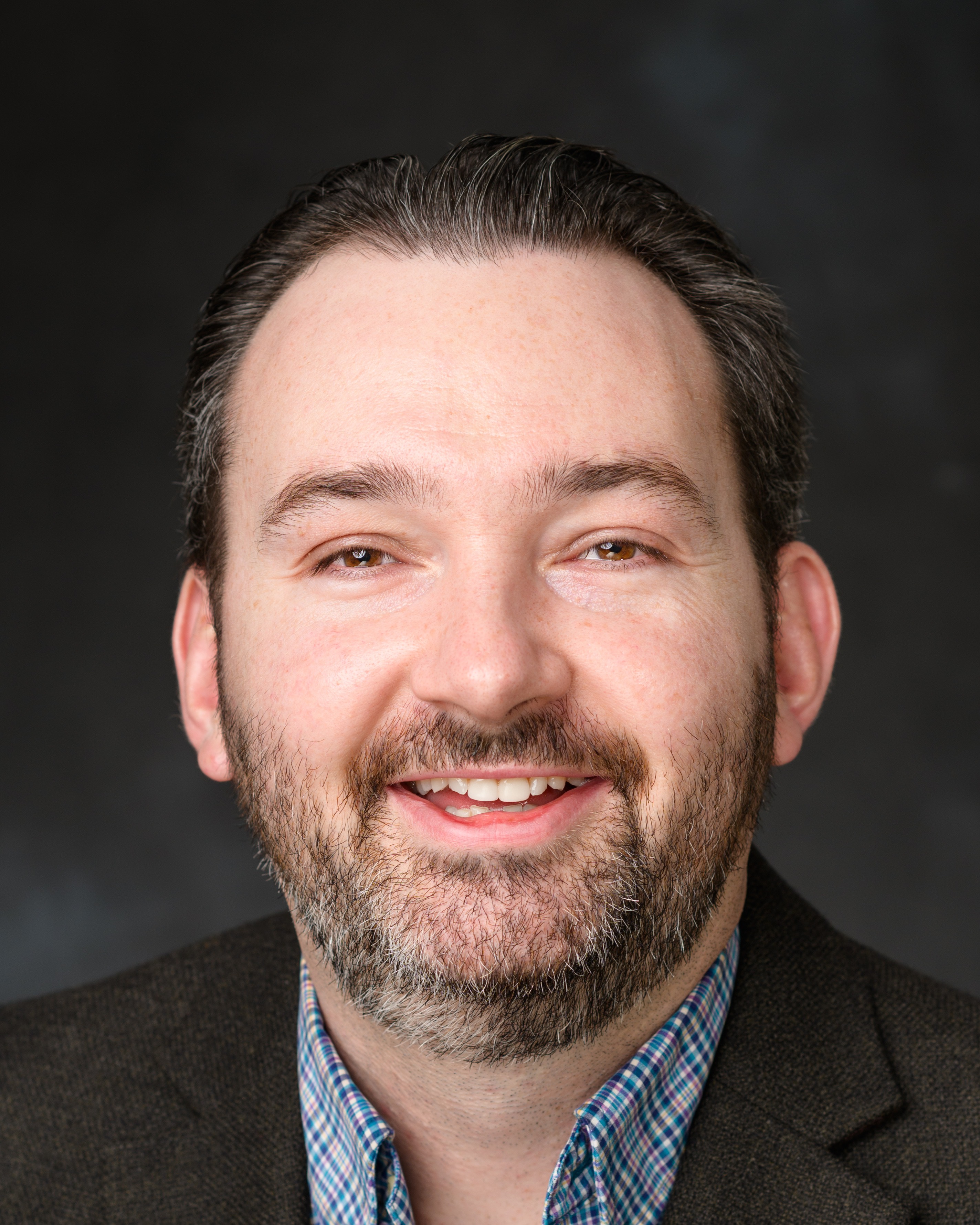}}]{Brendan Englot}
 (S'11--M'13--SM'20) received S.B., S.M., and Ph.D. degrees in Mechanical Engineering from Massachusetts Institute of Technology, Cambridge, MA, USA, in 2007, 2009, and 2012, respectively.

 He was a research scientist with United Technologies Research Center, East Hartford, CT, USA, from 2012 to 2014. He is currently an Associate Professor with the Department of Mechanical Engineering, Stevens Institute of Technology, Hoboken, NJ, USA. His research interests include motion planning, localization, and mapping for mobile robots, learning-aided autonomous navigation, and marine robotics. He is the recipient of a 2017 National Science Foundation CAREER award and a 2020 Office of Naval Research Young Investigator Award. 
 \end{IEEEbiography}

\vfill

\end{document}